\documentclass[11pt, a4paper, notitlepage]{report}
\usepackage{hyperref}
\hypersetup{colorlinks,%
citecolor=black,%
filecolor=black,%
linkcolor=black,%
urlcolor=black}
\usepackage{amsmath}
\usepackage{graphicx}
\usepackage{caption}
\usepackage{subcaption}
\usepackage[ruled, vlined, linesnumbered]{algorithm2e}
\usepackage[nottoc]{tocbibind}
\usepackage{hyphenat}
\usepackage{titling}

\author{Asem Khattab\\\href{mailto:a.a.m.f.khattab@student.utwente.nl}{a.a.m.f.khattab@student.utwente.nl}\\ Institute of Transportation,\\ German Aerospace Center (DLR)}
\title{Static and Dynamic Path Planning Using Incremental Heuristic Search}
\date{November 27, 2017}

\begin{document}
\vfill
\maketitle
\vfill
\hrule
\begin{abstract}
Path planning is an important component in any highly automated vehicle system. In this report, the general problem of path planning is considered first in partially known static environments where only static obstacles are present but the layout of the environment is changing as the agent acquires new information. Attention is then given to the problem of path planning in dynamic environments where there are moving obstacles in addition to the static ones. Specifically, a 2D car-like agent traversing in a 2D environment was considered. It was found that the traditional configuration\hyp time space approach is unsuitable for producing trajectories consistent with the dynamic constraints of a car. A novel scheme is then suggested where the state space is 4D consisting of position, speed and time but the search is done in the 3D space composed by position and speed. Simulation tests shows that the new scheme is capable of efficiently producing trajectories respecting the dynamic constraint of a car\hyp like agent with a bound on their optimality.
\end{abstract}
\hrule
\vfill
\thispagestyle{empty}

{\small\tableofcontents}
\thispagestyle{empty}

\newpage
\setcounter{page}{1}
\chapter{Introduction}
Automobile is perhaps one of the most influential inventions of the 20th century. It is widely used as the primary mode of transportation in urban areas. With this significance importance of cars, it is a surprising fact that insufficient innovation in their technology (especially with respect to safety) has occurred in past decades. Until now, they are greatly inefficient with respect to basic resources, like energy and human health. In the United States alone, 42,000 people die annually in nearly 6 million traffic accidents, not to mention the billions of hours wasted in traffic jams~\cite{roboticCars}.

Autonomous driving is a technology that is currently gaining a huge attention both from the industry and from governments as it has a great potential of improving the current situation. The technology is also referred to as driverless car, self-driving car and robotic car. When successfully utilized, and after having enough maturity, the technology is expected to significantly reduce crashes, energy consumption, air pollution and also congestion cost~\cite{roboticCars}. Autonomous cars are equipped by many inboard sensors, connected to a navigation system (and probably to other cars too) and contain fast computers able to process the received data to conceive the environment and take proper actions controlling the car safely. 

\begin{figure}[htb]
\centering
  \includegraphics[height=.4\textheight]{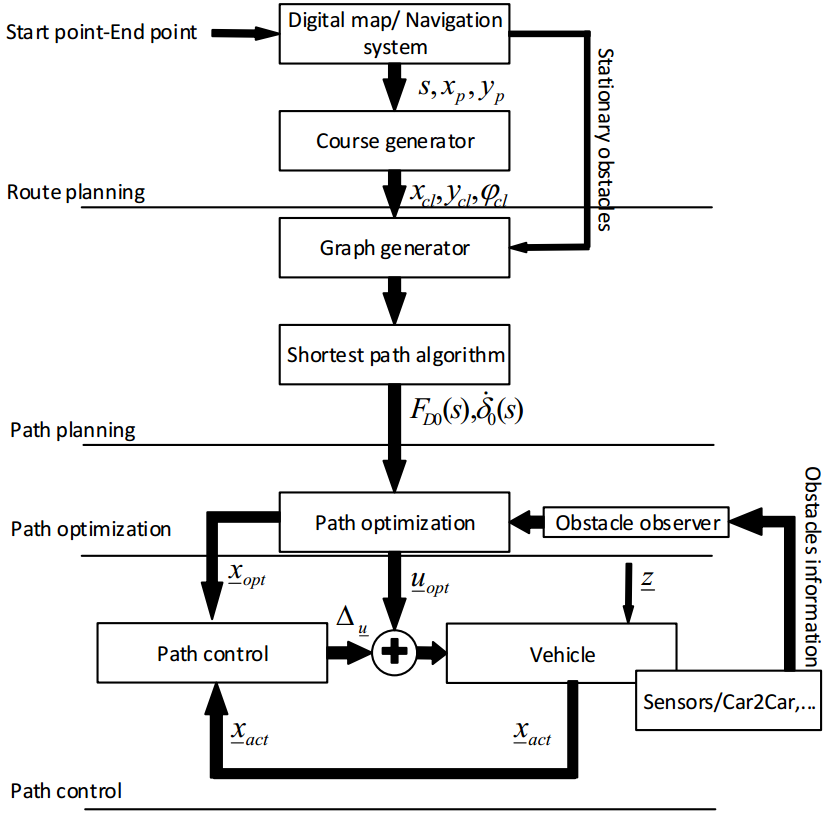}
\caption{A block diagram of the optimal control\hyp based system, taken from \cite{dar}.}
\label{fig:opBlock}
\end{figure}

An autonomous driving system is under development in the institute of transportation in DLR (the German Aerospace Center) based on an optimal control method. Details of the system can be seen in~\cite{dar}. A brief description of the main components of the system is given here for the sake of completeness. A block diagram of the system is shown in Fig.~\ref{fig:opBlock}. At the start, the system gets a start and end points (locations). The first level of the system, named ``route planning'' is where a general course from the start point to the end point is obtained from digital maps and navigation systems. A sensor fusion system is responsible for providing all the needed information about the surrounding area like static obstacles, dynamic obstacles and their expected trajectories. The second level is then concerned about path planning, where a sequence of actions that traverses the car from the start point to the end point is found based on different criteria that will be discussed later. The found path ensures that the car will avoid static obstacles and stay on road. The output of the path planning level serves as an initial solution for the path optimization level. In this level, an optimal control problem is defined and the dynamics of the car are considered by using a single track model. The aim of this level is to find the actuating parameters (driving force and steering angle velocity) that would lead the vehicle along an optimal path with respect to different objectives, like energy consumption, comfort and safety. In the final level, the vehicle is controlled in a feedback loop.

The levels shown in Fig.~\ref{fig:opBlock} are run periodically such that any change in the environment is accounted for. This is done in a very timely manner as the application is highly safety critical. The quality of the initial solution given by the path planning level affect the time needed by the path optimization level to converge. The better the found path is, the faster the convergence. What is meant by ``better'' is the more it is respectful of the kinematic and dynamic constraints and able to avoid static and dynamic obstacles.

This report describes the work done in an internship focused on the second level of the complete system, path planning. The current path planning component used in the system has two levels as seen in Fig.~\ref{fig:opBlock}. First, a graph, composed of nodes representing possible locations the car can be at, is generated. Then, the shortest path is found using A*, an incremental heuristic search method. The locations in the nodes of the generated graph are free from static obstacles (walls, road borders... etc.) and hence the found path from the search is guaranteed to avoid them. However, the current component does not consider dynamic obstacles (other cars, pedestrians... etc.), which are accounted for in the path optimization level. The technical details of the method mentioned here will become more clear shortly through the rest of the report.

The purpose of the internship was to investigate efficient ways to have a dynamic path planning where dynamic obstacles are also considered in addition to static obstacles. The resultant technique should be very fast and should produce trajectories that respect (or are as close as possible to respecting) the kinematic and dynamic constraints of cars. Achieving this would help speeding up the conversion of the system to a solution.

The rest of this report is organized as the following. In the rest of this chapter, the main concepts are briefly introduced and a small literature review is provided, then, the problem is formulated. In the following chapter, several incremental heuristic search algorithms are discussed from the perspective of static partially known environments. The third chapter then discusses the challenges of planning in dynamic environments while applying dynamic and kinematic constraints, and presents a proposed scheme for an efficient planning. The report is then concluded with the main findings.

\section{Concepts and Terminology}
Path planning is, in its general meaning, the task of finding a sequence of actions that allows transitioning from an initial configuration to a goal configuration~\cite{hgu}. Because of the generality of the concept, this report will use the word agent to refer to the object that executes the plan; a term that is commonly used in literature. 

Formally speaking the \emph{configuration} of an agent in its workspace is a set of parameters that define its current state. What a configuration represents is application dependent. For an agent translating in the 2D space, the configuration can be composed of two parameters defining its position (often denoted \emph{x} and \emph{y}). For a robotic arm, the parameters of the configuration can be the angular positions of the arm joints. The minimum number of parameters needed to uniquely describe the configuration of an agent is its \emph{degree of freedom (DoF)}. The parameters of the configuration form what is referred to as a generalized coordinates of the agent. The vector space defined by these coordinates is named the \emph{configuration space}. It includes all possible configurations of an agent and has a dimension equal to the degree of freedom this agent has.

Normally, the workspace of an agent contains obstacles. Configurations that cause the agent to collide with these obstacles are referred to as \emph{forbidden}. Forbidden configurations also result from all kinds of internal constraints of an agent (dynamic and kinematic constraints for example). Generally, a configuration is forbidden if it is not possible or desired for the agent to be at. Other configurations are denoted as \emph{free}.

Based on this, the task of path planning is to find a continuous path from the start configuration to the goal configuration where all the configurations along the path are members of the free space. If such path is not available, failure should be reported. In addition to that, quality measures can be defined for paths and the task of path planning may include optimization with respect to some or all of these measures. For instance, a planner that finds the shortest path for a translating agent optimizes the distance.

\section{Literature Review}

The problem of path planning has been given significance attention in research both as a general problem in fields like artificial intelligence and as an application specific problem. 

Two main ways exist for representing the environment over which the planning is performed ~\cite{phD}. The first one is Artificial Potential Field. In this approach, the motion of the agent is governed by potential force fields applied over the environment. Usually the goal configuration applies an attractive force and obstacles apply repulsive forces on the agent. The agent then goes through the steepest descent in these forces fields until reaching the goal~\cite{apf}.
Potiential field methods are less popular nowadays~\cite{phD}. One of the drawbacks of this technique is that the agent is prone to get stuck in local minima of the potential field, so generally, this method does not guarantee success~\cite{phD}. That is, there might be a valid path although the method has not found one.

Another way of representing the environment is by capturing the connectivity of the configuration into some sort of a \emph{roadmap}. This roadmap should cover the free configuration space such that any free configuration in the environment can be connected to the roadmap in a trivial way. The roadmap should also be connected such that if for any pairs of configurations on the roadmap a valid path exists in the free space, a valid path also exists through the roadmap. A roadmap that satisfies these two conditions allows \emph{complete} path planning methods, that is, if a valid path exists, it will be found. For many applications, it is very hard, or even impossible to generate a roadmap that meets these conditions.

To create a roadmap, the configuration space is discretized by selecting (sampling) configurations. These configurations are added to the roadmap and connections are created between them to capture the connectivity of the free space. The roadmap can then be represented as a directed graph structure on which nodes (or vertices) of the graph are the sampled configurations and a directed edge between two nodes exists if a valid transition from the configuration at the tail of the edge to the configuration at the head exists in the free space. 

Each edge can be associated with a cost that is given based on the wanted characteristics of the desired path. For instance, if the purpose is to minimize distance, the cost would be the distance between the two configuration. In other words, the cost function encodes the criteria of an acceptable path. A path with higher cost would generally be given a lower priority in path planning. An \emph{optimal} path planning method is a one that finds the optimal path, that is, the sum of its edge costs is minimal across all possible paths leading from the start configuration to the goal configuration. 

Different discretization (or sampling) schemes are used to create a road-map for path planning. In complex high-dimensional configuration spaces (e.g. 6 DoF robotic arm), the Probabilistic RoadMap (\textsc{Prm}) method has proven to be particularly useful \cite{phD}. The idea of the method is to randomly sample the configuration space to find free configurations and connect those close to each others by edges. However, in low dimensional configuration spaces, it is both tractable and more practical to use uniform sampling where nodes are uniformly away from each others yet close enough to capture the free and forbidden space and produce reliable plans \cite{dar}. More than that, unlike probabilistic method, the resultant roadmap has the two conditions mentioned above that would allow complete planning.

Generating the roadmap is considered a \emph{preprocessing} phase in path planning and is often done offline. The idea is to put as much effort as needed in creating a good roadmap. Then at real time, the roadmap can be used multiple times to answer queries in a relatively fast way.

There are different approaches in literature for computing paths given some
roadmap of the configuration space. Generally, the popular techniques can be divided to two main categories: deterministic, heuristic-based algorithms (A*-like algorithms) and
randomized algorithms (such as Rapidly-Exploring Random Tree (\textsc{Rrt})). When the planning is done over a low dimensional space,
(i.e. when the degree of freedom of the agent is low), deterministic algorithms are often preferred because they provide a bound\footnote{A factor $K\geq 1$ such that the found path has a cost lower than or equal to the cost of the optimal path multiplied by $K$.} on the quality (or \emph{optimality}) of the returned path~\cite{hgu}. 

\section{Problem Definition and Modeling}\label{prob}
For the application investigated in this report, it is suitable to treat the agent (the autonomous car in this case) as a 2D object moving in the 2D space. This is a sufficient representation since there is no big gain from working in the 3D space.\footnote{Colliding with a tall object is probably not better than colliding with a short one.}

\subsection{Planning in a Static Environment}\label{staticProblem}
The second chapter deals with the problem of an agent navigating in a complex environment where there are stationary obstacles like walls and buildings. Further, this environment contains roads with different degrees of traversability (e.g. different speed limits) in addition to desirable and undesirable areas according to some criteria (known protocols of driving). This agent is required to reach a specific goal state while keeping a minimal cost with regards to all mentioned criteria.

In a practical situation, the agent will never have a complete knowledge of the environment prior to planning. Instead, it may have an initial knowledge of the layout of the environment, including static obstacles, from readily available maps. While it is moving, it acquires new information (using on-board and\slash or off-board sensors) about stationary obstacles, which makes it necessary to continually update the planned path.

In this case, the configuration of the agent has only two parameters indicating its position in the 2D space. A configuration (position) is forbidden if there is an obstacle at it. Because of this representation, path planning returns a path consisting of a sequence of positions with no time information. 

Note that while the configuration space changes (new received information cause previously free configurations to become forbidden and vice versa), it does not contain any information about the dynamics of the environment. So, speed and heading of the moving obstacles are not known. This is what is referred to as a partially-known static space. In this case, even when there are moving obstacles, the configuration space is dealt with as it is at the moment of planning. The agent plans its path by frequently taking a \emph{snapshot} of the environment and finding a trajectory according to this \emph{static} image. This, of course, increases the risk of losing the validity of the plan soon after planning and requires replanning more often. Even with this, the agent may collide with obstacles moving at high speeds. 

Because of the low dimensionality of the problem, its most suitable to use a roadmap generated by uniformly discretizing the configuration space allowing a complete path planning. Uniform sampling in the 2D space creates a grid-like environment, where cells represent configurations (positions of their centers) as seen in Fig.~\ref{fig:conn}. Path planning is then solved as a directed graph search problem using deterministic, heuristic-based algorithms. This allows returning a path with a bound on its optimality. The desired criteria of the found path can be encoded in the edges costs. As new information arrives, the graph is updated (nodes and edges are added, removed and\slash or edge costs are changed) and the plan needs to be updated or recalculated (from scratch).

The second chapter focuses on heuristic-based search algorithms and their use in the context shown here. The purpose of this chapter is not to discuss the codes of theses algorithms rather than to provide insights on them and discuss how they apply to the considered problem. For that, the pseudo-codes of these algorithms are not included here, but can be seen in their respective references\footnote{I suggest consulting \cite{hgu} and \cite{phD} for a well presented view, history and bibliography on the topic.}. It is surely better to know these algorithms from the papers that presented them (all cited in the report).

\subsection{Planning in a Dynamic Environment}\label{dynamicProblem}
For a complete dynamic planning, the planner has to consider the dynamics of the moving obstacles. Ideally, the agent receives information about the speeds and heading angles of the moving obstacles. This information is used to estimate their future positions and planning is done such that the resultant path guarantees safety. This is the concentration of the third chapter. Different approaches exist to do this. One of the popular approaches is to plan in the so called \emph{time-configuration space}~\cite{aPa}. This means adding time as an additional dimension to the configuration space and extrapolating the positions of the moving obstacles along the time dimension. a roadmap is then generated in this 3D space and heuristic\hyp based search is used to find a path. It will be seen that this approach makes it complex to force dynamic and kinematic constraints on the resultant path (such as bounds on the speed and acceleration).

Another approach is to divide the planning task to two levels: one global and one local. On the global level, a path is planned on the configuration space without considering any of the moving obstacles. Then the agent starts moving along this global coarse, avoiding obstacles locally in a dynamic way by controlling its speed and heading angle. One of the most widely used local obstacle avoidance methods is the velocity obstacle approach~\cite{vo}. The idea of this approach is to geometrically construct what is known as the relative velocity space of the agent and the moving obstacles. In this space, the set of velocities that will lead the agent to a collision in the future is defined. The agent then chooses a velocity outside this set based on some predetermined preferences. While this approach nicely allows applying dynamic constraints on the chosen velocities, it doesn't guarantee global optimality as the general path is planned without considering the moving obstacles. Moreover, because of the locality of the approach, it is possible for an agent based on it to get stuck in local minima (e.g. U-shaped static obstacles)~\cite{rvo}.

The third chapter introduces a proposed approach based on incremental search that tries to capture the advantages of both methods discussed above having a global planning with optimality bounds while maintaining dynamic and kinematic constraints. A new online roadmap constructing approach is also presented increasing the flexibility of the agent movements and on the same time preventing the roadmap from explosion.

\chapter{Planning in Partially-Known Static Environments}\label{chapterStatic}
In this chapter, the main incremental heuristic\hyp based graph search algorithms are discussed from the point of view of a 2D agent moving in the 2D space. Then, a simulation grid-based environment designed on MATLAB is described. The discussed algorithms are then implemented and simulated on this environments and tests are done to gain deeper insights about them. The environments dealt with throughout this chapter are static partially\hyp known environments, meaning that only static obstacles are present, but as the agent gets more knowledge while it moves and senses, the perceived structure of the environment continuously changes.
 
\section{Heuristic\hyp Based Path Planning Algorithms}\label{heu}
It is hard to discuss path planning without mentioning Dijkstra's algorithm introduced in 1959~\cite{dij}. If there is a graph $G = (V, E)$ consisting of a set of nodes (or vertices) $V$ and a set of edges between nodes $E\subset V\times V$, and for each edge $(u, v)\in E$ an associated non negative cost $c(u, v)$, Dijkstra's algorithm is able to find optimal paths from a single start node $u_{start}\in V$ to all nodes in $V$. The algorithm is shown below. It starts at the start node and adds to a queue (often named $Open$) all the successors of that node. It maintains for each node $u$ the cost of the best found path to the start $g(u)$. nodes are popped out of the queue with priority given to lower $g$-values. In addition to that, the algorithm maintains for each node $u$ a \emph{back pointer} $bp(u)$ that indicates from which node in the predecessors $Pred(u)$ (the nodes where there are directed edges from them to $u$) does the best found path from $u_{start}$ to $u$ come. The search finishes when the queue becomes empty. If a path was found successfully, it can be traced using back pointers beginning from $u_{goal}$.

\begin{algorithm}[h]
\DontPrintSemicolon
\ForAll{$u \in V$}{
$g(u)\leftarrow \inf$\\
}
$g(u_{start})\leftarrow 0$\\
insert $u_{start}$ into $Open$\\
\Repeat{$Open=\emptyset$}{
$u\leftarrow$ element from $Open$ with minimal $g(u)$\\
Remove $u$ from $Open$\\
\ForAll{$v \in Succ(u)$}{
\If{$g(v) > g(u)+c(u,v)$}{
$g(v) \leftarrow g(u)+c(u,v)$\\
$bp(v)\leftarrow u$\\
insert $v$ to $Open$ or update it if it is in $Open$\\
}
}
}
\caption{Dijkstra\label{Dijkstra}}
\end{algorithm}

With the use of a priority queue, the running time of Dijkstra's algorithm is $O(m \log n)$, where m is the number of edges and n is the number of nodes\cite{dM}. For large graphs, Dijkstra's algorithm becomes very computationally expensive as it iterates through all the nodes in the graph. It is easy to notice that, in many applications, it might be unnecessary to compute the optimal paths to all nodes in the graph, but to only one goal node. This is what A* addresses. 

A* is an algorithm based on Dijkstra's and published in 1968~\cite{aSt}. It utilizes a \emph{heuristic} that helps focusing the direction of the search directly towards the goal. The difference between Dijkstra's algorithm and A* is that in A*, search ends as soon as $u_{goal}$ is popped out of the priority queue and that the \emph{key} of a node $u$ (its priority in the queue. Priority is given to nodes with lower key value) is computed by adding $g(u)$ to a heuristic $h(u, u_{goal})$. The algorithm is shown below for reference.

\begin{algorithm}[h]
\DontPrintSemicolon
\ForAll{$u \in V$}{
$g(u)\leftarrow \inf$\\
}
$g(u_{start})\leftarrow 0$\\
insert $u_{start}$ into $Open$\\
\Repeat{$u=u_{goal}$ \textbf{\emph{or}} $Open=\emptyset$}{
$u\leftarrow$ element from $Open$ with minimal $(g(u)+h(u))$\\
Remove $u$ from $Open$\\
\ForAll{$v \in Succ(u)$}{
\If{$g(v) > g(u)+c(u,v)$}{
$g(v) \leftarrow g(u)+c(u,v)$\\
$bp(v)\leftarrow u$\\
insert $v$ to $Open$ or update it if it is in $Open$\\
}
}
}
\caption{A*\label{aStar}}
\end{algorithm}

The heuristic is a value\footnote{It can be a function that takes a node and returns a real number, or a value for each node saved in a Lock-Up table.} associated with each node and is an estimation of the cost from this node to the goal. It defines the characteristics and affects the performance and results of the A* search. A heuristic is said to be admissible if it is a lower bound of the cost of moving from $u$ to the goal. More formally: 
\begin{equation}
h(u, u_{goal})\leq c^*(u, u_{goal})\quad\forall u \in V
\end{equation}
where $c^*(v, u_{goal})$ is the cost of the optimal path from $v$ to the goal. If the heuristic satisfies this, A* is guaranteed to find the optimal path to $u_{goal}$. An admissible heuristic for a graph embedded in the plane, for instance, where edge costs indicate Euclidean distances between the nodes, is the direct Euclidean distance to the goal (any other path to the goal will be longer). A tighter condition on heuristic is consistency, which is satisfied when:
\begin{equation}\label{consistency}
h(v, u_{goal}) - h(u, u_{goal})\leq c(u, v)\quad\forall u \in V\quad and\quad\forall v\in Succ(u)
\end{equation}
where $Succ(u)$ is the set of successors of $u$ (the nodes where there are directed edges from $u$ to them). 

For two heuristic functions $h_1$ and $h_2$, $h_1$ is said to be more informed if $h_1(u, u_{goal}) > h_2(u, u_{goal})$ for all nodes u. The least informed, yet admissible heuristic is $h(u, u_{goal}) = 0$ for all $u$. This heuristic makes A* equivalent to Dijkstra's algorithm. The better informed a heuristic is, the better the performance of  A* as less nodes are expanded. A perfect heuristic, that is, it represents the actual cost of arriving to the goal, would allow A* to reach the goal directly without expanding any nodes outside the optimal path.

Several notes are worth mentioning about A*. It is important to realize that initializing the $g$-values of all nodes at the start (lines 1-3) is not necessary. Instead, a node can be initialized when it is encountered for the first time in the search. This change can be significant when the graph contains a large number of nodes making initializing all of them an expensive task while only a subset of them will be considered. Further, it can be seen that \emph{expanding} states (executing lines 9-15 of the algorithm) is where the major computation time is spent. In other words, the time spent by the search algorithm can be looked at as linear function of the number of expanded nodes. It is interesting to know that if the heuristic is consistent, preventing any node from being expanded more than once would still lead to an optimal result~\cite{ai}. Finally, A* can as well work with more than one goal with nearly no addition to the algorithm. In this case we let $u_{goal}$ donate the set of goal nodes rather than one goal node. The heuristic function is then the minimum value of the heuristic between $u$ and each node in $u_{goal}$ and the search terminates whenever one of the goal nodes is at the top of $Open$.

A* is, without any doubt, a big improvement over Dijkstra's algorithm. However, there is still a large room for further developments. For example, the algorithm always computes the optimal path to the goal, which can still be very expensive in applications where the configuration space is very large or highly multidimensional. In these cases, one might prefer a suboptimal path computed in a short time over an optimal path taking much longer to be ready. Also, whenever a change happens in the graph, the search has to be carried out again from scratch throwing away all the efforts done before. This is unhandy and inefficient, especially in environments that are continuously changing by nature as in the case of autonomous cars. The following subsections review the evolution of heuristic-based search to account for all these situations. 

\subsection{Backward A*}\label{backA}
A useful note about A* is that it can be performed \emph{backwards}. This means that the search starts at the goal\footnote{Again this can be easily extended to multiple goals.} until finding the start. To do this, every $u_{start}$ occurrence in the algorithm shown above is replaced by $u_{goal}$ and vice versa. The used heuristic $h(u_{start},u)$ should be a lower bound of the cost of moving from the start to the node $u$. In other words, it should be \emph{backward admissible and consistent}:
\begin{multline}
h(v, u)\leq c(v, u)\quad\text{and}\quad h(u_{start},u)\leq c^*(u_{start}, v) + h(v, u) \\
\forall u \in V\quad \text{and}\quad \forall v \in Pred(u)
\end{multline}
Now, $g(u)$ represents the cost of the best found path from $u$ to the goal and when expanding a node $u$, its predecessors are examined instead of its successors (line 9). The found path can then be traced by back pointers (or perhaps \emph{front} pointers) starting from $u_{start}$.

\begin{figure}[htb]
\centering
\begin{subfigure}{.49\textwidth}
  \centering
  \includegraphics[width=.48\textwidth]{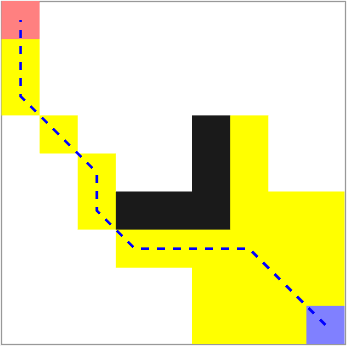}\hfill
  \includegraphics[width=.48\textwidth]{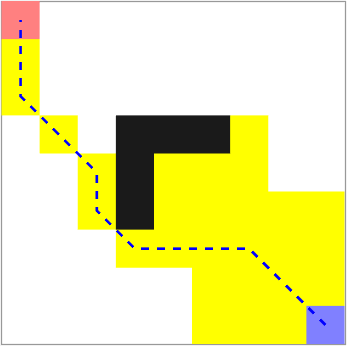}
  \caption{Forward A*}
\end{subfigure}
\hfill
\begin{subfigure}{.49\textwidth}
  \centering
  \includegraphics[width=.48\textwidth]{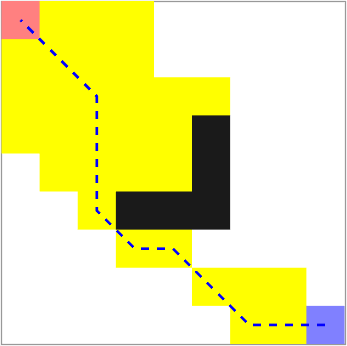}\hfill
  \includegraphics[width=.48\textwidth]{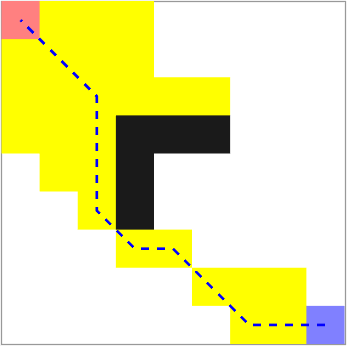}
  \caption{Backward A*}
\end{subfigure}
\caption[The effect of bent obstacles on forward and backward A*]{The effect of concaved obstacles on forward and backward A*.\\ blue: start, red: goal, black: obstacles, yellow: explored area, dashed line: path. Note that paths produced by both forward and backward A* are equivalent in total cost (distance).}
\label{fig:bent}
\end{figure}

While this produces the same result giving the optimal path, the behavior of the search is different. One difference is that forward search spends more time and expands more nodes around obstacles bent towards the start, while backward search struggle more with obstacles bent towards the goal as in Fig.~\ref{fig:bent}. Also, it was found that, when moving in a completely unknown terrain, it can be much more efficient to plan forward rather than backward~\cite{hgu}. This is because in such cases, the observed area is the surrounding of the start, while the edge costs near the goal are not known and typically given optimistic costs (less than the actual). Planning forward allows the agent to efficiently explore the observed area and rapidly progress planning towards the goal directly. In the opposite case, however, the search starts at the unknown area moving soon to the observed area where it encounters costlier edges and begin expanding a large area trying to find a cheaper path. 

However, backward search has a very important property that $g(u)$ represents the cost of the best found path from $u$ towards the goal not the start. This is useful because when the agent moves, the start position changes. On the other hand, the position of the goal is fixed, which means that the $g$-values of all the nodes are still valid. Retaining these values after the planning has finished may allow us to use them again when a change happens in the graph to update the plan instead of planning from scratch.

\subsection{Dynamic Replanning (D*)}\label{dstar}
As mentioned above, in real world scenarios, the agent would initially have an inaccurate graph to plan on. The graph gets continuously and regularly updated as time passes and as the agent moves and so the planned path may become invalid or suboptimal. It would be then prohibitively expensive to plan from scratch using A* to maintain validity and optimality every time an update happens on the graph, especially in such large and complex configuration spaces with tens of thousands of nodes. Further, the changes might actually not affect the optimality of the planned path or affect it in a very minor way such that the planned path can be easily fixed. In these cases, fixing the path or \emph{replanning} is much more logical than starting planning from scratch. 

Achieving this is the purpose of replanning algorithmst. Among the most well known and widely used algorithms under this category are Focused Dynamic\footnote{Note that `dynamic' here does not mean that it plans in the configuration-time space and accounts for the dynamics of the environment. Instead, it means that it can update a previously prepared plan to account for changes in the graph, whether this graph represents a configuration or configuration-time space.} A* (D*) (published 1995) and Dynamic A* Lite (D* Lite) (published 2002)~\cite{dSt}. The two algorithms behave equivalently, but the report will focus on D* Lite as it is simpler and considered a bit more efficient~\cite{dSt}. 

D* Lite is an extension of backward A*. It is based on the idea that when an update occurs in the graph, the cost of the shortest path to the goal ($g$-value) of only few nodes is affected. It tries to recalculate this cost only for nodes that were affected by the new update in the graph \emph{and} are relevant for calculating the new optimal path to the goal.\footnote{The costs of nodes to the goal are enough to determine the optimal path as it can be traced from $u_{start}$ to the goal by always transitioning from node $u$ to the successor node $v$ that minimizes $c(u,v)+g(v)$.} This is achieved by some updates on backward A*. Another value, beside $g$-value, is maintained for each node. It is named $rhs$, a one-step lookahead value based on the $g$-value and hence it is better informed than the $g$-value. it always satisfies:
\begin{equation}\label{rhs}
rhs(u) = 
\begin{cases}
0 & \text{if } u=u_{goal},\\
min_{v\in Succ(u)} [c(u,v)+g(v)] & \text{otherwise}.\\
\end{cases}
\end{equation}
A node $u$ is \emph{consistent} if $g(u)=rhs(u)$, otherwise it is overconsistent if $g(u)>rhs(u)$ or underconsistent if $g(u)<rhs(u)$. overconsistent nodes are those with an expectation that the cost of their path to the goal will decrease, while underconsistent nodes are those with an expectation that the cost of their path to the goal will increase. The $Open$ queue always holds the inconsistent nodes (over\slash under-consistent). They are expanded based on the key value:
\begin{multline}
key(u) = [k_1(u), k_2(u)]\\
= [min(g(u), rhs(u))+h(u_{start},u), min(g(u), rhs(u))]
\end{multline}
The key of a node has two elements and the nodes are ordered in a lexicographical way in the queue based on them. This means that the priority is given to nodes with lower values of $k_1$ and ties (i.e. when $k_1$ of a node is equal to that of another node) are broken by selecting the node with the lowest $k_2$. When a node is expanded, it is made consistent if it was overconsistent and the $rhs$-values of all of its predecessors are updated, otherwise it is made overconsistent and the $rhs$-values of itself and its predecessors are recalculated.

When D* Lite runs for the first time ($u_{goal}$ is overconsistent by initialization and all others have $g(u)=rhs(u)=\infty$), it works equivalently to a backward A* search. The search ends when there is no node in the queue with a key less than $u_{start}$. D* Lite then maintains the $Open$ queue as well as the $g$ and $rhs$ values of all the nodes. When any change is detected, $rhs$-values of the affected nodes are calculated, then inconsistent\footnote{If no inconsistent node found, the previously calculated path is still optimal.} nodes are inserted to the $Open$ queue (that might still has nodes from previous searches) and the nodes are expanded and updated until there is no node in the queue with a key less than the start node. The second element of the key that determines the priority in the queue insures that affected nodes lying along the previously planned path are processed in efficient order. So a minimal number of nodes is processed to find the new optimal path.

D* Lite is well suited for the path planning problem in unknown or partially known environments. It works well with moving start position (because it is a backward search), and replans in response to any change that happens in the graph. Some facts are significant about replanning. Replanning is more expensive when changes occur in the nodes closer to the goal. This is because these changes affect the $g$-values of a larger number of nodes. So, D* Lite is best used when the changes happen close to the agent (e.g. through its observation). It is also important to see that as replanning is based on using previous search results, if changes were significant and the graph was considerably altered, planning from scratch might be much more efficient than relying on search results of a nearly different graph. For that, it is common for systems utilizing D* Lite to abort replanning process if a significant change was detected or if a certain time threshold was exceeded and start planning from scratch \cite{hgu}. 

While D* Lite expands considerably less nodes than a normal A* because it uses previous search information, expanding a node in D* is more expensive as it usually involves iterating over the successors of predecessors of the node to update the $rhs$-values of the predecessors. So, for a small environment with no or very few changes happening in the graph, A* might be more efficient than D* Lite, but as the graph gets bigger and changes increase, the merits of D* Lite replanning feature can be seen and appreciated. An optimized version of D* Lite was presented in~\cite{dSt}. It considerably decreases the frequency of calculating $rhs$-values throughout the search making it significantly faster.

\subsection{Anytime Planning (ARA*)}\label{arastar}
All the algorithms shown in this report until now compute the \emph{optimal} path. However, calculating the optimal path might not be achievable under certain time constrains in complex multidimensional workspaces as it requires very large number of nodes to be expanded. A class of search algorithms often named anytime planners addresses this issue. Anytime planners construct an initial suboptimal solution quickly and improve this solution as time and resources allow. Heuristic\hyp based anytime algorithms generally rely on the fact that if an admissible heuristic of A* is inflated by a factor of $k$, the found path is guaranteed to be no more than $k$ times longer than the actual optimal path~\cite{hgu}.

One of the known heuristic\hyp based anytime algorithms is Anytime Repairing A* (ARA*) published in 2003~\cite{ara}. There is a forward version of this algorithm based on A*~\cite{ara} as well as a backward version based on backward A*~\cite{adS, hgu}. The backward version is superior for moving agents as it is able to improve its plan while moving ($u_{start}$ is changing) for the reasons explained in section \ref{backA}.

In backward ARA*, the priority of a node $u$ in the $Open$ queue is determined by the key:
\begin{equation} \label{keyara}
key(u) = g(u)+\epsilon\cdot h(u_{start},u)
\end{equation}
where $\epsilon\geq 1$ is the inflation factor. The algorithm starts by performing an inflated backward A* with an initial inflation factor $\epsilon_0$. This search is done very quickly and efficiently for one significant fact: each node in the $Open$ queue is only expanded once. This is different from normal A* search where a node is naturally expanded many times depending on its connectivity with neighbor nodes. In ARA*, once a node has been expanded in a particular search, then if it was found, during the search, inconsistent (the cost of the shortest path of the goal was found to be actually less or more than $g(u)$) it is inserted into a list named $Incons$. The search ends when the start node is popped out of $Open$ and a suboptimal plan results. After that, if the time and resources allow, while the agent is moving or before moving, the inflation factor $\epsilon$ is decreased by a certain \emph{step}. The nodes in the $Incons$ list are moved to the $Open$ queue (that might still contains nodes from previous searches) and the priorities of all nodes in $Open$ are updated according to~(\ref{keyara}). This continues until $\epsilon=1$ or until there are no more resources.

From search to search, ARA* retains the $Open$ queue and the $Incons$ list as well as the $g$-values of all expanded nodes. The inflation factor $\epsilon$ is decreased gradually each search getting the planned path closer and closer to the optimal path until resources end or until the path is satisfactory. Note that improving the plan (decreasing $\epsilon$) can only happen in a static environment where the graph stays as it is and nothing is moving except the agent on its previously planned path. What makes the ARA* subsequent searches fast is the fact that in each search only nodes that were found inconsistent during the previous search are inserted to $Open$ and that each node in $Open$ is expanded at most once. Yet, with that, ARA* is able to guarantee an optimality bound\footnote{A number $k$ such that the cost of the resultant plan is less than or equal $k$ times the cost of the optimal plan.} for each search done~\cite{ara}.

Anytime planning allows fast planning in time critical situations as well as in complex multidimensional configuration spaces where planning an optimal path would take a very long time. However, it has its limitations too. The most significant one is that it is not suitable in dynamic configuration spaces where nodes and edges can change. Any change will require ARA* to set back $\epsilon = \epsilon_0$, discard all the information of the previous searches and plan from scratch. Another limitation concerns the generality of the algorithm and comes from the requirement to select some initial inflation factor $\epsilon_0$ and a step by which $\epsilon$ is decreased at each subsequent search. These values have a strong influence on the performance and result of the search. Simulation experiments done on the simulation environment presented in this report have shown that these parameters are situation-dependent. The larger $\epsilon$ is the greedier the search through the configuration space is. With large values of $\epsilon$, the algorithm may expand low number of nodes but is then more prone to be temporarily stuck in local minima. It is recommended by the algorithm designers to decrease $\epsilon$ by small steps as large steps may result in expanding too much nodes~\cite{ara}. However, how small this step should be as well as the initial inflation factor are left for the user to decide based on his needs.

\subsection{Anytime Dynamic Replanning (AD*)}
Replanning algorithms allow maintaining an optimal path in the presence of changes, fixing the path instead of reconstructing it from scratch, which made planning in dynamic configuration spaces a lot cheaper. On the other hand anytime algorithms perform suboptimal searches and improve the plans while time and resources are available leading to a highly reactive planning that is flexible under resources constraints. Many real world applications would benefit from the features of both algorithms categories. For instance, autonomous cars operate in very dynamic configuration spaces where replanning is continuously needed. On the same time, plans are required to be produced quickly and efficiently allowing the agent to find valid (even if suboptimal) paths in critical situations and continue to improve its plans while moving and replanning. Under this motivation, the Anytime Dynamic A* (AD*) was developed in 2005~\cite{adS}.

AD* takes from D* Lite its key that has two elements and the lexicographical ordering of the $Open$ queue as well as the $rhs$-value that always satisfies (\ref{rhs}). It takes from ARA* the idea of inflating the heuristic with $\epsilon$ starting from an initial value of $\epsilon_0$ and gradually decreasing it as time allows to improve the path, in addition to the idea of expanding the nodes only once during a search and adding only the nodes that were found inconsistent at the previous search to $Open$ at the start of a subsequent search. As a result, it is able to plan and replan (in case of changes) in an anytime fashion while moving. 

The key by which nodes are ordered in the $Open$ queue shows how AD* is an intersection between D* Lite and ARA*:
\begin{multline}
key(u) = [k_1(u), k_2(u)] =\\
\begin{cases}
[rhs(u)+\epsilon\cdot h(u_{start},u), rhs(u)] & \text{if } g(u)>rhs(u),\\
[g(u)+h(u_{start},u), g(u)] & \text{otherwise}.\\
\end{cases}
\end{multline}
The heuristic is not inflated for underconsistent nodes (the cost of the path going through them is larger than anticipated) entering the queue as an admissible heuristic is needed so that their new costs propagate through their neighbors.

With this, AD* is able to handle changes in both the graph and the inflation factor giving it a great flexibility and capability in dynamic configuration spaces. However, this does not come without complexities and drawbacks. In fact, while AD* has the advantages of both replanning and anytime planning algorithms, it also suffers from the drawbacks in both of them. From the anytime side, the algorithm still requires the user to select suitable value for $\epsilon$ and the decrease step as discussed in section \ref{arastar}. From the replanning side, expanding a node is more expensive than in a normal A* search as seen in section~\ref{dstar}. Also, it is possible, if significant changes happened, that replanning in AD* would be more expensive than planning from scratch. In fact, it was pointed out that this is more so in AD* than D* Lite as AD* needs to reorder the whole $Open$ queue every time $\epsilon$ is changed~\cite{hgu}. Dealing with this issue in AD* is complex as one has three options confronting changes in the graph:
\begin{enumerate}
\item continue decreasing the inflation factor $\epsilon$ by the same usual step and let AD* replan and improve the path on the same time. This is recommended when the changes are minor and insignificant.
\item increase the inflation factor $\epsilon$ to an adequately large value to allow AD* to replan quickly and then improve the plan by gradually decreasing $\epsilon$ as time allows. This is recommended if the changes are significant but replanning is still more affordable than starting from scratch.
\item discard all the previous search information and start planning from scratch. This should be the decision when replanning is taking a long time. 
\end{enumerate}
The problem is to determine how significant the change should be such that planning from scratch is more affordable. This is highly application dependent. One can set a time threshold, for example, such that when it is exceeded in replanning, planning from scratch begins, but then this would waste valuable time and resources. Instead, simulations and experiments can be done to determine the change conditions that would motivate a new fresh planning.

Despite all these complexities, AD* has proven to be very powerful. The algorithm was used for planning in \emph{Boss}, the autonomous car that won the 2006 DAPRA urban driving challenge moving 97 km in a dynamic urban environment and Obeying the California  Driver Handbook~\cite{boss}. An optimization that considerably decreases the cost of expanding a node and extends the algorithm to plan in the configuration-time space was proposed in 2006~\cite{aPa}. The original algorithm as well as the optimized version were implemented to plan in simulated dynamic configuration spaces as seen further below.
 
\section{Test Environment}\label{MATLAB}
\begin{figure}[htb]
\centering
  \includegraphics[width=\textwidth]{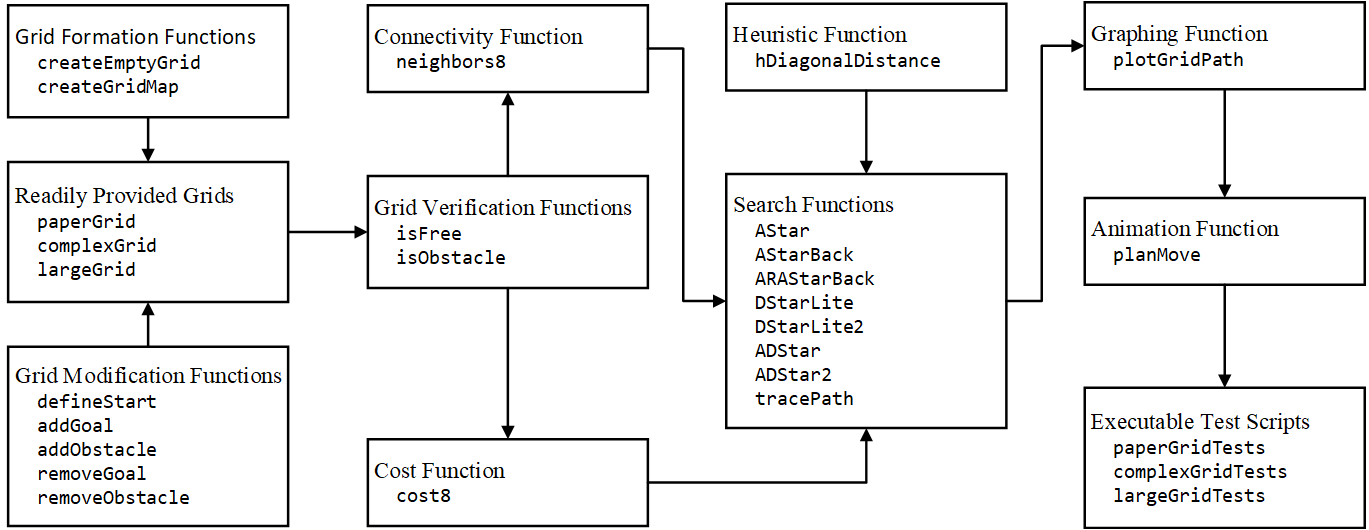}
\caption{A block diagram of the simulation environment}
\label{fig:sysBlock}
\end{figure}

A grid-based simulation environment was designed on MATLAB to test the different shown algorithms. In this section, the backbone of this environment is described. A block diagram of the main environment components is shown in Fig.~\ref{fig:sysBlock}. An arrow from a component $A$ to a component $B$ indicates that $B$ uses $A$.

\subsection{Forming a Grid}
A 2D grid is represented as a MATLAB structure named \textsf{grid}. The structure contains three fields: (1) \textsf{obstacles} which is a binary matrix of the width and height of the grid. A cell in that grid can be either free or occupied by an obstacle. If it is free, its respective value in the \textsf{obstacles} matrix is 0 (or \textsf{false}) otherwise it is 1 (or \textsf{true}). (2) \textsf{start} which is the cell where the initial state or configuration is. A cell (or node) is represented by a horizontal vector containing two numbers representing the $x$ and $y$ coordinates of the cell in the grid respectively. (3) \textsf{goal} which is a $n\times 2$ matrix containing $n$ goal nodes, where $n$ is at least one. This allows to have more than one goal in a grid.

It can be concluded from this that the \textsf{obstacles} field is what gives the grid its general structure and size. It is defined such that if it is given the $x$ and $y$ coordinates of a node as its first and second indices respectively, it determines whether this node has an obstacle on it or not. Hence, it is hard to get a visual image of the grid by simply printing this matrix\footnote{The first index corresponds to a row in MATLAB, so it represents a position in the vertical direction while, in the used definition here, it actually represents the $x$ horizontal coordinate.}. A grid can be generated by calling one of the following two functions:
\begin{enumerate}
\item \textsf{createEmptyGrid}: takes the width and length of the grid and returns a grid with no obstacles (\textsf{grid.obstacles} is all zeros) and with no defined start and goal nodes.
\item \textsf{createGridMap}: takes an obstacle matrix and returns a \textsf{grid} structure with the given obstacle matrix and with no defined start and goal nodes.
\end{enumerate}

Table \ref{gridfun} shows additional functions that operate on the \textsf{grid} structure. All of these functions take two arguments: a \textsf{grid} structure and a node. Verification functions returns a Boolean that is true when the input node is both in the grid and is obstacle\slash free. Modification functions facilitate changes on a node in the grid. These functions do not allow an obstacle to be placed over a goal or a start node and vice versa. However, the start node can be in the same cell as a goal node.\footnote{In which case no need to do any path planning!} While multiple goals can be \emph{added}, only one start node can be \emph{defined}, so when calling the \textsf{defineStart} function, the old start position is replaced by the new one. It is worth noting that these modification functions do not, in fact, change the input \textsf{grid} structure. Instead, they return a new \textsf{grid} structure that is a copy of the input one with the changes applied on it.

\begin{table}[ht]
\centering
\caption{Functions operating on the \textsf{grid} structure.\label{gridfun}}
\begin{tabular}{lll}
\hline
\multicolumn{3}{c}{Grid Verification Functions}\\
\hline
\textsf{isObstacle} & \textsf{isFree}\\
\hline
\multicolumn{3}{c}{Grid Modification Functions}\\
\hline
\textsf{addObstacle}  & \textsf{addGoal} & \textsf{defineStart}\\
\textsf{removeObstacle} & \textsf{removeGoal}\\
\hline
\end{tabular}
\end{table}

\subsection{Search Algorithms and Search Information}
All the algorithms discussed in section \ref{heu} were implemented as functions. Seven search functions are provided, namely: \textsf{AStar} and \textsf{AStarBack} for forward and backward A* search respectively, \textsf{ARAStarBack} for backward ARA* search, \textsf{DStarLite} and \textsf{DStarLite2} for normal and optimized D* Lite search respectively and \textsf{ADStar} and \textsf{ADStar2} for normal and optimized AD* search respectively. All of them were implemented such that they support multiple goals. 

For any path planning algorithm to be compatible with the designed simulation environment, it should be in a form of a function that takes two arguments: a \textsf{grid} structure and another structure named \textsf{searchInfo}, and returns a new version of the \textsf{searchInfo} structure. As its name suggests, this structure holds the search results produced by the path planning algorithm. If the input \textsf{searchInfo} is empty, the function plans from scratch. Otherwise, different procedures are taken based on the different capabilities of the algorithms. Table \ref{sifields} shows the common fields in the \textsf{searchInfo} structure that are present in the output of all the search functions included. In addition to fields shown in table \ref{sifields}, \textsf{searchInfo} may contain additional fields depending on the needs of the search function that uses it. For instance, \textsf{ADStar} and \textsf{ADStar2} output a \textsf{searchInfo} structure that have the additional fields \textsf{rhs}, \textsf{open}, \textsf{incons} and \textsf{eps} (for $\epsilon$).

\begin{table}[ht]
\centering
\caption{Common fields of the \textsf{searchInfo} structure.\label{sifields}}
\begin{tabular}{lp{10cm}}
\hline
Field & Description\\
\hline
\textsf{success} & A binary that is true if a path was found successfully.\\
\textsf{start} & The start of the found plan. This is particularly important when the grid has multiple goals and a backward search is done. This field then determines which goal is in the found plan. Note that, for a backward search, the search is from goals towards the start, so the start of the plan is a goal not the start node. \\ 
\textsf{bp} & A back-pointer matrix. It indicates for each node the next node to be taken to arrive to the goal according to the found plan.\\ 
\textsf{g} & A matrix containing the $g$-value of each node calculated in the search.\\ 
\textsf{expanded} & A binary matrix that is true for nodes that were expanded during the search.\\ 
\textsf{grid} & The \textsf{grid} structure on which the plan was found. This is needed so that when a search function is called again, a comparison between the input \textsf{grid} and the \textsf{grid} field in the input \mbox{\textsf{searchInfo}} can determine if the grid was changed and planning\slash replanning is needed.\\ 
\textsf{time} & The time consumed to complete the search in seconds. Note that this is machine dependent.\\ 
\hline
\end{tabular}
\end{table}

Normally, search algorithms do not output a complete path from $u_{start}$ to the goal. This is because the complete path is not needed, only the next step matters. However, in the simulation environment presented here, computing the complete path was necessary to be able to visualize it. It was important, however, to separate the task of calculating the path from the search functions themselves so that when time measurements are done, they only reflect the effort spent on planning. For that a function named \textsf{tracePath} is included. This function takes a \textsf{searchInfo} structure and returns a copy of it with an additional field named \textsf{path} that is computed in the function. The function simply checks the \textsf{successful} field, if it is 1, it traces the path beginning by the node \textsf{searchInfo.start} and moving to the next one using back pointers \textsf{searchInfo.bp} until it reaches the end of the path. It is able to do this for both forward and backward functions and in both cases the outputted path is a $n\times 2$ matrix containing $n$ nodes, where $n$ is the length of the path, beginning by the start node and ending with a goal node. Note that although most of the implemented algorithms (D* Lite, ARA* and AD*) didn't include back pointers in their published versions, it was possible, yet not very easy, to extend them to produce back pointers.

\subsection{Heuristic and Connectivity Functions}
The search functions need several information to find a valid plan as explained in section \ref{heu}. The needed information includes a heuristic function that takes two nodes and return an admissible estimate of the cost of transitioning from one to the other. In addition to that, the search may need to access the predecessors and\slash or the successors of a node. For that, predecessors and successors functions are required. They take as their arguments a grid and a node and return the predecessors\slash successors nodes of the input node in the input grid. Finally, the edge cost of moving from a node to another (if an edge exists between them) is required.

The simulation environment was designed with generality in mind. So at the start of all search functions, there is a code section named ``options'' where the handles of the four functions (\textsf{h} for the heuristic, \textsf{pred} for predecessors, \textsf{succ} for successors and \textsf{c} for cost) should be given. The functions can be provided by the user according to the purposes of the needed simulation and the whole code will work smoothly without needing to change anything else. 

\begin{figure}[htb]
\centering
\begin{subfigure}{.49\textwidth}
  \centering
  \includegraphics[width=\textwidth]{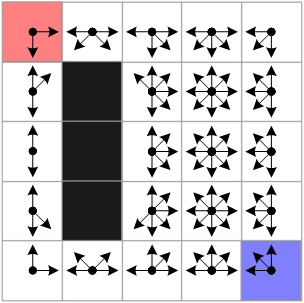}
  \caption{8-connected grid: for every outgoing edge in a cell there is an ingoing edge in the opposite direction.}
  \label{fig:conn1}
\end{subfigure}%
\hfill
\begin{subfigure}{.49\textwidth}
  \centering
  \includegraphics[width=\textwidth]{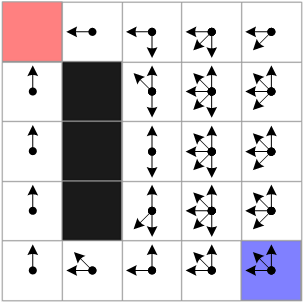}
  \caption{connectivity that doesn't allow moving back in the direction of the start. Generally $Pred(u)\neq Succ(u)$}
  \label{fig:conn2}
\end{subfigure}
\caption[Different connectivity schemes]{Different connectivity schemes. A black arrow represents an edge that allows moving from a node to a neighbor.}
\label{fig:conn}
\end{figure}

For simplicity, the tests presented in this report are all based on an 8-connected grid model (Fig.~\ref{fig:conn1}). In this model, a node is connected to the eight surrounding nodes except obstacle nodes, which are not connected. This implies that the predecessors of a node are the same as its successors. So for both predecessors and successors, one function named \textsf{neighbors8} is used to return the non-obstacle of the surrounding 8 nodes of a node. In addition to that, a simple cost function named \textsf{cost8} that only considers distance is used. If there is no edge between a node and another (as the case with obstacles), the cost of transitioning is infinity. Moving to an adjacent node (straight) costs 1 and moving to a corner node (diagonally) costs $\sqrt{2}$. For the heuristic, the diagonal heuristic \cite{heur} is used in a function named \textsf{hDiagonalDistance}. In the heuristic, the unit straight distance is considered as 1 and the unit diagonal distance is considered as $\sqrt{2}$.

Note that, with the provided functions, the agent is allowed to move backward any time. In addition, it cannot move more than one node a time. These conditions can be easily changed by changing the connectivity functions \textsf{pred} and \textsf{succ}. Suppose, for example, we want the agent to only move forward in the direction of the goal, then \textsf{pred} and \textsf{succ} cannot be linked to the same function. Instead \textsf{pred} should returns neighbors that are in the opposite direction of the goal and \textsf{succ} should return neighbors that are in the direction of the goal as in Fig.~\ref{fig:conn2}. Similarly, the agent can be allowed to move more than one node a time, by connecting to it more surrounding nodes than eight, provided that there is no obstacle between the node and any of the connected nodes. Connecting the surrounding 24 nodes, for example, would allow moving one or two nodes a time in any direction. It is clear that the more connectivity there is in the graph the more expensive the computation needed to find a plan.

\subsection{Graphing and Animation}
The principal graphing function in the presented simulation environment is named \textsf{plotGridPath}. Its purpose is to allow plotting grids, paths, as well as expanded cells in the grid during the search. For that it takes a group of inputs. The inputs include a \textsf{grid} and \textsf{searchInfo} structures. This allows plotting the grid and the found path from the search as well as the expanded nodes during the search. If the input \textsf{searchInfo} is empty, the function only plots the grid, and if the input \textsf{grid} is empty while \textsf{searchInfo} is not, the graph updates the path on a readily available plot. For that, a handle to the figure where the grid was plotted before has to be inputted. In addition to that, handles to specific plotted objects are among the inputs, namely: handles to the plotted start cell, expanded cells and path line. These inputs allow updating the plotted figure with minimal computational efforts instead of redrawing the whole figure again and again. If a plot is being produced for the first time, the input handles can be replaced by empty inputs. The function will then create a new plot from scratch. 

In the top of the function file, there is a code section named ``options'' where the user can control various options regarding the produced plots. This includes controlling whether specific elements appear or not like expanded cells, path line, grid lines and coordinates ticks. The options also allow changing the colors of all elements of the figure and enabling or disabling a \emph{color map} for the expanded cells. When this color map option is enabled, the expanded cells are plotted with different colors based on their g-value. The color is yellow for low g-values close to zero (expanded cell was found by the search to be very close to the search start), it moves gradually to cyan as the g-value increase (meaning that the cell is considered far from the search start). This creates more informative figures giving the user an idea on why the found path was chosen by a specific search algorithm. The effect of some of these options can be seen in the figures included in this report. For instance, the plots in Fig.~\ref{fig:bent} are generated with grid lines and color map disabled, while plots in Fig.~\ref{fig:paperGrid} has grid lines and color map enabled, and Fig.~\ref{fig:complexGrid} and \ref{fig:largGrid} have grid line disabled and color map enabled. In all of these figures, coordinates ticks option was disabled.

After its execution, \textsf{plotGridPath} returns handles to the plotted figure as well as objects inside it (again, start cell, expanded cells and path line). These outputs can then be used as inputs to the same function in future calls which would allow creating fast animations. Suppose, for example, we have a grid that has not been plotted before, and its associated search information from a search function that operated on the grid. We can call \textsf{plotGridPath} with these two inputs and empty inputs for the handles. The function would produce a plot for the grid and the path on it and return handles to the created figure and the individual items inside it. Suppose then we want to animate movement of the agent along the planned path, the function can be called repeatedly with the output handles of each call as the input handles of the following call and with updated start position in input grid and updated path in the input \textsf{searchInfo}. The same principle can be used to animate the updated paths and plans while the agent is moving in case of anytime planning or replanning. 
 
To make animation easier and more accessible to the user, an animation function named \textsf{planMove} is also included in the simulation environment. It takes a \textsf{grid} and a \textsf{searchInfo} structures as well as all the handles taken by \textsf{plotGridPath} and a handle to a search function. If there is no plan, \textsf{planMove} uses the search function to plan a path and animate movement of the agent along the planned path for one step using \textsf{plotGridPath}. It then outputs all the handles outputted by \textsf{plotGridPath} as well as an updated version of the inputted \textsf{grid} (where the start position is changed to the node the agent moved to) and \textsf{searchInfo} (where the path is reduced by one node because the agent already executed part of the path). Using these outputs as inputs to the same function in a future call (naturally in a loop) will animate another one step movement and so on until the agent reaches the goal. 

It is also possible if a plan was already found in an initial call of \textsf{planMove} to input a grid where some obstacles have been changed (using \textsf{addObstacle} and \textsf{removeObstacle} functions) to represent the situation of the grid after the one step movement. \textsf{planMove} will then animate moving the agent one step along the old path. Then plot the new situation of the grid and use the search function to update the plan, or plan from scratch if it is not capable of replanning, and show the new plan on the figure. With this, \textsf{planMove} is able to animate moving the agent, changes in the grid and the subsequent replanning as well as anytime planning. Because updating the figure can be too fast depending on the size of the grid and the time needed to do the search, the function includes an ``options'' section to change the pause time between updates to make the animation easier to follow.

In addition to the outputs mentioned above, \textsf{planMove} also outputs the number of expanded cells and the time taken to do the search if a search was done. This allows the function to be used for time tests too. To make the tests faster and more accurate, the function doesn't show any animation if the input figure handle is empty, which allow doing fast and accurate time tests by using \textsf{planMove} to move the agent along the path and compute the total time taken by all planning\slash replanning. This can be done repeatedly then the average of this total time can be taken as a performance measure.

\section{Results and Analysis}
Several tests have been performed on the simulation environment. The tests started on a simple grid so that it is easy to inspect and debug when there is a problem in a search algorithm. This grid is the same $7\times 6$ grid that can be seen in~\cite{adS, hgu}. The idea was to regenerate the same results shown in these papers to ensure the correct functioning of the whole system. The grid can be generated in the environment by calling a function named \textsf{paperGrid}. 

Another larger and more complex grid was designed and can be generated in the environment by calling a function named \textsf{complexGrid}. This grid has two goal nodes. Its size gives a better opportunity to test the replanting and the effect of different rates and inflation factors for anytime planning.

Finally a third huge grid is included in the environment and can be generated by calling \textsf{largeGrid}. This grid was initially generated by a maze generator. However, the generated maze had very narrow passages only one cell wide, so it was extensively modified to have wider passages and multiple paths to the different places in it. Because the grid is very large, its \textsf{obstacles} matrix is saved in a \textsf{.txt} file. When \textsf{largeGrid} is called, it loads the matrix, adds a start positions and four different goal positions. The large size of this grid allows more reliable and realistic testing for the search functions. Table~\ref{grids} gives a summary of the properties of the three grids.

\begin{table}[htb]
\centering
\caption{Properties of the grids included in the simulation environment.\label{grids}}
\begin{tabular}{l|ccc}
\hline
Property &\multicolumn{3}{c}{Grids}\\
\hline
 & \textsf{paperGrid} & \textsf{coomplexGrid} & \textsf{largeGrid}\\
Dimensions (verti. $\times$ horiz.) & $7\times 6$ & $15\times 26$ & $99\times 55$\\
Number of cells   & 42 & 390 & 5445 \\
Number of goals   & 1 & 2 & 4\\
Start position $(x,y)$ & $(1,6)$ & $(1,1)$ & $(1,1)$\\
Obstacles density & $35.71\%$ & $21.54\%$ & $37.25\%$ \\
Movement scenario changes & 1 & 3 & 7\\
\hline
\end{tabular}
\end{table}

The reasons for including multiple goal positions in some of the test grids is that dealing with multiple goals is particularly important when planning in the configuration-time space (that is to be inspected later), where the search has multiple configuration-time goal nodes usually in the same configuration but at multiple points in time.

For each grid there is an associated test MATLAB \emph{script}. The script contains what we call a movement (or navigation) scenario where the \textsf{planMove} function is used to plan a path and move the agent through it. During movement, the grid is changed and the agent is faced by a new situation where it has to replan or plan from scratch. The number of times each grid is changed during the movement scenario is shown in table~\ref{grids}. The test script iterates through all the seven included search functions performing the test on them. Every test script has two choices: (1)~To disable time tests. In this case, an animation for the movement and planning in the agent is shown to the user to follow and see. This usually useful for what the author names ``Behavior'' tests, where the behavior of the search can be followed and verified. (2)~To enable time tests. The plotting is then disabled and the script moves the agent silently through the grid until it reaches a goal. While doing this, it records the total time spend, the total number of expanded cells in planning, replanning and anytime improvement as well as the length of the followed path to the goal. It does this repeatedly for a number of repetitions inputted by the user depending on the available time and computational power. The tests are done for each search function and the average metrics of them over the repetitions are printed on the screen at the end.

\subsection{Behavior Tests}\label{correct}
The first performed test was, as mentioned before, the regeneration of the search results that was shown in~\cite{adS,hgu} over a small grid. The test script of the \textsf{paperGrid} performs the same simple navigation scenario shown in these papers, where the agent plans, moves for two cells then finds a new information about the environment realizing that there was a falsely placed obstacle in the old graph. So, it plans or replans depending on its capabilities. 

\begin{figure}[htb]
\centering
\begin{subfigure}{.49\textwidth}
  \centering
  \begin{subfigure}{.31\textwidth}
    \centering
    \includegraphics[width=\textwidth]{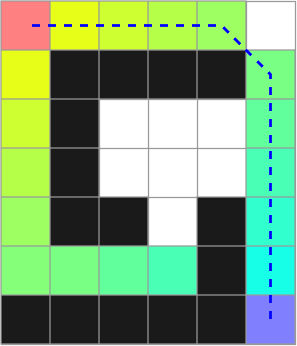}
    \caption*{$\epsilon=1.0$}
  \end{subfigure}
  \begin{subfigure}{.31\textwidth}
    \centering
    \includegraphics[width=\textwidth]{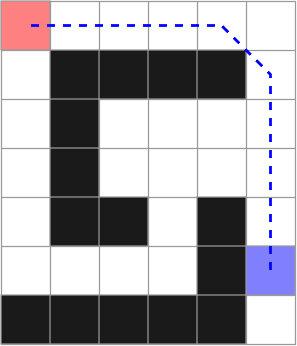}
    \caption*{$\epsilon=1.0$}
  \end{subfigure}
  \begin{subfigure}{.31\textwidth}
    \centering
    \includegraphics[width=\textwidth]{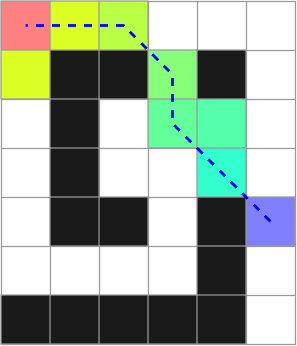}
    \caption*{$\epsilon=1.0$}
  \end{subfigure}
  \caption{Backward A*}
\end{subfigure}
\hfill
\begin{subfigure}{.49\textwidth}
  \centering
  \begin{subfigure}{.31\textwidth}
    \centering
    \includegraphics[width=\textwidth]{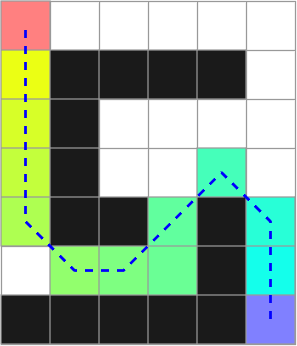}
    \caption*{$\epsilon=2.5$}
  \end{subfigure}
  \begin{subfigure}{.31\textwidth}
    \centering
    \includegraphics[width=\textwidth]{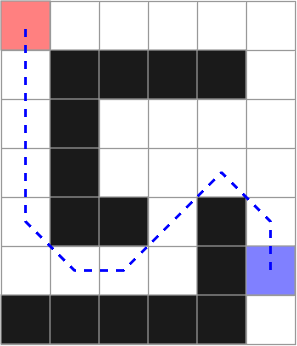}
    \caption*{$\epsilon=2.5$}
  \end{subfigure}
  \begin{subfigure}{.31\textwidth}
    \centering
    \includegraphics[width=\textwidth]{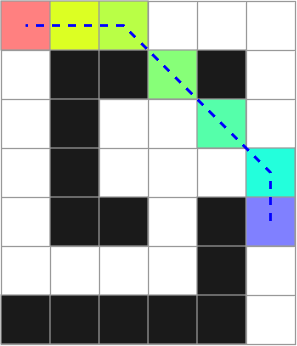}
    \caption*{$\epsilon=2.5$}
  \end{subfigure}
  \caption{Backward A* with $\epsilon=2.5$}
\end{subfigure}
\par\medskip
\begin{subfigure}{.49\textwidth}
  \centering
  \begin{subfigure}{.31\textwidth}
    \centering
    \includegraphics[width=\textwidth]{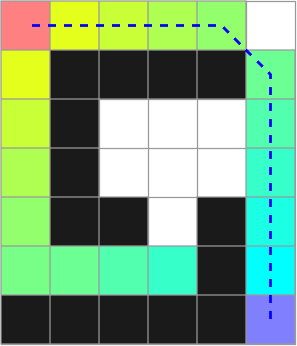}
    \caption*{$\epsilon=1.0$}
  \end{subfigure}
  \begin{subfigure}{.31\textwidth}
    \centering
    \includegraphics[width=\textwidth]{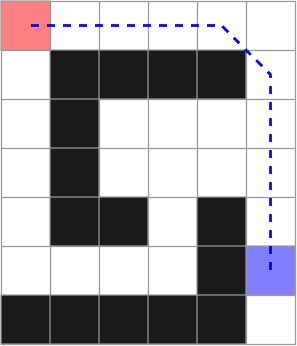}
    \caption*{$\epsilon=1.0$}
  \end{subfigure}
  \begin{subfigure}{.31\textwidth}
    \centering
    \includegraphics[width=\textwidth]{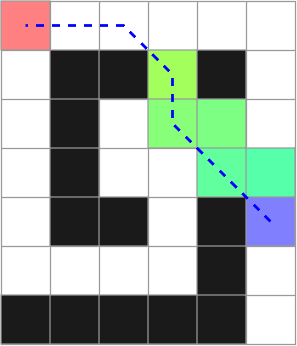}
    \caption*{$\epsilon=1.0$}
  \end{subfigure}
  \caption{D* Lite}
\end{subfigure}
\hfill
\begin{subfigure}{.49\textwidth}
  \centering
  \begin{subfigure}{.31\textwidth}
    \centering
    \includegraphics[width=\textwidth]{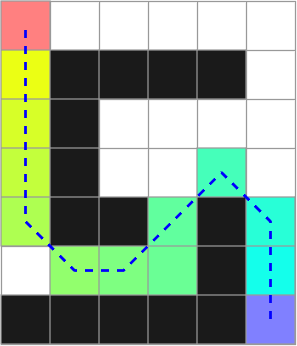}
    \caption*{$\epsilon=2.5$}
  \end{subfigure}
  \begin{subfigure}{.31\textwidth}
    \centering
    \includegraphics[width=\textwidth]{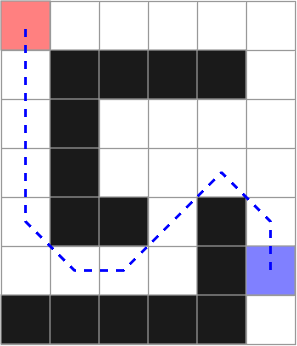}
    \caption*{$\epsilon=2.5$}
  \end{subfigure}
  \begin{subfigure}{.31\textwidth}
    \centering
    \includegraphics[width=\textwidth]{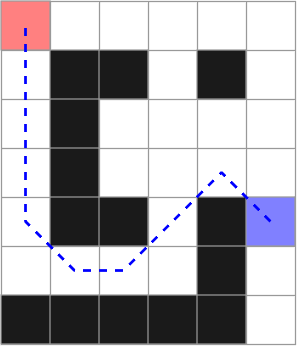}
    \caption*{$\epsilon=2.5$}
  \end{subfigure}
  \caption{D* Lite with $\epsilon=2.5$}
\end{subfigure}
\par\medskip
\begin{subfigure}{.49\textwidth}
  \centering
  \begin{subfigure}{.31\textwidth}
    \centering
    \includegraphics[width=\textwidth]{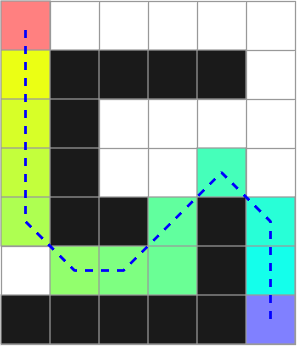}
    \caption*{$\epsilon=2.5$}
  \end{subfigure}
  \begin{subfigure}{.31\textwidth}
    \centering
    \includegraphics[width=\textwidth]{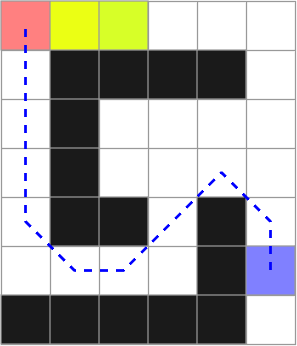}
    \caption*{$\epsilon=1.5$}
  \end{subfigure}
  \begin{subfigure}{.31\textwidth}
    \centering
    \includegraphics[width=\textwidth]{i/plot1c.png}
    \caption*{$\epsilon=1.0$}
  \end{subfigure}
  \caption{ARA*}
\end{subfigure}
\hfill
\begin{subfigure}{.49\textwidth}
  \centering
  \begin{subfigure}{.31\textwidth}
    \centering
    \includegraphics[width=\textwidth]{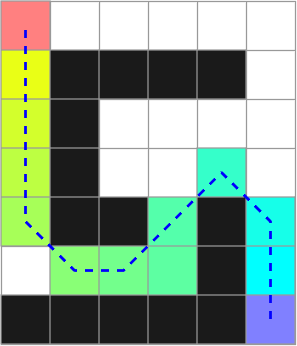}
    \caption*{$\epsilon=2.5$}
  \end{subfigure}
  \begin{subfigure}{.31\textwidth}
    \centering
    \includegraphics[width=\textwidth]{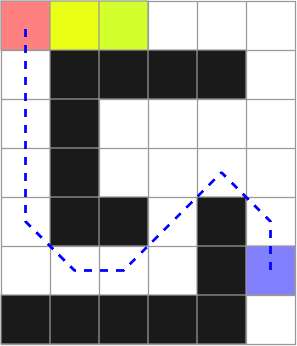}
    \caption*{$\epsilon=1.5$}
  \end{subfigure}
  \begin{subfigure}{.31\textwidth}
    \centering
    \includegraphics[width=\textwidth]{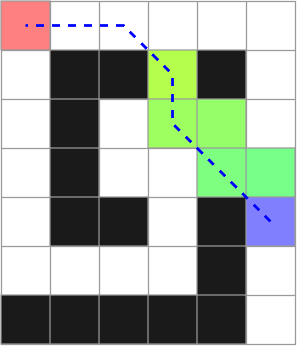}
    \caption*{$\epsilon=1.0$}
  \end{subfigure}
  \caption{AD*}
\end{subfigure}
\caption{Behavior of the search functions on a simple navigation scenario.}
\label{fig:paperGrid}
\end{figure}

The grid is small enough for all the results to be shown. The resultant plans can be seen in Fig.~\ref{fig:paperGrid}. It can be seen that they are very similar to the ones shown in~\cite{adS, hgu}. However, the number of expanded cells of all the search functions is generally slightly less. This is due to the fact that these papers use a different heuristic and cost function. The heuristic they use is the minimum of the horizontal and vertical distance between two nodes and the cost is the same for moving straight or diagonally. While this heuristic is valid and admissible, it creates ties\footnote{Situations where two paths are equivalent according to the provided cost function.} that are broken arbitrarily for visually undesired trajectories sometimes. Because the same discussion shown in~\cite{adS, hgu} applies here, it is not repeated.

Because of the large sizes of the other grids, it is not possible to show the complete behavior of all the search functions over these grids in this brief report. Instead, light will be spotted on interesting and significant details of the different functions in different situations.

\begin{figure}[htb!]
\centering
\begin{subfigure}{.49\textwidth}
  \centering
  \includegraphics[width=\textwidth]{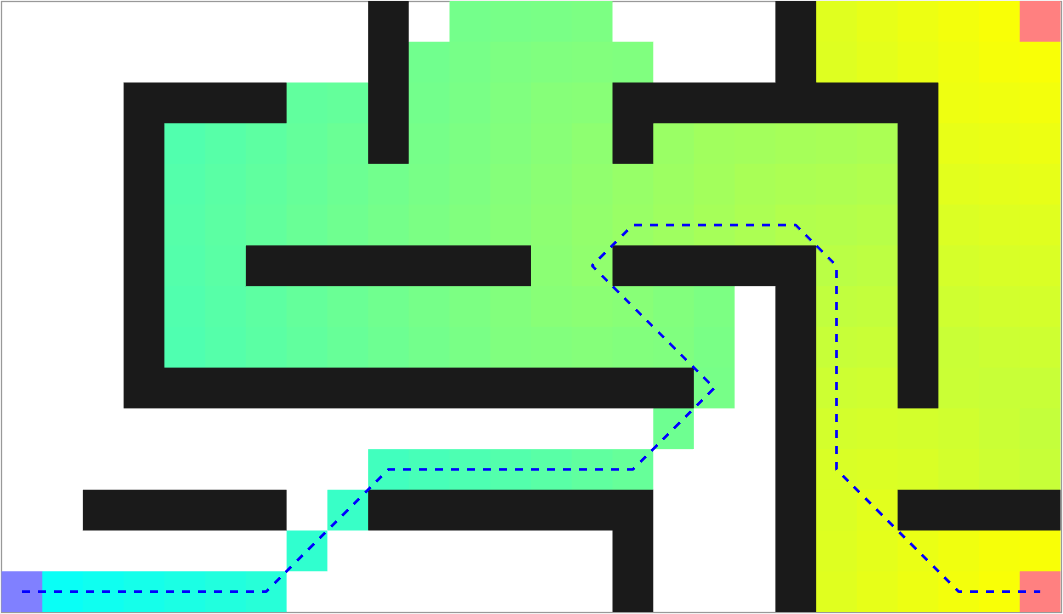}\par\medskip
  \includegraphics[width=\textwidth]{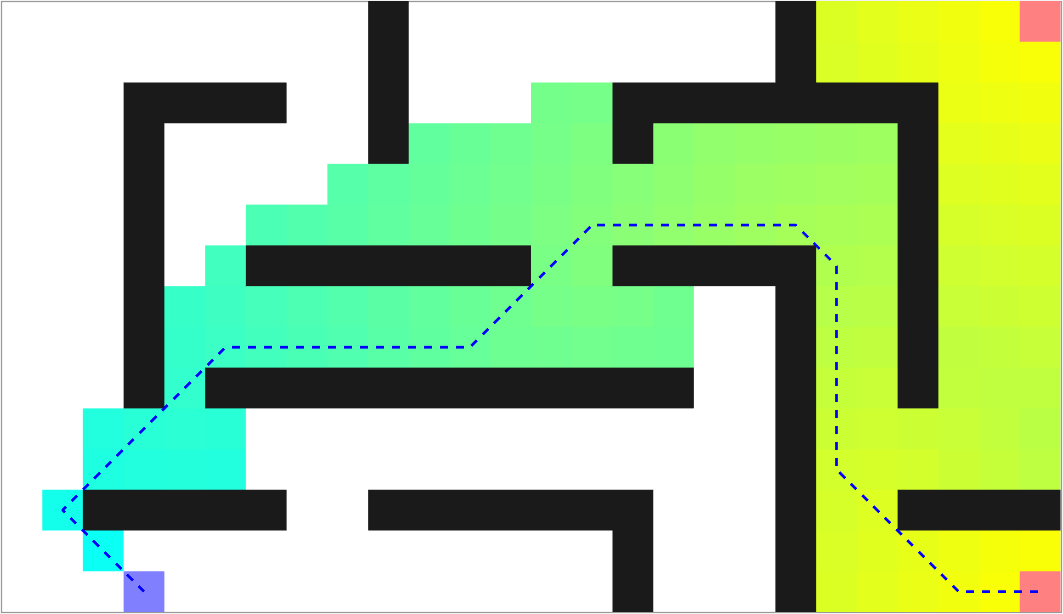}\par\medskip
  \includegraphics[width=\textwidth]{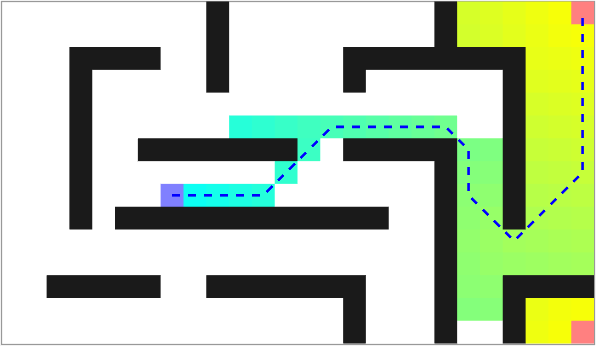}\par\medskip  
  \includegraphics[width=\textwidth]{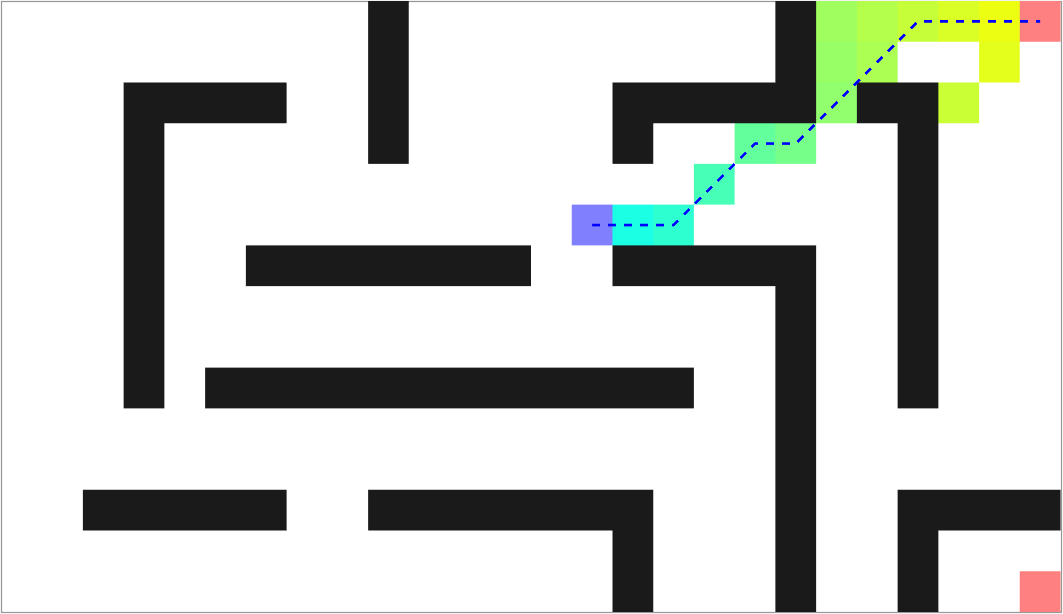}\par\medskip
  \caption{Backward A*}
  \label{fig:comp1}
\end{subfigure}%
\hfill
\begin{subfigure}{.49\textwidth}
  \centering
  \includegraphics[width=\textwidth]{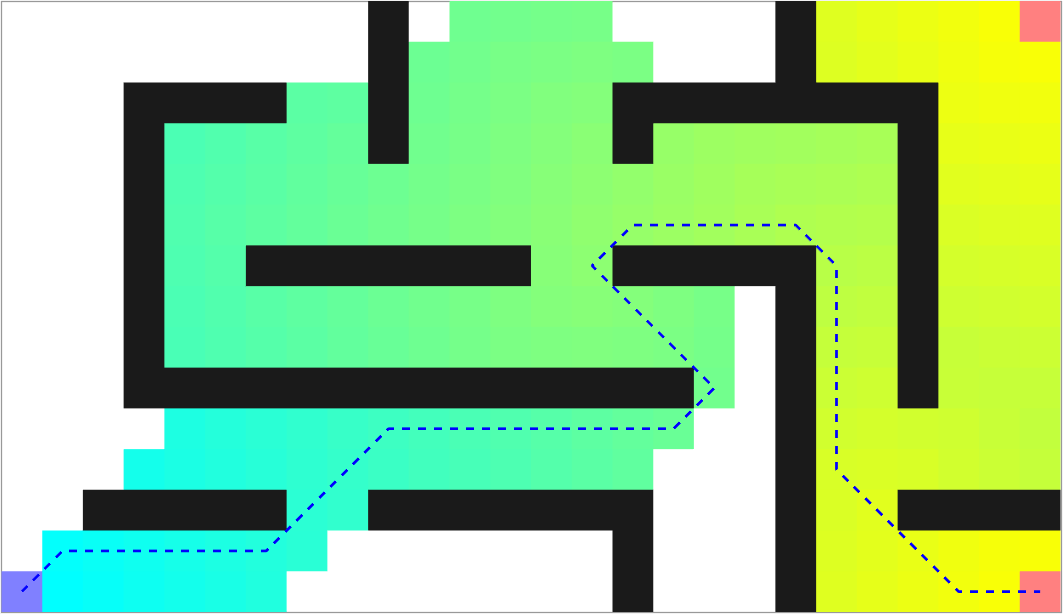}\par\medskip
  \includegraphics[width=\textwidth]{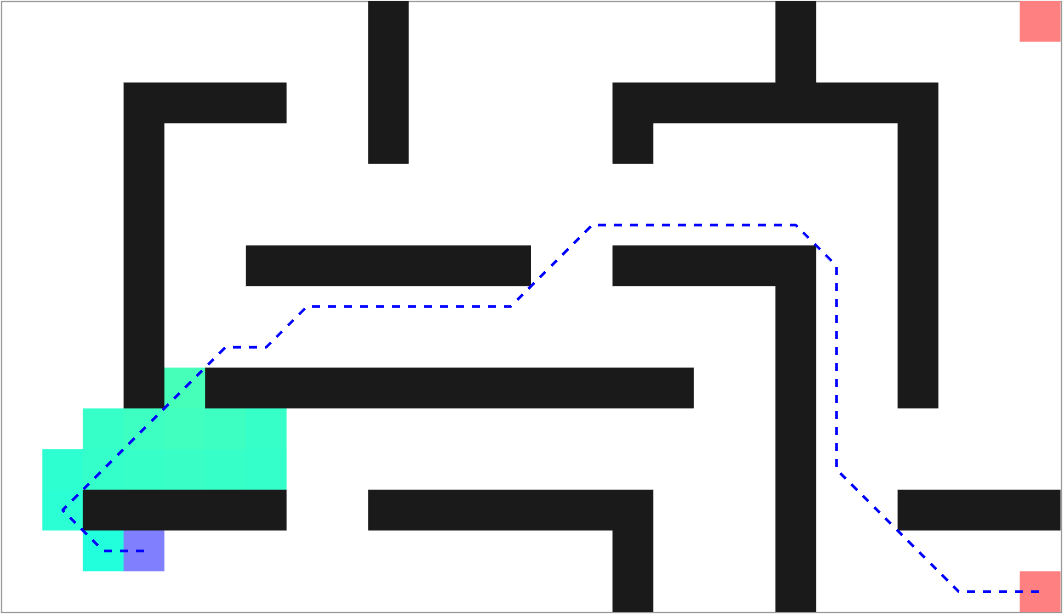}\par\medskip
  \includegraphics[width=\textwidth]{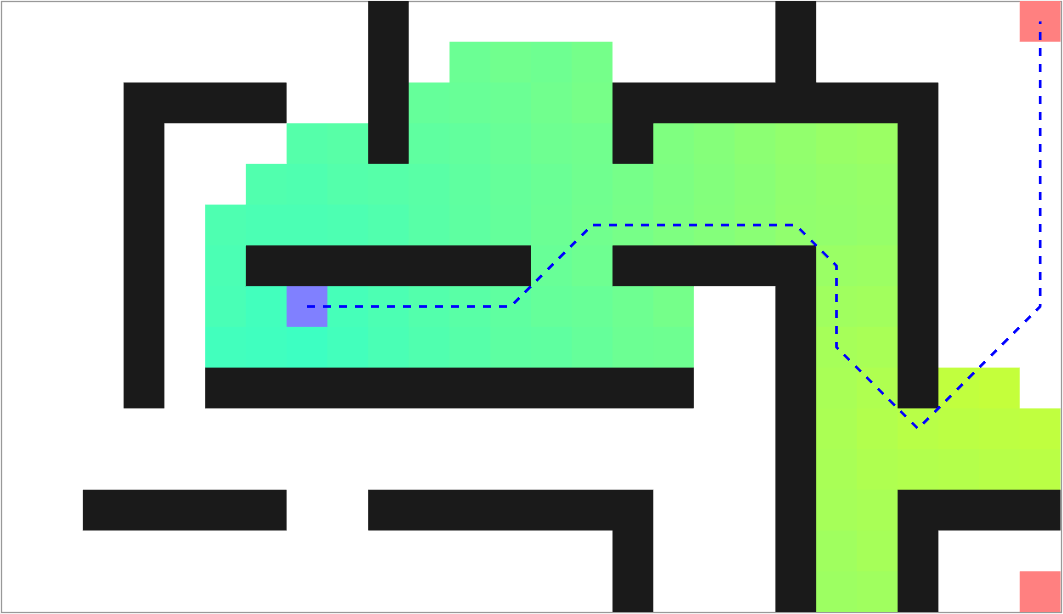}\par\medskip  
  \includegraphics[width=\textwidth]{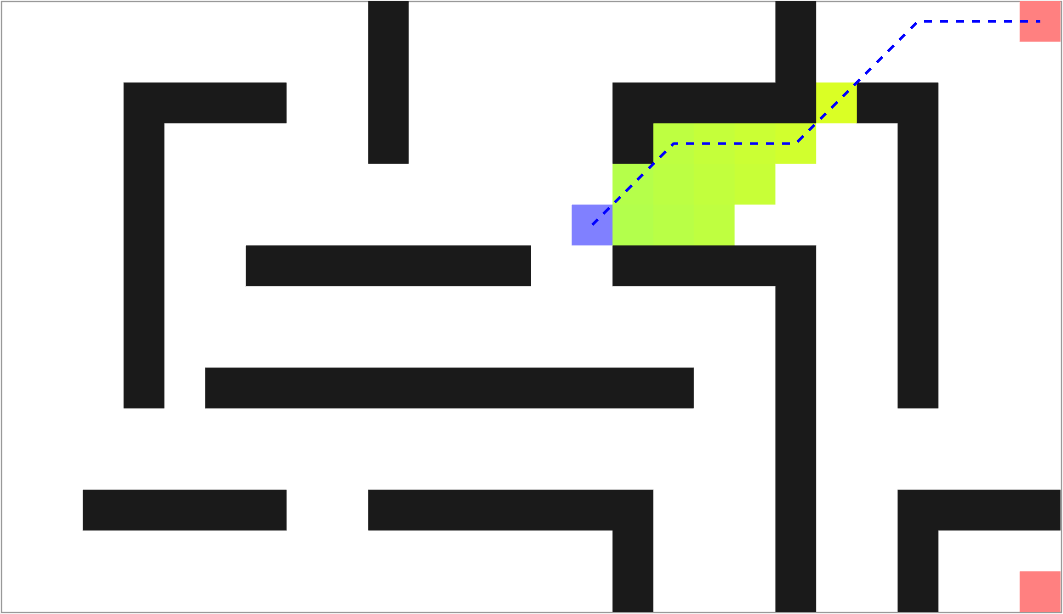}\par\medskip
  \caption{D* Lite}
  \label{fig:comp2}
\end{subfigure}
\caption{Behavior of backward A* compared to D* Lite on the \mbox{\textsf{complexGrid}}.}
\label{fig:complexGrid}
\end{figure}

Figure~\ref{fig:complexGrid} shows the behavior of both backward A* and D* Lite it the \textsf{complexGrid}. Initially (first row of the figure), the two plan optimal paths\footnote{Note that although the two resultant paths are slightly different, they both have the same length and are optimal} using a backward search. It can be noticed, however, that backward A* expanded less nodes than D* Lite in the initial plan. This is due to the different ties braking strategy used by both algorithms. While backward A* breaks ties arbitrarily when there is two nodes with equal keys in the queue, D* Lite uses keys with two elements to break  ties in a more tactical way. The strategy of D* Lite is beneficial in replanning, but in planning from scratch it results in more expanded cells. In fact, the author believes that rethinking how D* Lite should break ties (redesigning the key of the queue) may lead to significant improvements for both D* Lite and AD*.

After the agent moves for 3 cells, it finds that one of the obstacles is removed (second row of Fig.~\ref{fig:complexGrid}). Here, the difference between backward A* and D* Lite is seen clearly. D* Lite was able to quickly find the new optimal path expanding a small amount of nodes, while backward A* had to plan from scratch. After moving again, the agent discovers that the goal it was headed to is inaccessible (third row of Fig.~\ref{fig:complexGrid}). It can be seen that, in that case, planning from scratch was almost equivalent to replanning, again due to the way ties are broken. Finally, while approaching the second goal, the agent realizes that there is a shorter path to it (last row of Fig.~\ref{fig:complexGrid}). Backward A* plans from scratch, expanding states from the goal until the agent position. D* Lite replans. In short, while D* Lite spends a big effort in initial plans, it manages to save a lot when facing a change in the environment in most of the cases. The total number of expanded cells and the spent time over all the grid are shown in table~\ref{metrics}.

\begin{figure}[htb!]
\centering
\begin{subfigure}{\textwidth}
  \centering
  \includegraphics[width=.49\textwidth]{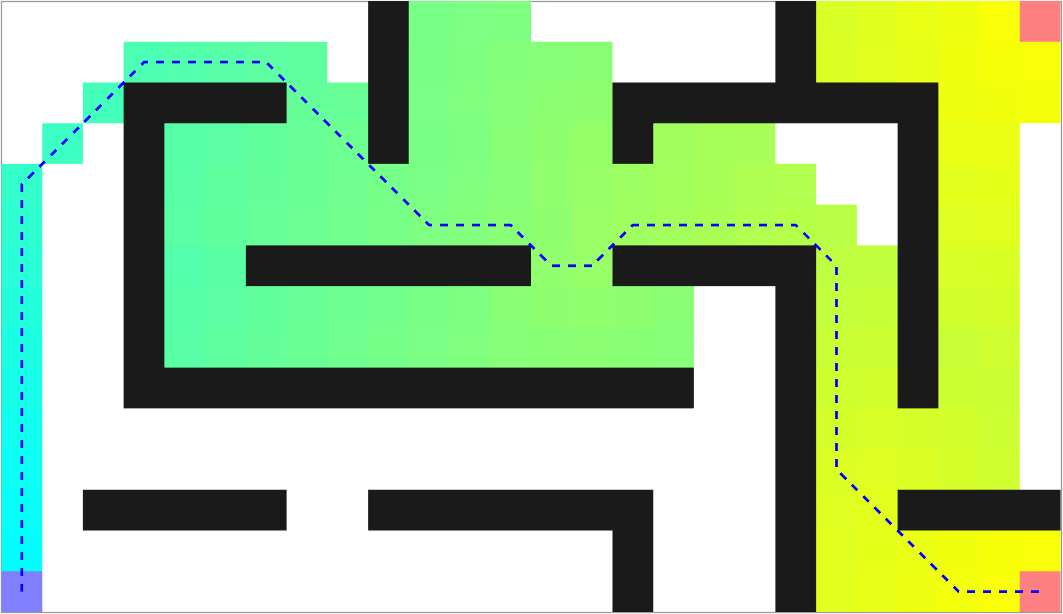}\hfill
  \includegraphics[width=.49\textwidth]{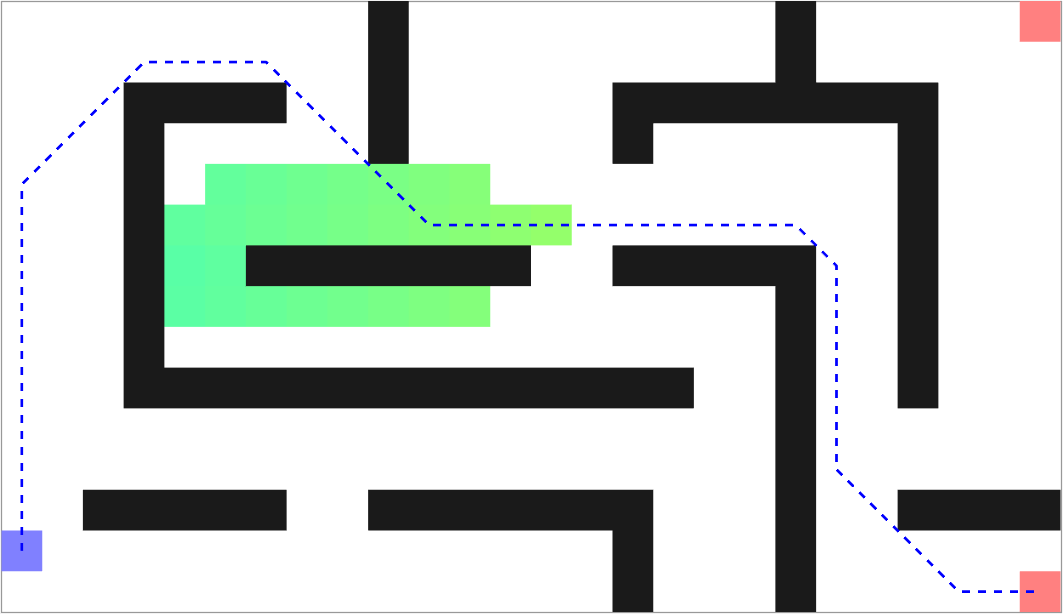}
\end{subfigure}%
\par\medskip
\begin{subfigure}{\textwidth}
  \centering
  \includegraphics[width=.49\textwidth]{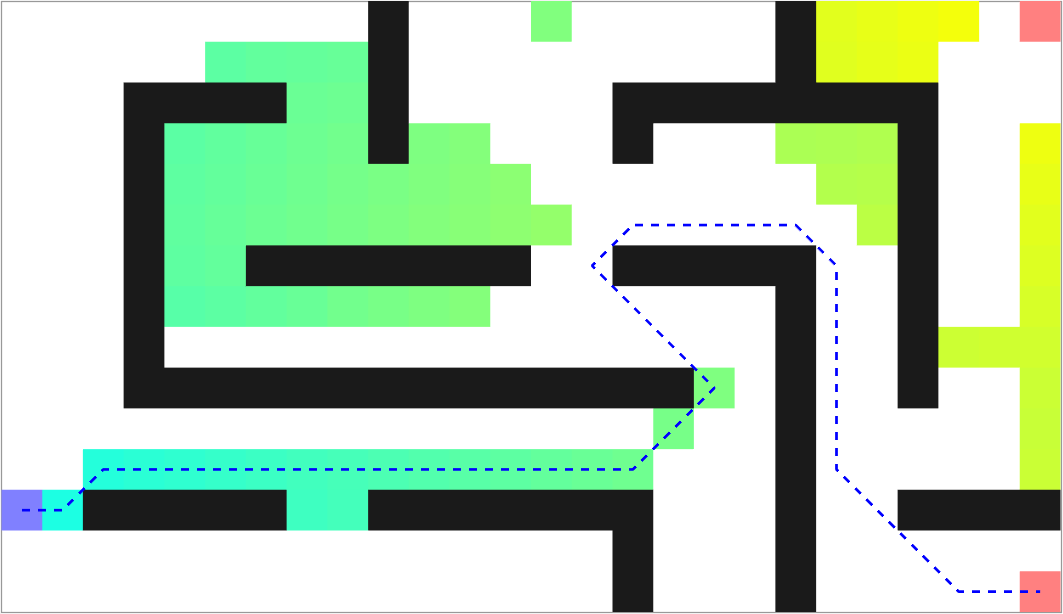}\hfill
  \includegraphics[width=.49\textwidth]{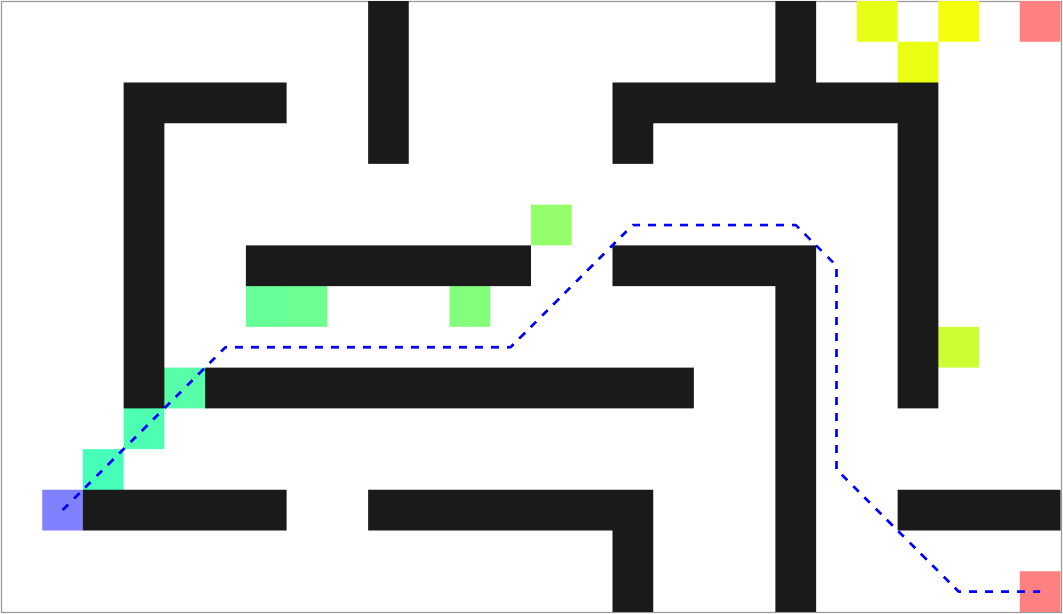}
\end{subfigure}
\caption{Planning and replanning of AD* on the \textsf{complexGrid}.}
\label{fig:complexGridAD}
\end{figure}

An example of how AD* works is shown in Fig.~\ref{fig:complexGridAD}. The agent starts planning on the \textsf{complexGrid} (upper left of the figure), expanding as few nodes as possible at the expense of the optimality of the found path. At the subsequent step (upper right), the inflation factor $\epsilon$ is decreased and nodes are expanded trying to find a better path. AD* succeeds to find the optimal path after the second step (bottom left) as $\epsilon$ was decreased more. After three steps (bottom right), the grid is changed and AD* is able to replan and find the new optimal path expanding a minimal number of nodes. An interesting case happens here. The initially suboptimal path allowed the agent to be closer to the shortcut road that appears in the grid after the third step (compare this to the second row of fig~\ref{fig:complexGrid}). This made the final total path taken by AD* from start to goal actually shorter than the one taken by methods that find the optimal path from the beginning like A* and D* Lite. In addition to that AD* expands less nodes to find this shorter path as seen in table~\ref{metrics}.

Finally, Fig.~\ref{fig:largGrid} compares the behavior of D* Lite, ARA* and AD* at different stages on the \textsf{largeGrid}. It can be seen in Fig.~\ref{fig:largD} (top) that D* Lite expands most of the grid (2824 nodes) to find an initial optimal path of length 143 nodes. After moving for some time, the agent is faced with a new situation (middle) where the old path is not valid anymore. D* Lite expands only 13 nodes to find the new optimal path of length 124. Later, A new more significant change\footnote{Note that other changes that required modifying the plan have occurred in the grid previously in instances that are not shown in the figure because of the large number of changes that are programmed to occur in the \textsf{largeGrid}.} is detected (bottom), so D* Lite replans again expanding 307 nodes to find an optimal path of length 95.

\begin{figure}[htb]
\centering
\begin{subfigure}{\textwidth}
  \centering
  \includegraphics[height=.3\textheight]{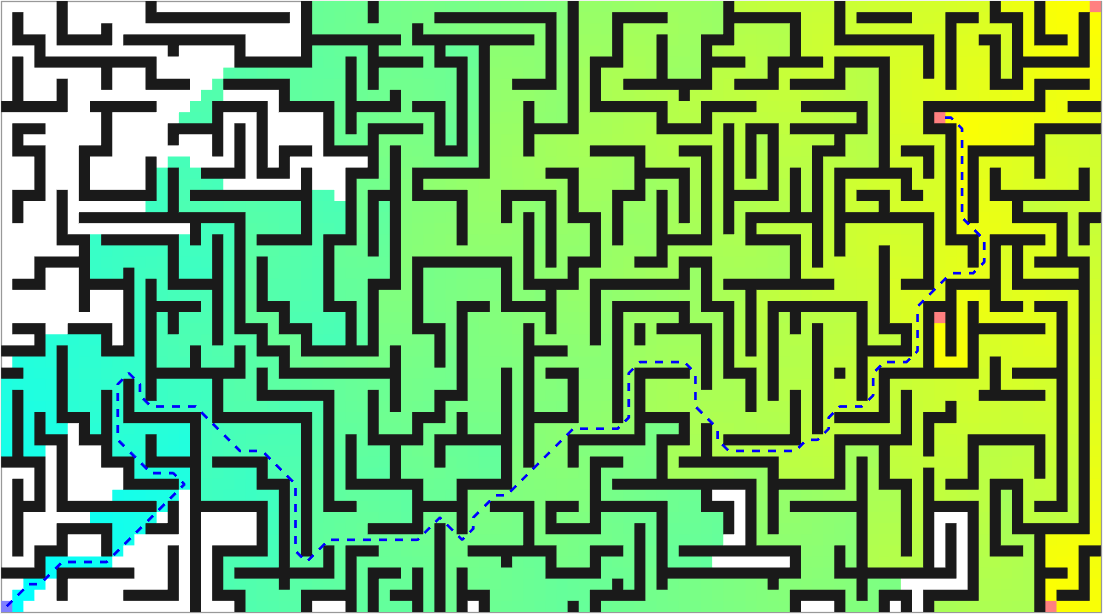}\par\medskip
  \includegraphics[height=.3\textheight]{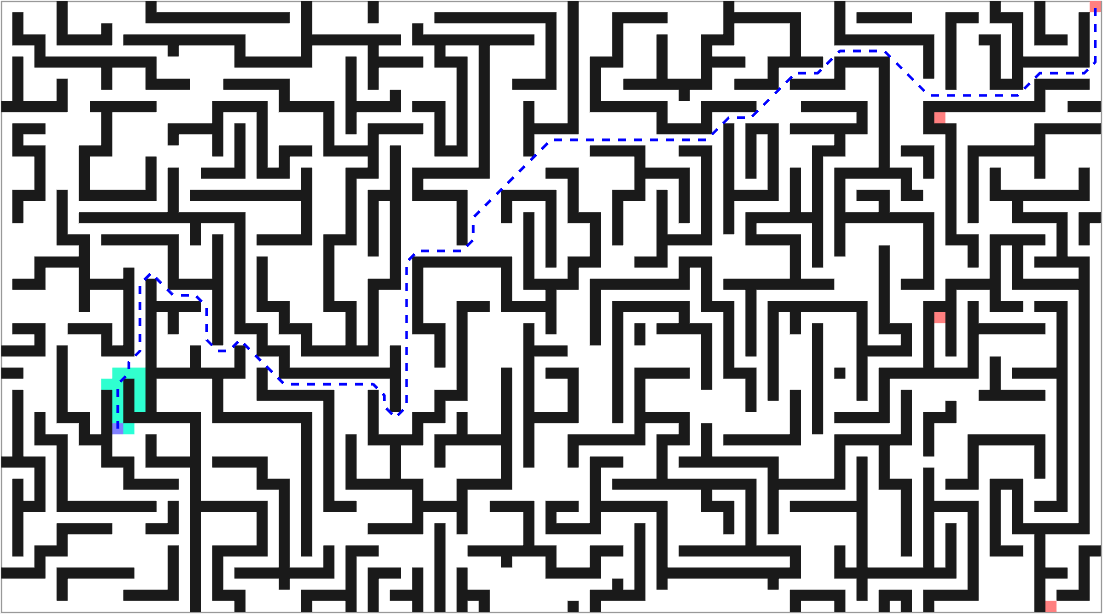}\par\medskip
  \includegraphics[height=.3\textheight]{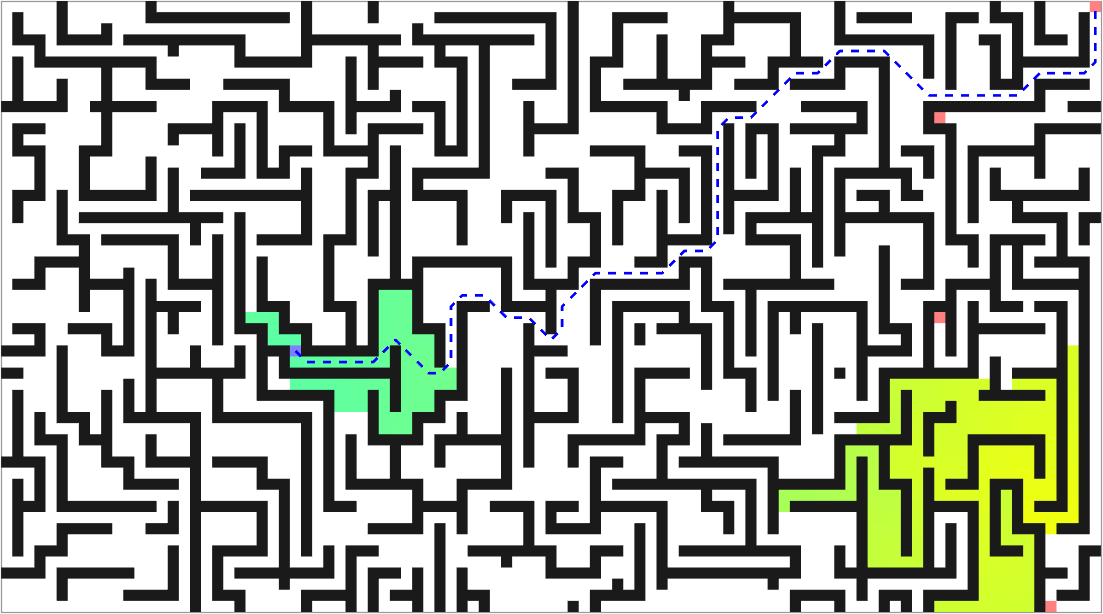}
  \caption{Optimized D* Lite}
  \label{fig:largD}
\end{subfigure}
\caption{Behavior of D* Lite, ARA* and AD* on the large grid.}
\label{fig:largGrid}
\end{figure}

Figure~\ref{fig:largARA} shows how ARA* deals with the same situation. Initially (top), the algorithm finds a suboptimal path using an inflation factor $\epsilon=4.5$ expanding only 359 cells. The found path is surprisingly optimal with a length of 143. However, the user and the algorithm have no guarantee on that. So as the agent moves, ARA* expands more cells trying to find shorter paths. These intermediate searches are not shown in the figure. When it reaches some point, it finds that the path is not valid (middle). So, the whole previous search results are discarded. The inflation factor is reset to 4.5 again and a quick new plan is prepared expanding 326 cells to find a path of length 136. The same thing happens again at a later point (bottom) and again ARA* discards all the previous efforts, reset $\epsilon$ and plan from scratch expanding 240 nodes to find a path of length 105. Note that the found paths in the last two situations are not optimal and are different than the ones D* Lite found. 

As the agent is moving, ARA* will continue trying to improve the found path. While ARA* is able to converge to the optimal path after some amount of time, the continuous changes in the grid makes ARA* loses its optimality frequently. The total path followed by ARA* from start to goal is highly suboptimal compared to other methods as seen in table~\ref{metrics}. One way to fix this is to stop resetting the inflation factor $\epsilon$ whenever a change happens and keep decreasing it as usual. While this will result in more optimal paths, it will gradually convert ARA* to be backward A* as $\epsilon$ gets closer to 1, which means that ARA* will expand much more nodes when a change happens in the grid as seen in table~\ref{admetrics}. Stopping resetting $\epsilon$ caused expanding more than four times the number of nodes expanded when $\epsilon$ was reset at each change in the grid.

\begin{figure}[htb]\ContinuedFloat
\centering
\begin{subfigure}{\textwidth}
  \centering
  \includegraphics[height=.3\textheight]{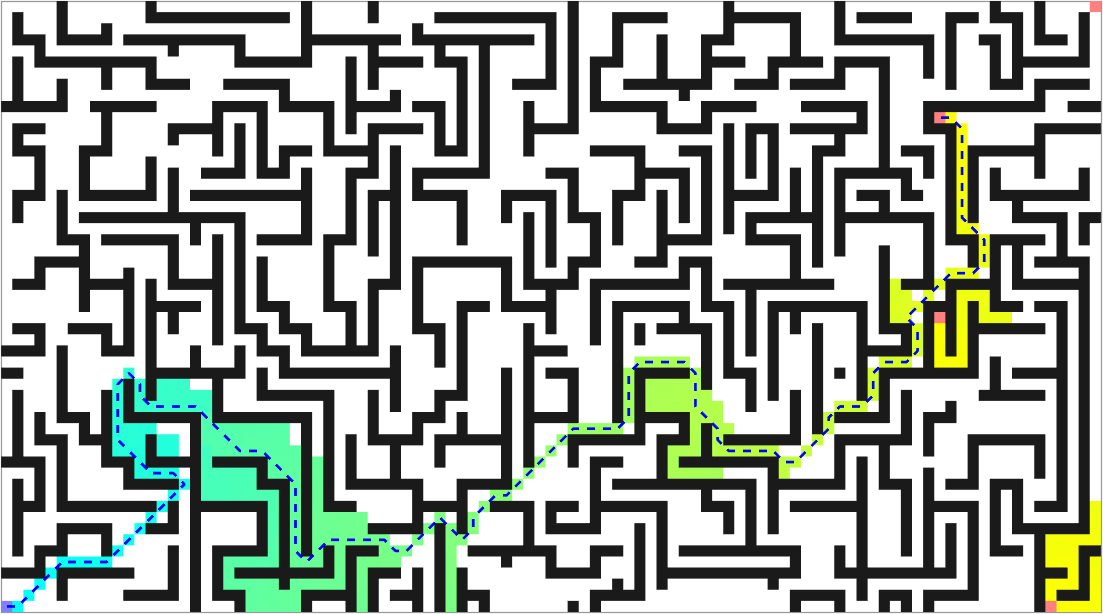}\par\medskip
  \includegraphics[height=.3\textheight]{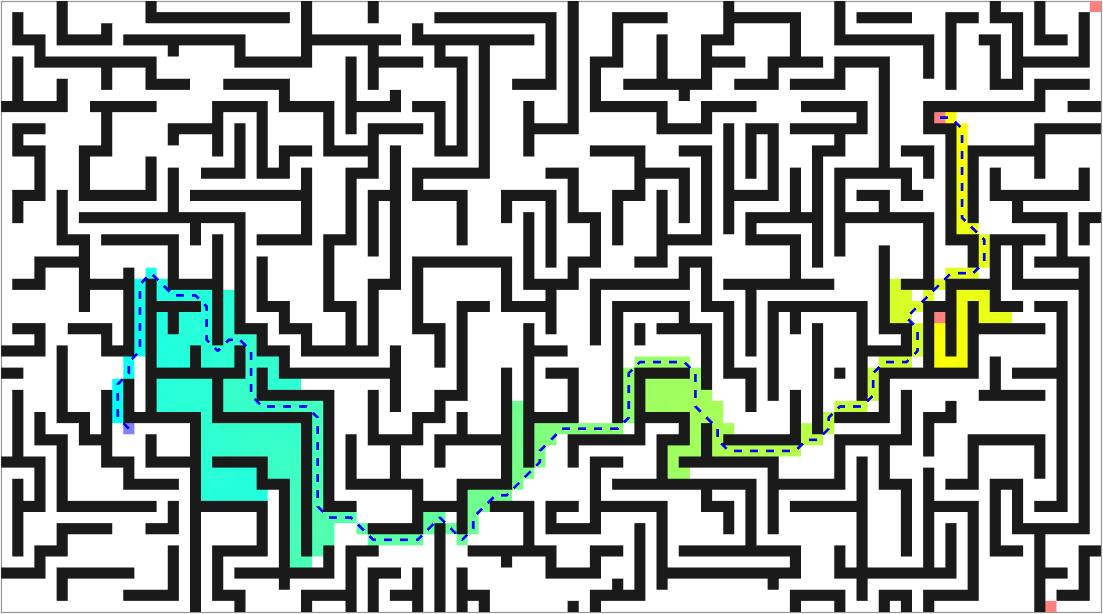}\par\medskip
  \includegraphics[height=.3\textheight]{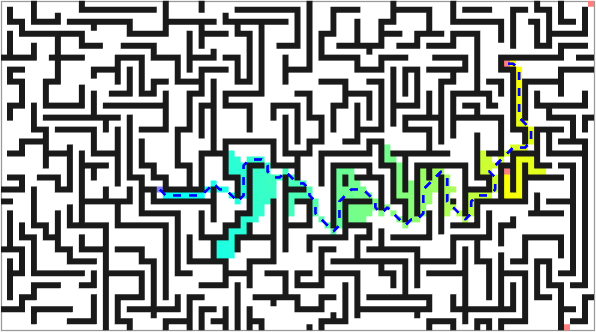}
  \caption{ARA*, $\epsilon_0=4.5$, step = 0.08.}
  \label{fig:largARA}
\end{subfigure}
\caption{Behavior of D* Lite, ARA* and AD* on the large grid (cont.).}
\end{figure}

In Fig.~\ref{fig:largAD}, we see how AD* deals with this situation. Initially (top) AD* finds a quick plan the same way ARA* did. It begins with an inflation factor of 4.5 and finds a plan by expanding only 359 nodes. Again, the found path happens to be optimal of length 143, but AD* is not aware of that and while it moves, it will expand larger parts of the grid trying to find a shorter path but will not find one. The intermediate searches are not shown in the figure. At some point, the agent finds that the old path is not valid (middle), so replanning is needed. However, AD* doesn't have to start from scratch and doesn't have to reset $\epsilon$.\footnote{Unless the change was very significant, which is not the case here.} At that instance, $\epsilon$ hasn't reached 1 yet, and AD* is doing something between ARA* and D* Lite. It expands 270 nodes (much less than what ARA* did at the same situation but also much more than what D* Lite did) to find a suboptimal path of length 135. Note that the expanded cells are not around the goals but are mostly around the changed area. After sometime, when the agent detects a new change in the grid, AD* replans again. At this moment, AD* has converged to D* Lite as $\epsilon$ reached 1 and it is able to repan and find optimal paths. It expands 301 nodes (slightly less than D* Lite) to repair its path and find the optimal path of length 95, the same one found by D* Lite. It can be seen in table~\ref{metrics} that the total number of expanded cells of D* Lite on \textsf{largeGrid} is much less than that of AD*. However, this comes with a great advantage for AD* that will be discussed in the next section.

\begin{figure}[htb]\ContinuedFloat
\centering
\begin{subfigure}{\textwidth}
  \centering
  \includegraphics[height=.3\textheight]{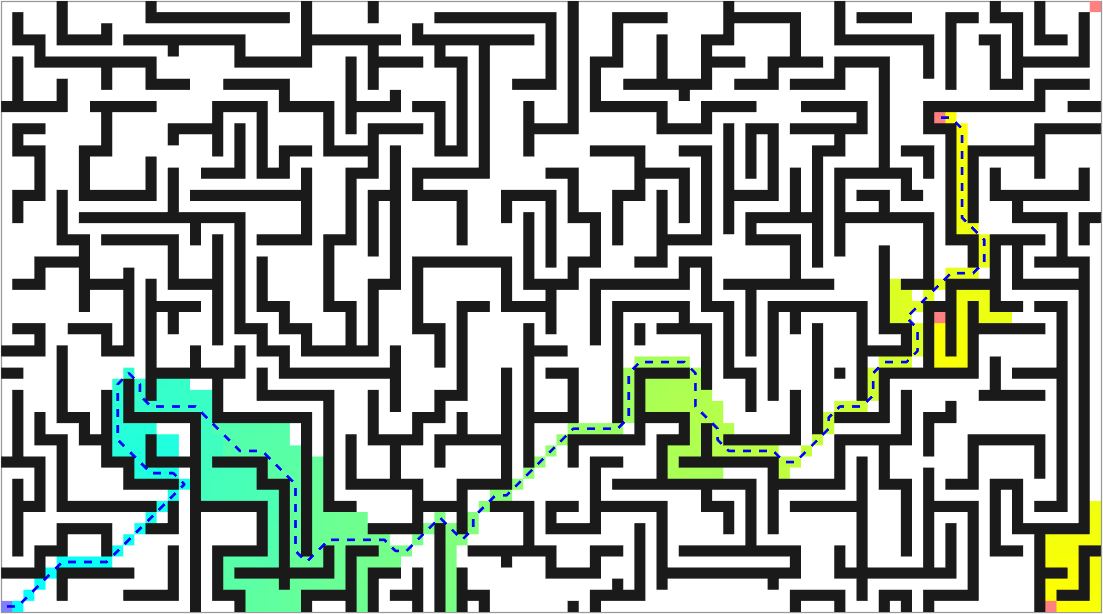}\par\medskip
  \includegraphics[height=.3\textheight]{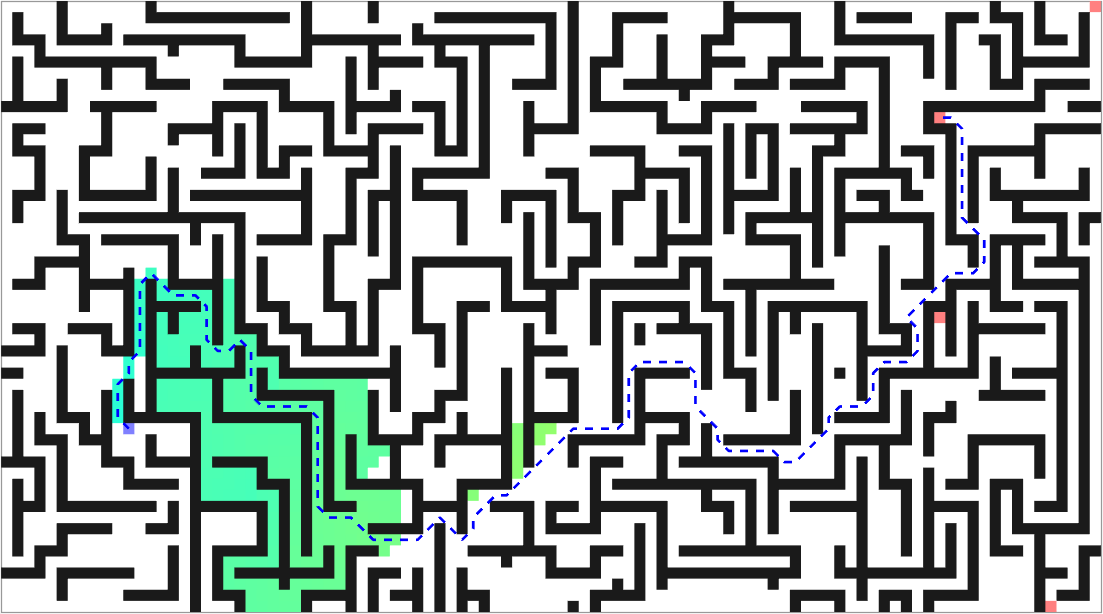}\par\medskip
  \includegraphics[height=.3\textheight]{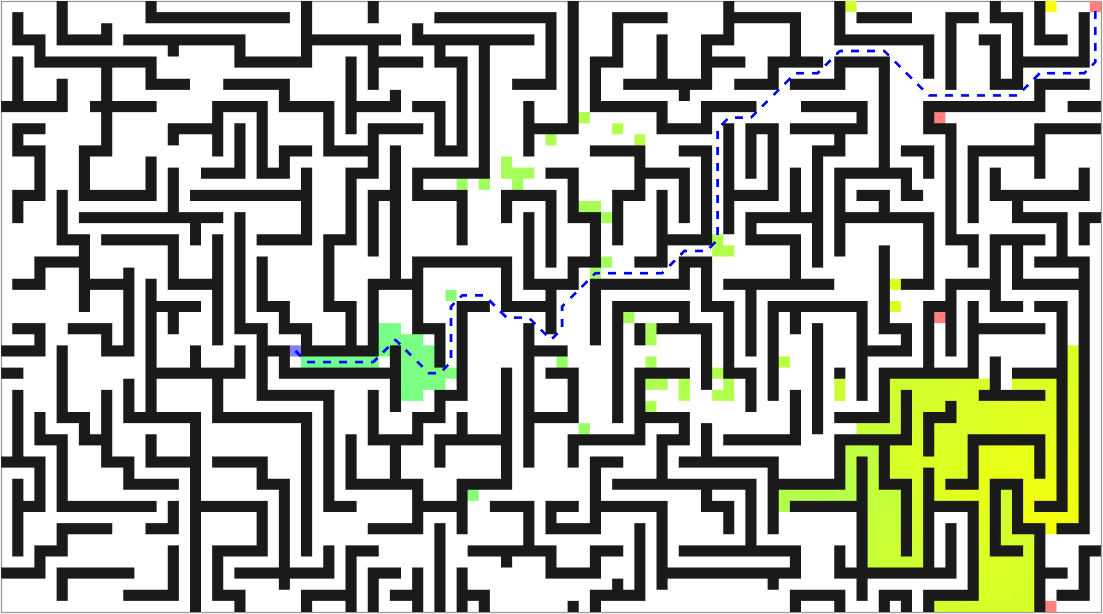}
  \caption{Optimized AD*, $\epsilon_0=4.5$, step = 0.08.}
  \label{fig:largAD}
\end{subfigure}
\caption{Behavior of D* Lite, ARA* and AD* on the large grid (cont.).}
\end{figure}
\clearpage

\subsection{Performance Tests}

Time tests were done in MATLAB to have an indicator of the relative performance of the different search functions on a real system. Table~\ref{metrics} summarizes the results of these tests for the three grids using seven different search functions: A*, backward A* (B.~A*), D* Lite, optimized D* Lite (O.~D* Lite), ARA*, AD* and optimized AD* (O. AD*). The time measurements are taken as averages over a number of repetitions depending on the grid size. The other included metrics are the number of expanded cells and the length of the followed path from start to goal. 

\begin{table}[htbp]
\centering
\caption{Test results for the search functions.\label{metrics}}
\resizebox{\textwidth}{!}{\begin{tabular}{l|ccccccc}
\hline
Metric & A* &  B. A* & D* Lite &  O. D* Lite & ARA* & AD* & O. AD* \\
\hline
\multicolumn{8}{c}{\textsf{paperGrid} (average over 100 repetitions)}\\
\hline
time (s) & 0.0065 & 0.0052 & 0.0196 & 0.0063 & 0.0077 & 0.0240 & 0.0079
\\
expanded & 26 & 26 & 25 & 23 & 21 & 21 & 19\\
path & 8 & 8 & 8 & 8 & 8 & 8 & 8\\
\hline
\multicolumn{8}{c}{\textsf{complexGrid} (average over 50 repetitions)}\\
\hline
time (s) & 0.1721 & 0.0994 & 0.5560 & 0.5113 & 0.1359 & 0.5900 & 0.2016\\
expanded & 734 & 439 & 354 & 349 & 357 & 243 & 245\\
path & 29 & 29 & 29 & 29 & 28 & 28 & 28\\
\hline
\multicolumn{8}{c}{\textsf{largeGrid} (average over 10 repetitions)}\\
\hline
time (s) &  3.4615 & 2.6417 & 3.1217 & 1.2692 & 0.9787 & 22.9397 & 9.8100\\
expanded & 15,903 & 13,528 & 3,357 & 3,349  & 2,265 & 10,462 & 10,658\\
path & 159 & 159 & 159 & 159 & 183 & 159 & 159\\
\hline
\end{tabular}}
\end{table}

The time taken in the tiny \textsf{paperGrid} is not very indicative of the performance of the functions due to the very low number of nodes but is put for completeness. The search steps and the resultant paths over the \textsf{paperGrid} were also shown in Fig.~\ref{fig:paperGrid}. It can be seen that the length of the followed path is the same for all the functions. The number of expanded cells of the different search function on this small grid will not be discussed here as it was discussed before in~\cite{adS, hgu} and is not a good representative of the performance as in the case of larger grids.

It was possible to repeat the tests on the \textsf{complexGrid} 50 times and get the average of these tests for the time measurements. It can be seen that A* expands more nodes than any other method. It expands more than backward A* due to the nature of the obstacles as discussed in section~\ref{backA}. However, the cost of expanding a node in both methods is equivalent at about $230\mu s$ per node. D* Lite expands less nodes than both A* and backward A*, but the cost of expanding one node is much higher than in A*, so the total time is longer. The optimized D* Lite slightly decreases the number of expanded cells and also the time spent per one node, however, the significance of the optimization is not very clear in this grid as it is in the \textsf{largeGrid}. The total path followed by these four method is 29 cells long. ARA* expands slightly more nodes than D* Lite but still much lower than forward and backward A*. The cost of expanding one node is higher than that of A* reaching $380\mu s$ per node, though still considerably lower than that of D* Lite needing only one fifth of its total time. AD* expands nodes less than any other method, but the cost of expanding one node is higher than all other methods, the optimization succeeds to significantly decrease this cost and hence the total time needed by more than 65\%. The length of the followed path by the three last functions is shorter than that followed by methods that plan optimally for the reason explained in section~\ref{correct} and seen in Fig.~\ref{fig:complexGridAD}.

It appears from the tests over the \textsf{complexGrid} that the benefits of methods like D* Lite and AD* can only be seen in situations where the number of nodes they expand is much less than the number of nodes expanded by other methods, which should be the case in the \textsf{largeGrid}. Indeed, forward and backward A* expand a huge number of nodes reaching 15903 and 13528 expanded nodes for the two respectively. Again, forward A* expands more because of the nature of the gird that slightly favors backward searches. D* Lite expands about one fifth of the nodes expanded by A* but because of the higher cost of expanding nodes, the total time of the two methods is close. The optimized version of D* Lite tremendously decreases the total time taken by the algorithm by about 60\% making its total time less than half of that of backward A*. The four methods follow a path of length 159 from start to finish. 

ARA* expands less nodes than all previous methods but follow a highly suboptimal path of length 183 cells. This is because it resets the inflation factor $\epsilon$ at each change in the grid as discussed in section~\ref{correct}. The initial inflation factor $\epsilon_0$ is 4.5 and the step by which this factor is decreased each time is 0.08. This means that 44 steps are needed until $\epsilon=1$ and ARA* is guaranteed to find an optimal path. However, changes are happening frequently such that $\epsilon$ never reaches 1. If $\epsilon$ is not reset at each step, ARA* will soon behave like backward A* and will expand a large number of nodes whenever a change happens. This can be seen in table~\ref{admetrics}, stopping resetting $\epsilon$ made the number of nodes expanded by ARA* close to that of backward A* but allowed ARA* to follow an optimal trajectory. On the other hand, AD* expands a large number of nodes that is comparable to that of backward A*, and because of the high cost of expanding a node, the total time is much higher than all other functions at 23 s. The optimized version decreases this time by more than half reaching less than 10 s but this is still much higher than the other methods, which would make the reader question the whole usefulness of AD*. 

\clearpage
\subsection{Insights on the Performance of AD*}

The performance of AD* depends highly on the values of the initial inflation factor $\epsilon_0$ and the step by which this factor is decreased each time. Different choices of these two parameters lead to different results as shown in table~\ref{admetrics}. So, why choose $\epsilon_0 = 4.5$ and step = 0.08 while this leads to the largest number of expanded cells?

\begin{table}[htbp]
\centering
\caption{Test results for ARA* and Optimized AD* at different sittings on the \textsf{largeGrid}.\label{admetrics}}
\resizebox{\textwidth}{!}{\begin{tabular}{l|cc|ccccc}
\hline
 & \multicolumn{2}{c|}{ARA*} & \multicolumn{5}{|c}{Optimized AD*} \\
\hline
\multicolumn{8}{c}{Settings}\\
\hline
$\epsilon_0$ & 4.5 & 4.5 & 3.0 & 3.5 & 4.5 & 6 & 4.5
\\
$\epsilon$ step & 0.08 & 0.08 & 0.2 & 0.08 & 0.08 & 0.2 & 0.08\\
rest $\epsilon$ & yes & no & no & no & no & no & yes\\
\hline
\multicolumn{8}{c}{Metrics}\\
\hline
time (s) & 0.9787 & 3.3126 & 3.5613 & 9.6405 & 9.8100 & 4.1915 & 5.0884
\\
expanded & 2,265 & 9,331 & 6,162 & 11,844 & 10,658 & 6,789 & 4,374\\
path & 183 & 159 & 159 & 159 & 159 & 159 & 155\\
\hline
\end{tabular}}
\end{table}

\begin{figure}[htb]
\centering
\begin{subfigure}{\textwidth}
  \centering
  \includegraphics[width=\textwidth]{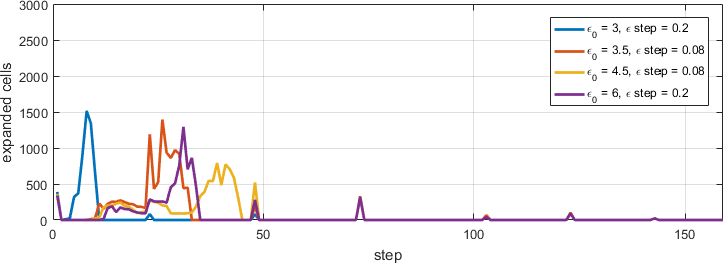}
  \caption{AD* at different values of $\epsilon_0$ and $\epsilon$ step.}
  \label{fig:rates1}
\end{subfigure}
\par\medskip
\begin{subfigure}{\textwidth}
  \centering
  \includegraphics[width=\textwidth]{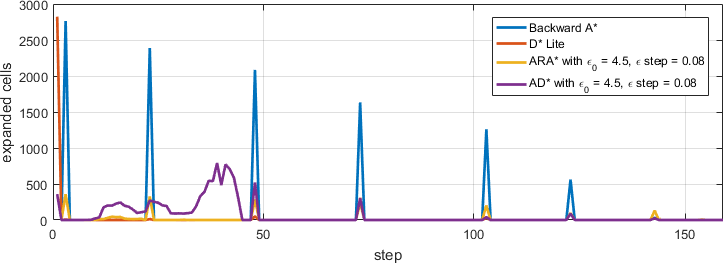}
  \caption{AD* compared to other algorithms}
  \label{fig:rates2}
\end{subfigure}
\caption{The number of nodes expanded at each step as the agent is moving on the \textsf{largeGrid}.}
\label{fig:rates}
\end{figure}

D* Lite does a very good job in replanning, but it spends a very large time to produce its initial plan as was seen in Fig.~\ref{fig:largD}. AD* can be thought of as a way to use D* Lite without needing to spend a long time (that might not be available) on producing an initial plan. For that, the usefulness of AD* is not in decreasing the total number of expanded nodes, but to avoid having to expand a large amount of nodes at one search until AD* converges to the optimal replanning behavior of D* Lite. To better understand this, refer to Fig.~\ref{fig:rates} which shows the number of expanded cells at each step along the path followed by the agent from the start until reaching a goal on the \textsf{largeGrid}. 

AD* begins by behaving like ARA* performing a suboptimal search that expands few number of nodes. The value of $\epsilon_0$ forces a bound on the optimality of the initial plan as explained in section~\ref{arastar}. It can be seen in Fig.~\ref{fig:rates1} that the number of nodes expanded at the first step by AD* with different values of $\epsilon_0$ is almost identical at around 400 nodes. The value of $\epsilon$ step defines how the optimality bound will change from a search to a subsequent one. If it is too large, AD* might need to expand many nodes in one search to satisfy the new optimality level. The ratio between $\epsilon_0-1$ and the step determines the number of searches needed for AD* to converge to D* Lite. The lower this ratio is, the faster for AD* to become D* Lite. 

For the curves shown in the Fig.~\ref{fig:rates1}, the ratio between $\epsilon_0-1$ and the step is lowest for AD* with $\epsilon_0=3.0$ and step = 0.2. With these parameters, AD* converges fast to D* Lite and hence expands a minimal number of nodes as in table~\ref{admetrics}. However, AD* needs to expand a large number of nodes per one search reaching 1500 nodes at the peak. When this step is kept and $\epsilon_0$ is increased to be 6, AD* takes longer to converge to AD* and the total number of expanded nodes is slightly higher than the previous case, but AD* still needs to expand a large number of nodes per one search reaching around 1350. This indicates that the step should be decreased. When the step is decreased to 0.08 and $\epsilon_0$ is set to 3.5, AD* takes less to converge to D* Lite, and at some searches it needs to expand as much as 1400 nodes to improve the path while planning and replanning, which means that the initial inflation factor was not large enough for AD* to be able to expand less nodes per one search. The total number of expanded cells in this situation is almost double that of the previous two situations. If this step is kept and $\epsilon_0$ is increased to 4.5, we get a smoother curve with a peak at 700 nodes per one search. Note that most searches are lower than the peak, so when time is available AD* might do more than one search per step. However, a relatively low upper bound is successfully set on the number of nodes expanded per search. This comes at the expense of the total number of expanded nodes as seen in table~\ref{admetrics}. It can be seen in Fig.~\ref{fig:rates1} also that after AD* converges to D* Lite it behaves the same way whatever was the values of $\epsilon_0$ and the step, replannning only when changes are detected in the grid (small peaks after the 50th step).

To conclude, decreasing the step means that optimality will be increased more slowly, which means that more searches will be done until $\epsilon=1$ and hence more nodes will be expanded. Selecting a suitably small step \emph{and} an adequately large value for $\epsilon_0$ will modify AD* to expand a maximum number of nodes based on the needs and the limitations of the task. However, the appropriate values of these parameters depend on the nature of the graph, which makes the task of selecting them difficult.

Figure~\ref{fig:rates2} shows the number of cells expanded by AD* compared to other methods. Backward A* expands a large number of nodes at the start, then does nothing until a change is detected in the grid, when it plans from scratch expanding also a large number of nodes. The number of expanded nodes in planning decreases as the agent gets closer to the goal because the space of the search gets smaller. D* Lit has a very high peak at the start expanding more than 2800 nodes (which is slightly more than backward A*), then it expands very few nodes (generally less than 500) whenever a change is detected in the grid. Again, the number of expanded cells in replanning decreases as the agent gets closer to the goal. ARA* expands few nodes at the start, then expands fewer or no nodes at the subsequent steps until a change happens where it expands again a few number of nodes. This is because it is never given more than 30 subsequent steps without a change in the grid making it unable to get $\epsilon$ close to 1. Finally, AD* starts by expanding few nodes, then improves the plan expanding more nodes whenever possible. It never expands more than 800 nodes in a search. Around the 50th step, AD* becomes equivalent of D* Lite expanding nearly the same number of nodes for replanning.

\chapter{Planning in Dynamic Environments}\label{chapterDynamic}
In this chapter, the problem of planning in dynamic environments is discussed. For safe planning in dynamic environments, it is inevitable to consider the dynamics of the observed moving obstacles. Planning using static snapshots of the environment as explained in section~\ref{staticProblem}, cannot guarantee safe paths even if the rate of taking these snapshots is high. This example from~\cite{aPa} nicely clarifies the reason behind this:
\begin{quote}
Imagine an agent trying to cross a road on which cars are driving. If the agent was to take a snapshot of the environment with its sensors and assume fixed positions for each object, then it would be in serious risk of getting runover if it attempted to cross the road. In order to successfully accomplish its task, the agent really needs to model the cars as dynamic objects so that it can anticipate where they will be at future times.
\end{quote}

More than that, the interest is in a path planning method that allows applying kinematic and dynamic constraints on the resultant planned paths. For the agent considered in this chapter, simple constraints are required. The required kinematic constraint is that the agent does not change its heading angle from one step in the path to the other by more than a maximum of $\pm\theta$. For many cars, $\theta$ is around $30^o$. The required dynamic constraint is that the agent does not have an acceleration or deceleration higher than $\pm a$.

The aim in this chapter is to find a way to incorporate these constraints in a dynamic planning using incremental heuristic\hyp based search algorithms seen in the previous chapter. These algorithms proved attractive since they return paths with a bound on their optimality. Moreover, they allow more flexibility in setting preferences for the planned path through the cost function.

This chapter begins by discussing a suitable representation of the state of the agent to allow applying these constraints in path planning. The problems in existent dynamic incremental heuristic\hyp based path planning techniques are clarified. Then, the details of a new proposed methods are explained. This is followed by a brief description of a simulation environment written in MATLAB to test the proposed algorithm. The chapter is ended by an analysis of the simulation results.

\section{Dynamic Constraints in Incremental Search}
As discussed before in section~\ref{dynamicProblem}, A popular way of dynamic planning using incremental heuristic\hyp based algorithms is to plan in the time\hyp configuration space as in~\cite{aPa}. In this method, the state is described by the parameters of the configuration in addition to time. A backward AD* search is then done over this space after constructing a roadmap that extrapolates the positions of the moving obstacles along the time dimension. As the space is large and complex, a probabilistic roadmap technique is used.

Because the search is backward, it starts with a goal state trying to reach the start state. However, the goal state is not known because the time of arrival of the agent to the goal cannot be predetermined with certainty. To get around this, a minimum and maximum arrival times need to be determined. Multiple goals are then inputted to the search at the same position but with different time values starting from the minimum arrival time until reaching the maximum arrival time with a small time step between each goal and the next.

In addition to that, the heuristic and cost function have to include time such that the planner can be asked to prioritize time or distance as well as other factors (like smoothness, closeness to the center line... etc.) while finding a path. The cost of traversing a path is given by:
\begin{equation}
C(path) = w_t\cdot t_{path} + w_c\cdot c_{path}
\end{equation}
where $t_{path}$ is the time taken to go through the path, $c_{path}$ is the cost of the path based on all relevant criteria other than time (in the simplest form, it is only the distance), and $w_t$ and $w_c$ are weighting factors specifying the extent to which each of these values contributes to the overall cost of the path $(w_t + w_c = 1)$. If $w_t$ is larger than $w_c$, paths that take shorter time will be favored in the planning. 

Obtaining a heuristic is more complex. The heuristic in this case should be a lower bound estimate on the cost from the start node to a certain node $u$. Since the cost is composed of two parts, time and other aspects, lower bound estimations of the needed time as well as cost according to other criteria from the start node to $u$ are needed. To get these two estimations as accurate as possible (such that the search becomes as efficient as possible), Dijkstra's algorithm is used for one time at the beginning over the configuration space only (without considering the time dimension) to find the heuristic values.

Despite all these complexities, this scheme, with the way it is presented in~\cite{aPa}, only allows the agent to either move with a fixed speed or stop. Stopping after moving with a certain fixed speed and vice versa results in an infinite acceleration violating the dynamic constraints of a car\hyp like agent. The reason this planner cannot plan for multiple speeds is that the state itself does not include any information about the speed. To set a bound on the change of speed, it has to be included in the state, such that when the planner expands a state, it checks the speed of the agent at this state and selects successors (or predecessors because this is a backward search) with speeds within a limited range of the speed at the current state.

The most straightforward way of adding speed to the state in the scheme described above is to make the state space 4-dimensional consisting of the position, time and speed. This would explode the state space making the search very time consuming. In addition, it would create new complexities for this backward search. As the goal states are needed at the start of the search and as there is no information about the speed of the agent when it arrives to the goal, goal states that include all possible speeds \emph{and} all possible times of arrival of the agent at the goal position must be added to the queue at the start of the search. This would create a huge number of goals.

From this, it can be seen that this scheme is unsuitable for applying the desired constraints on the agent. The complexities of this method and its disadvantages, from the perspective of a car\hyp like agent and in the context of the autonomous driving system this path planning should be integrated in, come from several factors summarized here:
\begin{enumerate}
\item \textbf{State representation:} 
performing the search in the time\hyp configuration space makes it very complex to have control over the speed. In fact, the standard way of adding multiple speeds to time\hyp configuration roadmaps is to make connections between nodes in the roadmap denser along the time axis, such that an agent at a node can arrive at the next time step to a nearby node or to further nodes. The further the node that will be reached in one time step, the higher the speed. It is clear that this dramatically increases the complexity of constructing the roadmap and of planning itself. Even when a dense roadmap is constructed allowing multiple speeds, no rational decision can be taken for the speed of the next state because the speed at the current state is not known to the planner (speed is not part of the state).

\item \textbf{Backward search:}
The use of backward search complicates controlling the speed along the planned path as it requires entering multiple goals states that include all possible times and speeds of the agent when it arrives to the goal position. The range of the possible values might be large and taking only part of them exposes the agent to the risk of losing the optimality of the planned path. To set bounds on the change of speed, it is more natural and intuitive to start the search from the start node where the initial speed of the agent is already known. Successors can then be chosen according to the current speed. 

It is important to remember that the reason of using backward search is to allow anytime planning and replanning while the agent is moving for the reasons mentioned in section~\ref{backA}. However, the tests done in chapter~\ref{chapterStatic} showed that AD* and D* methods are only efficient for systems where replanning is frequently required. While they expand much fewer nodes, the time needed to expand one node in these algorithms is much higher than the time needed to expand one node in A*. So, for roadmaps containing a relatively small number of nodes and do not require replanning many times, A* is more efficient. This is the case with the system the designed path planning method will be integrated in. The system only plans for a relatively short horizon of distance between 100 - 200 meters. Roadmaps created over this distance would not be of a size that rationalizes the use of AD* or D*. This is especially true noticing that the system runs periodically such that no replanning would be needed between successive runs. 

\item \textbf{Preprocessing:} 
The scheme described above uses a lot of preprocessing (which are supposed to be done offline) such that path planning is done fast online. This includes constructing a large roadmap in the time\hyp configuration space and running Dijkstra's algorithm on the configuration space to obtain heuristics. This is unhandy for a system like the one dealt with here because roads data are obtained periodically from navigation systems and this preprocessed information is, most probably, not available from these systems. It would be difficult to do all this preprocessing every time the agent enters a new road and receives new data because of the very reactive nature of the task.
\end{enumerate}

\section{A Proposed Efficient Planning}
At this point, a new scheme for dynamic planning using incremental search algorithms is proposed. The proposed planner is based on forward search for the reasons explained in the previous section. First, the idea of constructing the roadmap as needed while the planner is searching is explained. This is combined with a new idea to give the planner more flexibility in selecting the heading angle of the agent. All of this is done in the configuration space only. The Dynamics of the environment are then considered by adding velocity and time to the state space and on the same time limiting the search to three dimensions only rather than four.

\subsection{Graph Formation and Position Discretization}\label{newPosition}
In the traditional 8-connected grid shown in chapter~\ref{chapterStatic}, the agent is in the center of a square cell and can move to the free centers of the eight surrounding cells. While this representation of the environment is efficient and simple, it is clearly not suitable for a car\hyp like agent traversing a road. Such an agent is typically not able to take very sharp turns due to its kinematic constraints. Moreover, the flow of the road itself prevents the agent from having some heading angles that would violate the road protocols. For example, an agent cannot move backward in a road that goes forward. 

\begin{figure}[htb]
\centering
\includegraphics[width=.5\textwidth]{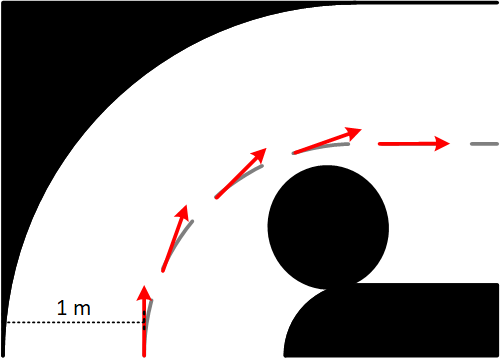}
\caption{Typical heading angle values (red arrows) on a simple road.}
\label{fig:roadPhi}
\end{figure}

To solve this issue, usually road data received from navigation systems includes a parameter known as the heading angle ($\phi$). This parameter indicates the forward direction of the road at each point along its center line as can be seen in Fig.~\ref{fig:roadPhi}. In this proposed scheme, the idea is to limit the heading angle of the agent such that it is within a certain range of $\phi$ at any point on the road.

It is more efficient to build a roadmap while searching rather than constructing a complete roadmap first and then search over it. For that, a \emph{successor function} is used. This function takes a position  $u = [x,y]$  and returns a list of positions that are valid for the agent to reach from the input position. The pseudo\hyp code below shows the basic operation of this function.

\begin{function}[htb]
\DontPrintSemicolon
$Successors = \emptyset$\\
\For{$\theta =-\theta_{max},-\theta_{max} + \theta_{step},\cdots$ \KwTo $\theta_{max}$}{
$v = u + moveLength\cdot\cos(\phi + \theta), \sin(\phi + \theta)]$ \\
\If{$isFree(u,v)$}{
add $v$ to $Successors$ with an associated cost = $c(u, v)$\\
}
}
\Return{$Successors$}
\caption{Succ(u)}\label{succFun}
\end{function}

The function initializes a list named $Successors$ as an empty list. If no valid successors were found, the list will be returned empty as it is. The function then iterates over a range of values for the deviation angle $\theta$ starting from the maximum allowed deviation angle $-\theta_{max}$ from the heading angle of the road $\phi$ until reaching $+\theta_{max}$ with a step value of $\theta_{step}$. All successors are equally away from the current node. That is, they are all on the circumference of a circle centered at the current node. This makes calculations simpler and more efficient while in the same time increases the flexibility. The $moveLength$ parameter determines the fixed length of movement. For each deviation angle, the location of the successor $v$ is determined and then a static collision checking function named $isFree$ checks if there is any static obstacles between the current node and the successor node. If no obstacle is found. The successor is added to the successor list. 

The successor list does not only hold the successors but also the cost of moving from the current node to each one of them (hence no need for an additional cost function). In the simple case of considering only distance in the cost, all successors have the same cost of $moveLength$. If the cost is more complex (for example, if it considers also the deviation angle), they may have different cost values. It is more efficient to calculate this cost in the $succ(u)$ function because it has access to all information about the current node and the successors. For example, if it was decided to include the angle of deviation in the cost, only this function is able to do this because it is the only place where this data is known.

The complexity of the collision checking function $isFree$ depends greatly on the chosen value of $moveLength$. If this value is small enough such that if there is a static obstacle between $u$ and one of the successors, it will be touched by that successors (on real roads this can be safely 1 or 0.5 meter), then it is enough to check if the successor location touches any of the static obstacles in the environment. Otherwise, multiple points should be checked along the line between node $u$ and the successor.

With this successor function, a modified A* algorithm is used as shown in algorithm~\ref{aStarClosed}. This A* version does not allow any node to be expanded more than once. For this to still guarantee an optimal solution, the heuristic has to be consistent satisfying the condition in \eqref{consistency}. Luckily, most naturally admissible heuristic (like the Euclidean distance) are also consistent~\cite{ai}. A consistent heuristic guarantees that the expanded node was reached in the lowest possible cost and hence it will not be reached in a lower cost in the future such that it must be expanded again.

\begin{algorithm}[htb]
\DontPrintSemicolon
$Open = Closed = \emptyset$\\
$g(u_{start}) = 0$\\
insert $u_{start}$ into $Open$\\
\Repeat{$cell_u=cell_{u_{goal}}$ \textbf{\emph{or}} $Open=\emptyset$}{
$u$ = element from $Open$ with minimal $(g(u)+h(u))$\\
move $u$ from $Open$ to $Closed$\\
$Successors$ = Succ($u$)\\
\ForAll{$v \in Successors$}{
\If{\textbf{\emph{not}}\emph{(}$\exists v_2\in Closed$ \textbf{\emph{such that}} $cell_{v_2} = cell_v$\emph{)}}{
$g(v) = g(u)+c(u,v)$\\
$bp(v)= u$\\
\If{$\exists v_2 \in Open$ \textbf{\emph{such that}} $cell_{v_2} = cell_v$}{
\If{$g(v_2) > g(v)$}{
replace $v_2$ by $v$ in $Open$\\
}
}
\Else{
insert $v$ in $Open$\\
}
}
}
}
\caption{A* with a $Closed$ set\label{aStarClosed}}
\end{algorithm}

The proposed A* search forces two other limitations on the search making it very efficient. Once a node is expanded it enters the $Closed$ set. Then, it is not possible to expand this node again and also to expand any node within the same cell (line 9). More than that, the $Open$ queue can only hold one node per cell (line 12). A node $v_2$ in the $open$ queue is replaced by another node $v$ in the same cell only if the cost of reaching $v$ from the start (the $g$-value) is lower (line 13). After successfully finishing the search. The path can be traced by back pointers $pb$ starting from the node found on the goal cell.

The cell $cell_u$ in which a node $u=[x,y]$ is located is calculated very easily:
\begin{equation}\label{cellEq}
cell_u = [\text{floor}(x/cellLength),\text{floor}(y/cellLength)]
\end{equation}
where $cellLength$ is a system parameter determining the length of one cell. The lower this parameter is, the higher the resolution of the search, but also the more expensive as will be seen in section~\ref{cpp}.

With these restrictions, and under the condition that the heuristic is consistent with respect to the cost, it can be easily intuitively proven that the resultant scheme returns paths that are either cheaper or have the same cost of paths returned by normal uniform grid schemes, with at most the same computational cost (but in most cases far lower because the step of constructing the roadmap is omitted). A mathematical proof of this statement, however, is not within the scope of this report.

\begin{figure}[htb]
\centering
\begin{subfigure}{.49\textwidth}
  \centering
  \includegraphics[width=\textwidth]{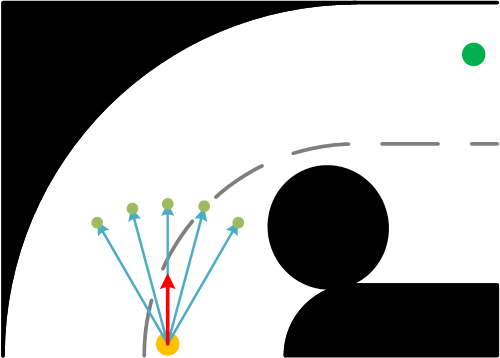}\par\medskip
  \includegraphics[width=\textwidth]{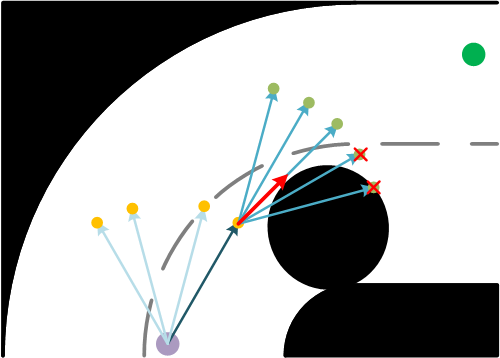}\par\medskip
  \includegraphics[width=\textwidth]{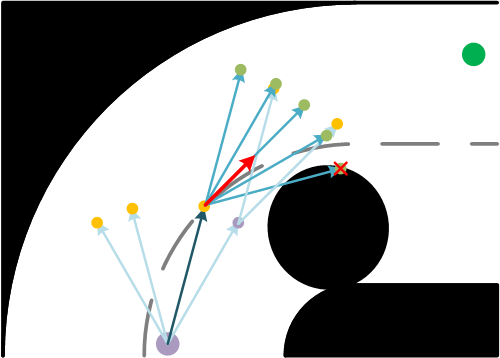}\par\medskip
  \includegraphics[width=\textwidth]{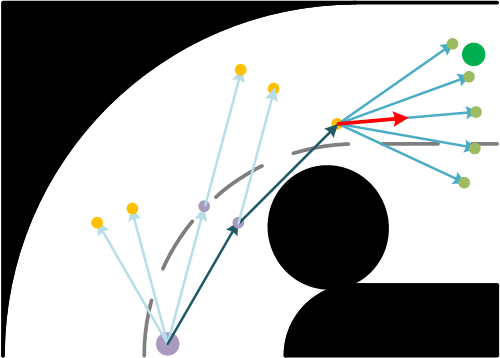}
  \caption{successors returned by $Succ(u)$ (green) and $\phi$ at $u$ (red).}
\end{subfigure}
\hfill
\begin{subfigure}{.49\textwidth}
  \centering
  \includegraphics[width=\textwidth]{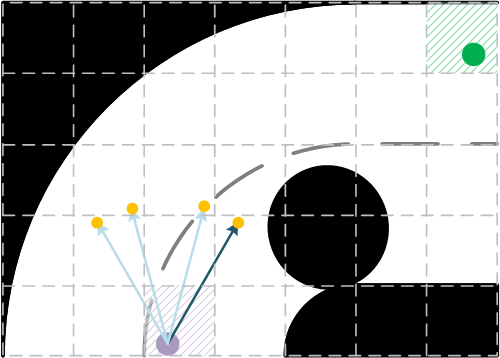}\par\medskip
  \includegraphics[width=\textwidth]{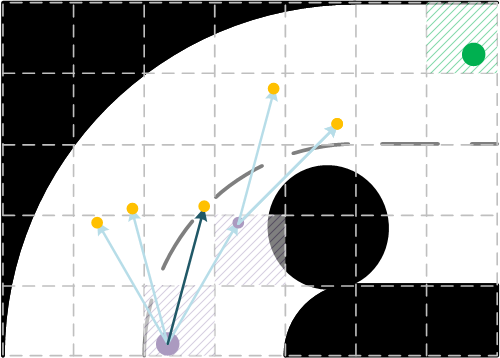}\par\medskip
  \includegraphics[width=\textwidth]{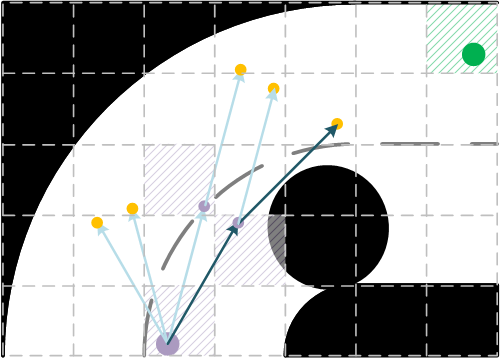}\par\medskip
  \includegraphics[width=\textwidth]{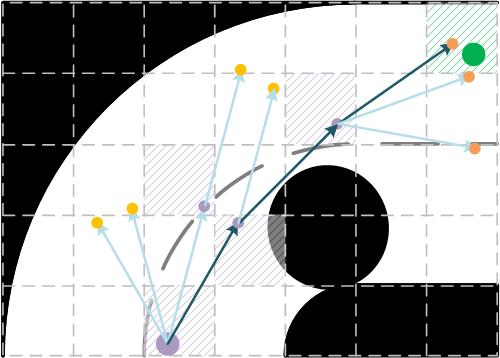}
  \caption{nodes kept in the $Open$ queue (orange) and in the $closed$ list (magenta).}
\end{subfigure}
\caption{Operation of the successor function and A* on a simple path planning problem.}
\label{fig:succFun}
\end{figure}

The operation of the new successor function and the proposed A* search is best explained through an example. Figure~\ref{fig:succFun} shows the operation of $Succ(u)$ (left part) and the nodes A* keeps in the $Open$ queue and the $Closed$ set (right part). At first (first row - left), the algorithm is given a start (orange) and a goal (green) positions. The start position is expanded and the successor function is called. Here, $moveLength=1m$, $cellLength=0.5m$, $\theta_{max} = 30^o$ and $\theta_{step}=15^o$. The successor function returns five valid successors. These successors are to be added to the $Open$ queue. However, the modified A* algorithm doesn't allow more than one node in a cell. So, only four nodes are admitted to the queue (first row - right). At the same time, the expanded state is added to the $Closed$ set and hence it is not possible to add any node to the same cell anymore.

The modified A* then, based on the priorities of the nodes in $Open$, expands another state (second row - left) and calls the successor function on it. The function returns three successors only because the others pass through an obstacle. Since two of them lie on the same cell, only two of the successors are added to the $Open$ queue (second row - right), while the expanded state is put in the $Closed$ set.

A third node in the $Open$ queue is expanded (third row - left). The successor function returns four valid successor positions. Since three of these positions are on the same cells as other nodes in $Open$ and since they don't have a lower cost from start ($g$-value), they do not replace the existent ones. At the end, only one successor is added to the queue, and the parent state is added to the $Closed$ set (third row - right).

The search algorithm then expands a fourth node (fourth row - left). The successor function finds five valid nodes. Only one node per cell is accepted to the queue. So, three nodes are added (fourth row - right). It can be seen that one of the nodes touches the goal cell. The search will end successfully whenever this node is expanded (most probably the next step of the search).

\begin{figure}[htb]
\centering
\includegraphics[width=.5\textwidth]{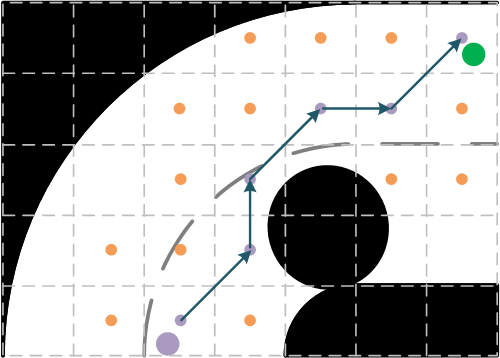}
\caption{A path that results from a typical uniform roadmap approach.}
\label{fig:succUni}
\end{figure}

Comparing the resultant path (shown in dark arrows in the last plot in Fig~\ref{fig:succFun}), to a one that would result from a typical uniform roadmap method with the same cell size as seen in Fig.~\ref{fig:succUni}, demonstrates the merits of the suggested method. It can be seen that the suggested approach expands less nodes (magenta), keeps less nodes in the $Open$ queue (orange) and finds smoother and shorter (or cheaper) paths. Another important advantage is that the position of a node (including the start and goal nodes) can be anywhere in the path while in the uniform sampling approach, only predefined locations are allowed.

It should be mentioned that for this modified A* algorithm to work properly, $moveLength$ has to be greater than the diagonal length of the a cell ($\sqrt{2}\cdot cellLength$). This ensures that all the successor positions will be outside the cell of the parent node. Otherwise, all successors inside the cell of the parent node will be automatically eliminated by the modified A* which can result in the failure of the search. 
\clearpage

\subsection{Collision Checking and Speed Discretization}\label{newSpeed}
To be able to avoid dynamic obstacles and at the same time force dynamic constraints on the velocity of the agent, a suitable state representation is needed. In the proposed method, the state is represented by $state = (pos, speed, t)$, where $pos$ is the position of the center point of the agent, $speed$ and $t$ are the speed the agent will be at when reaching this state, and the time at which this state is reached respectively. This is a 4D representation. However, it will be seen that the search is done over three dimensions only. 

A new successor function named DynSucc is used to return successor states with this 4D representation as shown in the pseudo\hyp code below. The new successor function is very similar to the one presented in previous section, but it returns successors with different speeds. The function iterates over several values for acceleration $a$, starting from the maximum allowed deceleration $-a_{max}$ until reaching the maximum allowed acceleration $a_{max}$ returning valid successors for each acceleration value in the different directions determined by the inner loop iterating over values of the deviation angle $\theta$. 

\begin{function}[htb]
\DontPrintSemicolon
$Successors\leftarrow\emptyset$\\
\For{$a =-a_{max},-a_{max} + a_{step},\cdots$ \KwTo $a_{max}$}{
\If{$u_{speed}^2 + 4\cdot a\cdot moveLength\geq 0$}{
$speed = \left(u_{speed}+(a)\cdot\sqrt{u_{speed}^2 + 4\cdot a\cdot moveLength}\right)/2$\\
\If{$0\leq speed\leq maxSpeed$}{
$moveTime =  moveLength/speed$\\
$t = u_t+moveTime$\\
\For{$\theta =-\theta_{max},-\theta_{max} + \theta_{step},\cdots$ \KwTo $\theta_{max}$}{
$pos = u_{pos} + moveLength\cdot[\cos(\phi + \theta), \sin(\phi + \theta)]$ \\
$v=(pos, speed, t)$\\
\If{\emph{isDynFree}($u, v$)}{
add $v$ to $Successors$ with an associated cost = $w_t\cdot moveTime + w_c\cdot c(u, v)$\\
}
}
}
}
}
\Return{$Successors$}
\caption{DynSucc(u)}\label{succFunV}
\end{function}

For each acceleration value, the speed required to achieve this acceleration is calculated (line 4). The calculation assumes that the agent will change its speed instantly after leaving the parent node $u$ and continue applying a fixed speed until reaching the successor $v$. This assumption is very important in simplifying collision checking calculations as will be seen later. With this, the time needed to reach the successor with this speed is: 
\begin{equation}
moveTime = moveLength/speed
\end{equation}
It can be seen that the average acceleration between node $u$ and a successor $v$ is then:
\begin{equation}
a = \frac{speed - u_{speed}}{moveTime} = \frac{(speed - u_{speed})\cdot speed}{moveLength}
\end{equation} 
Since all variables are known except $v_{speed}$, the equation above is a quadratic equation that can be solved to get:
\begin{equation}
speed = (u_{speed} + \sqrt{u_{speed}^2+4\cdot a\cdot moveLength})/2
\end{equation}
Because speed is not allowed to be negative (the agent is not allowed to go backward), only the positive solution is taken.

The function checks if the resultant $speed$ will be real (line 3). If so, it is calculated (line 4) and checked against the speed limits of the agent (line 5). After ensuring that this speed is valid, the time needed to reach successors that are $moveLength$ away with this speed is calculated (line 7), then successors with this speed and time but different deviation angles from $\phi$ are calculated exactly as done in the function Succ. The only difference is how collisions are checked, which happens in a function named isDynFree. It returns true only if it is safe for the agent to go from node $u$ to the successor $v$ with the specified speed. The principle behind this function is explained below.

After iterating through all acceleration values and deviation angles, DynSucc returns a list of successor. The maximum number of returned successors is equal to the number of possible acceleration values times the number of possible deviation angles. Each successor is returned with an associated cost. The cost is composed of a time part and a path part weighted by weighting factors $w_t$ and $w_c$ respectively such that $w_t+w_c = 1$. The path part includes the costs of all factors other than time. In the simplest form, it is only distance. The weighting factors allow the search function to find paths optimized for either time or other factors based on what is desired by the user. For example, if $w_t>w_c$ paths that take shorter time will be favored over paths that have shorter distances (if distance is what is considered in the path part of the cost). The cost is calculated in the successor function for the same reasons explained in the previous section. 

\begin{figure}[htb]
\centering
\includegraphics[width=.7\textwidth]{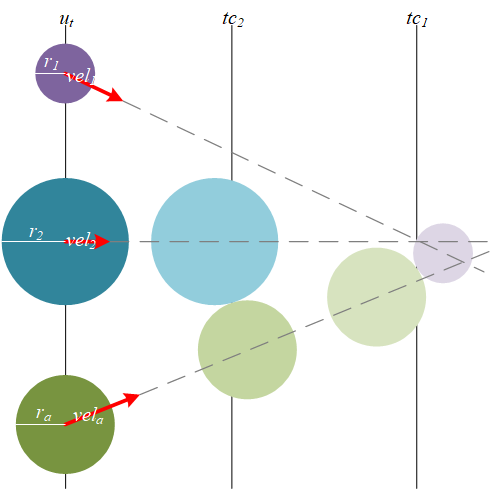}
\caption{Checking the minimum collision time}
\label{fig:coll}
\end{figure}

To ensure that the transition of the agent from node $u$ to a successor $v$ is collision\hyp free, a new method, based on collision times, is used. In this method, the agent as well as the moving obstacle are modeled as moving 2D shapes with different velocity vectors\footnote{A velocity vectors has a magnitude equal to the speed and a direction equal to the direction of movement.}. For simplicity, the agent and the obstacles are considered as circles with various radii as shown in Fig.~\ref{fig:coll}.

Suppose that the agent is positioned, at the time of the current node $u_t$, in $pos_a = u_{pos}$, and has a velocity vector $vel_a$ and a radius $r_a$. Suppose also that the environment has one moving obstacle with position $pos_o$ at the time of the current node, velocity vector $vel_o$ and radius $r_o$. Assuming that the agent and the obstacle are moving with fixed speeds, their positions, after a time interval $tc_o$ passes, are given by:
\begin{align}\label{colltime1}
pos_a(tc_o) = pos_a + vel_a\cdot tc_o\\
pos_o(tc_o) = pos_o + vel_o\cdot tc_o\nonumber 
\end{align}
Now, if a collision between the agent and the obstacle will happen at $tc_o$ (collision time with obstacle $o$), the distance between their centers at that time will be equal to the sum of their radii:
\begin{equation}\label{colltime2}
\parallel pos_a(tc_o) - pos_o(tc_o)\parallel = r_a + r_o
\end{equation}
Squaring both sides and rearranging the terms leads to:
\begin{multline}\label{colltime3}
\parallel vel_a - vel_o\parallel^2\cdot tc_o^2 + 2\cdot(vel_a-vel_o)\cdot (pos_a-pos_o)\cdot tc_o\\
+ \parallel pos_a - pos_o\parallel^2 - (r_a+r_o)^2= 0
\end{multline}
which is a quadratic equation that leads to two solutions for $tc_o$. If no one of them is a positive real number, then no collision will ever happen between the agent and the obstacle. If only one of the solutions is a positive real number then it is the time of collision. Otherwise, the collision time is the minimum of the two solutions.

If the collision time is greater than $moveTime$, the agent is safe from colliding with the obstacle while moving from $u$ to $v$. When the environment contains multiple moving obstacles, collision times are calculated for all of them. It is enough to find out one collision time below $moveTime$ to conclude that the transition is not valid as done in the pseudo\hyp code of the isDynFree function shown below.

\begin{function}[htb]
\DontPrintSemicolon
\If{\emph{isFree($v_{pos}$)}}{
extrapolate the positions of the moving obstacles to time $u_t$\\
\ForAll{$m \in$ moving obstacles}{
calculate $tc_m$ (the time of collision with $m$) by solving~\eqref{colltime3}\\
\If{$tc_m\leq moveTime$}{
\Return false 
}
}
\Return true\\
}
\Else{
\Return false\\
}

\Return{$Successors$}
\caption{isDynFree(u,v)}
\end{function}

Usually the agent receives information about the positions and velocities of the moving obstacles before planning, at the time of the start state. So, before calculating the collision times, isDynFree linearly extrapolates the positions of the moving obstacles to the time of the parent node. In addition to that, the radii of the agent can be made such that they grow up with time to compensate for uncertainty. This extrapolation step is better done in the successor function itself since the result is the same for all successors of one parent node. This is what was done in the MATLAB implementation that will be introduced in the next section.

With this successor function and this collision checking technique, the proposed scheme uses a modified ARA* (anytime repairing A*) algorithm to plan a path and construct the search space while planning. The pseudo\hyp code of the algorithm is shown below. The algorithm contains three functions: key($u$) which calculates the priority of a node in the $Open$ queue, improvePath which performs an $\epsilon$-suboptimal search, and main which initializes the search and manages it. The space over which this algorithm performs the search is three\hyp dimensional composed of the position and velocity. Time is only used to calculate the cost and the heuristic functions and to check for collisions.

\begin{algorithm}[htb]
\DontPrintSemicolon
\SetKwProg{Fn}{Function}{}{end}
\Fn{key(u)}{
\Return $g(u)+\epsilon\cdot h(u)$
}
\Fn{improvePath}{
\While{$Open\neq\emptyset$ \emph{\textbf{and}} $g_{goal} > \min_{u\in Open}(key(u))$}{
$u$ = element from $Open$ with minimal $key(u)$\\
move $u$ from $Open$ to $Closed$\\
$Successors$ =  DynSucc($u$)\\
\ForAll{$v \in Successors$}{
\If{\textbf{\emph{not}}\emph{(}$\exists v_2\in Closed$ \textbf{\emph{such that}} $block_{v_2} = block_v$\emph{)}}{
$g(v) = g(u)+c(u,v)$\\
$bp(v) = u$\\
\If{$cell_v = cell_{u_{goal}}$ \emph{\textbf{and}} $g_{goal}>g(v)$}{
$g_{goal} = g(v)$\\
$u_{goal} = v$\\
}
\If{$\exists v_2 \in Open$ \textbf{\emph{such that}} $block_{v_2} = block_v$}{
\If{$g(v_2) > g(v)$}{
replace $v_2$ by $v$ in $Open$\\
}
}
\Else{
insert $v$ in $Open$\\
}
}
\ElseIf{{\textbf{\emph{not}}\emph{(}$\exists v_2\in Incons$ \textbf{\emph{such that}} $block_{v_2} = block_v$\emph{)}}}{
insert $v$ in $Incons$\\ 
}
}
}
}
\Fn{main}{
$Open = Closed = Incons = \emptyset$\\
$g(u_{start})= 0$; $g_{goal} = \infty$; $\epsilon = \epsilon_0$\\
insert $u_{start}$ into $Open$\\
improvePath\\
publish current $\epsilon$\hyp suboptimal solution\\
\While{$\epsilon>1$ \emph{\textbf{and}} time is available}{
decrease $\epsilon$\\
move states in $Incons$ to $Open$\\
empty $Closed$\\
update priorities of all states in $Open$ using $key(u)$\\
improvePath\\
publish current $\epsilon$\hyp suboptimal solution\\
}
}
\caption{Modified ARA*\label{mAra}}
\end{algorithm}

\begin{figure}[htb]
\centering
\includegraphics[width=\textwidth]{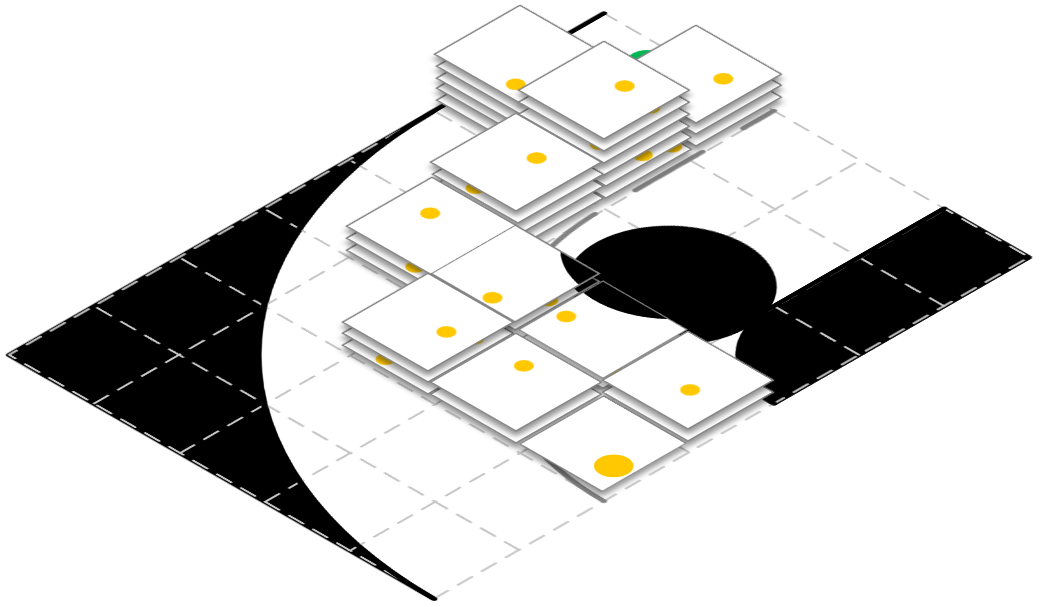}
\caption{Sample nodes with velocity dimension.}
\label{fig:succS}
\end{figure}

Several optimization ideas are applied in this algorithm in addition to an anytime feature allow producing quick sub-optimal paths in time\hyp critical situations. The idea of limiting the number of nodes per cell to one node is still applied here. However, a cell is a 2D area over the 2D space. Speed serves as a third dimension in the search space here. So, this idea is extended to allow only one node per a speed value in a cell. Figure~\ref{fig:succS} shows a visualization of this idea where every cell carries a pile of layers. Each layer represents a different velocity and only one position is allowed per layer. 

\begin{figure}[htb]
\centering
\includegraphics[width=0.75\textwidth]{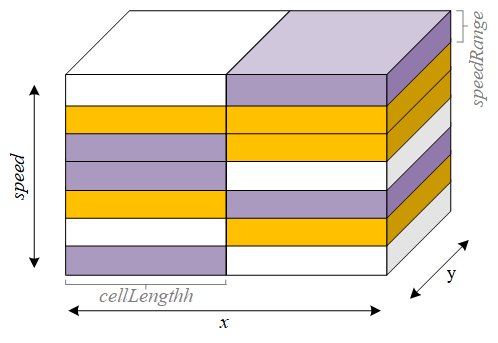}
\caption{Blocks over two adjacent cells.}
\label{fig:nodesBlocks}
\end{figure}

It can be noticed that the number of layers grows up as the cell gets away from the start position. In general, the number of possible speeds at a node grows fast the deeper this node is in the search tree, reaching very high numbers and making the search very time consuming. In many cases the differences between these speed values is very small. To solve this problem, the algorithm allows only one node per \emph{block}. The block to which a node $u$ belongs is defined as a 3-integers array:
\begin{equation}
block_u = [cell_{u_{pos}}, \text{floor}(u_{speed}/speedRange)]
\end{equation}
where $cell_{u_{pos}}$ is defined in~\eqref{cellEq} and $speedRange$ is a system parameter chosen based on the needs of the user. The lower the value of $speedRange$ the more precise the search with respect to  the speed but also the more expensive it is. Figure~\ref{fig:nodesBlocks} shows blocks over two adjacent cells. Some of these blocks contain nodes that have been expanded before and hence no node can be added to any position in these blocks (magenta), other blocks contain nodes in the $Open$ queue (orange), while others are empty (white). 
\clearpage

The algorithm gets a start state $(pos, speed, t)$ and a goal \emph{position}. It doesn't get a goal \emph{state} because it is unknown\footnote{speed and time at reaching the goal are unknown as explained before.}. The algorithm then starts by initializing the $Open$ queue and the $Closed$ and $Incons$ (for inconsistent) sets to be empty (line 24). Three values are also initialized: $g$-value of the start node is initialized to zero, $g_{goal}$ (a variable that holds the minimum cost the goal position has been reached by) is initialized to infinity and the inflation factor $\epsilon$ is initialized to its initial value $\epsilon_0$. $\epsilon_0$ must be greater than or equal to 1. If it is one, the algorithm will perform an A* search.

The start node is then inserted to the $Open$ queue (line 26) and a search is performed using improvePath. improvePath has a while loop that continues running as long as the minimum key in the $Open$ is lower than the minimum found goal cost (initially infinity) and the $Open$ queue is not empty. It picks the state $u$ that has the minimum key in the $Open$ queue. Since this node is expanded, it is moved to the $closed$ set (line 6) and it is not possible anymore to add any nodes from the same block to the $Open$ queue. 

The valid successors of the expanded state are obtained using the function DynSucc($u$). A successor $v$ is added to the queue only if there is no state in $block_v$ inside $Closed$. If that is not the case, then $v$ is added to the $Incons$ set if there is no node inside that set that lie on the same block. The $Incons$ set contains nodes to be reconsidered in next searches because they may have g-values lower than the g-values of the nodes that were expanded before in the same blocks (hence the name). 

If no node was expanded before in $block_v$, the successor is considered for admittance to $Open$. First, it is checked if it lies on the same cell as the goal position. If so, $g_{goal}$ is changed to the $g$-value of $v$, and $v$ is labeled as the goal state. If there is another node $v_2$ in open that is on the same block as $v$ then it is replaced by $v$ only if $v$ has lower cost from the start node ($g$-value). If $v$ is the first state in $block_v$, it is directly inserted in the $Open$ queue. When the search finished successfully, the path can be traced by following back pointers $bp(u)$ starting from $u_{goal}$ until reaching $u_{start}$

After improvePath returns, the found solution is published and is guaranteed to be $\epsilon$-suboptimal (its total cost is guaranteed to be less than or equal $\epsilon$ times the cost of the optimal path). If time is still available and $\epsilon$ has not reached one yet, the following sequence of actions is repeated: $\epsilon$ is decreased by a certain determined amount, states in $Incons$ are moved to $Open$ and all keys in $open$ are updated (keys have changed because $\epsilon$ changed). Then improvePath is called again.

It is to be noted that, for this scheme to work, a consistent heuristic with respect to path cost \emph{and} time cost is needed. A simple one would be as shown below assuming that the path cost only considers the distance. This heuristic estimates the distance to the goal as the Euclidean distance and estimates the time needed to reach the goal as the time needed to traverse this Euclidean distance at the maximum speed the agent can have. Note that the weighting factors in the heuristic have to be the same as those in the cost. While this heuristic is consistent, it is not very informing. Obtaining a more informing heuristic would be more computationally expensive. However, the proposed anytime planner can quickly find paths even with this not so informing heuristic.

\begin{function}[htb]
\DontPrintSemicolon
\If{$cell_{u_{pos}}=cell_{goal}$}{
\Return{0}
}
euclideanDistance = $\parallel goal_{pos} - u_{pos}\parallel$ \\
timeToGoal = euclideanDistance/maxSpeed\\
\Return{$w_t\cdot timeToGoal+ w_c\cdot euclideanDistance$}
\caption{h(u)}
\end{function}

The resultant scheme that has been shown in this section produces paths that are more consistent with the kinematic constraint of the agent (heading angle) as well as the dynamic constraint (speed). The presented algorithm gives flexibility to optimize the found path with respect to time or other aspects (like distance, closeness to the center line, smoothness,... etc.). It is also able to quickly produce suboptimal paths in time\hyp critical situations and continue improving the path as time allows.

The speed of an agent following a path produced by this scheme is a staircase function of time, whenever the agent leaves a state $u$, it immediately change its speed at time $u_t$ to the speed of the following state along the path $v_{speed}$. This section is concluded by discussing the consequences of changing the speed function of the agent to a more linear one. Suppose when the agent leaves a state $u$ along the path, it does not change its speed immediately. Instead, it linearly accelerates until reaching the next state. In this case, we should go back to the successor function and change some calculations.
Suppose that we want to check if $v$ is a dynamically valid successor of $u$ for some acceleration value $a$. From equations of motions, it can be found that:
\begin{equation}
moveLength = u_{speed}\cdot moveTime +1/2\cdot a\cdot moveTime^2
\end{equation}
all the variables in this equation are known except $moveTime$. This is a  quadratic equation that can be solved easily and has only one solution due to the fact that moveTime cannot be negative:
\begin{equation}
moveTime = (\sqrt{u_{speed}^2 + 2\cdot a\cdot moveLength}-u_{speed})/a
\end{equation}
The speed that will be reached by the agent applying constant acceleration from $u$ to the successor $v$ is then:
\begin{equation}
v_{speed} = u_{speed} + a\cdot moveTime
\end{equation}

Now, the position of $v$ is calculated as in the DynSucc function (line 9) and the validity of the successor is checked. The transition between $u$ and $v$ was found to be free of static obstacles, and the collision times of dynamic obstacles are calculated. However, with the current modeling of the agent motion between the two states, it does not move with a constant speed. So, \eqref{colltime1} and \eqref{colltime3} are not valid anymore. For an agent $a$ and a moving obstacle $o$, we continue to assume that the obstacle is moving in a constant speed, but the agent has acceleration now. Hence:
\begin{align}
pos_a(tc_o) = pos_a + vel_a\cdot tc_o + 1/2\cdot a\cdot tc_o^2\\
pos_o(tc_o) = pos_o + vel_o\cdot tc_o\nonumber 
\end{align}
When substituting this in \eqref{colltime2}, squaring both sides and rearranging terms, the result is a fourth degree polynomial in $tc_o$ that has four different possible solutions. Obtaining these solutions analytically or numerically is very computationally expensive that it will significantly affect the execution time of the whole algorithm. After obtaining the solution, the collision time can be decided as done with solutions of the quadratic equation in \eqref{colltime3}.

This modeling, however, allows having a much more informing heuristic where the time needed to reach the goal is estimated assuming that the agent will continue accelerating with the maximum available acceleration until reaching the goal instead of assuming that it will directly move with the maximum speed. It could be possible, with a small value of $moveLength$ and some assumptions about the environment to use this motion modeling (linearly accelerating form a node to a successor) and still check the collision time assuming that the agent will move from $u$ to $v$ with a fixed speed equal to that reached at $v$. This idea, however, is not further investigated in this report.

\section{Simulation Environment}\label{MATLAB2}

\begin{figure}[htb]
\centering
\includegraphics[width=\textwidth]{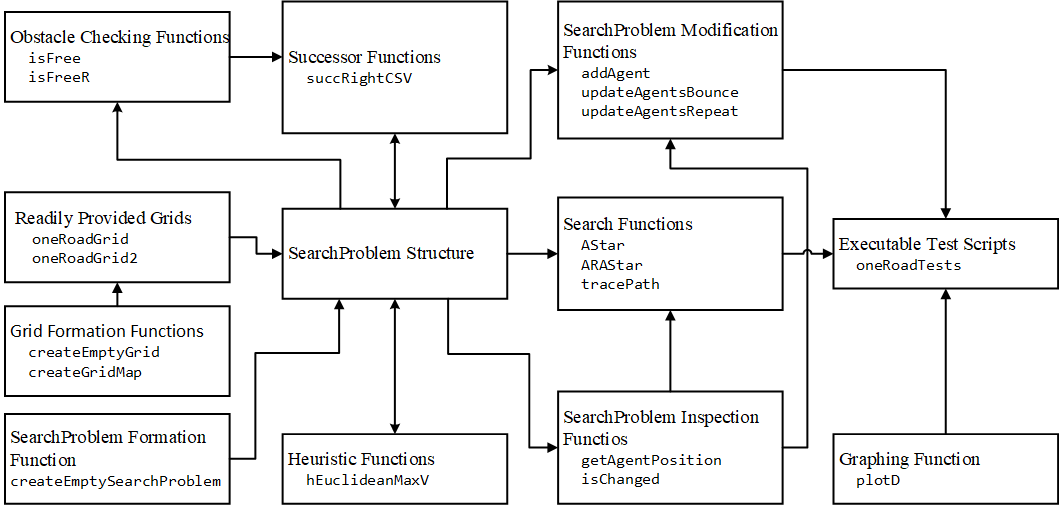}
\caption{A block diagram of the components of the simulation environment.}
\label{fig:sysBlockD}
\end{figure}

A simulation package, very similar to the one discussed in section~\ref{MATLAB}, was written. Because of the strong similarity, this section is only to discuss the main differences. The main aim for this package is to provide a quick and reliable way to validate and test the ideas presented in the previous section, mainly using fixed\hyp step simulation. A block diagram of the main components of the package is shown in Fig.~\ref{fig:sysBlockD}.

The main difference is the use of a new structure named \textsf{searchProblem}. This structure holds all the information needed to do the search, including the grid, the moving obstacles and the known information about them, the successor and heuristic functions as well as the system parameters like the weighting factors of the cost and heuristic, $moveLength$, $cellLength$ and $speedRange$. An empty version of this structure can be created using the function \textsf{createEmptySearchInfo}. The different elements of the search problem can then be added to this empty structure. 

The grid is still represented with the structure \textsf{grid} similar to the one seen in section~\ref{MATLAB} with a slight difference. Now when creating a grid using grid formation functions, an input representing the length of one cell inside the grid in meters has to be provided. This input does not have to be the same as $cellLength$. However, the input is necessary to determine which position lies in which cell (which index in the \textsf{obstacles} matrix) and hence be able to check for static obstacles. The grid is used just as a simple mean of representing the shape of the environment and the positions of the static obstacle. It doesn't play any other role beyond that, unlike what was done on the simulation package seen in section~\ref{MATLAB} where the search itself was done over the cells of the grid. Here the search and discretization of movement are completely independent of the grid cells. In a practical situation, other representations of the environment might be more efficient\footnote{as will be seen in the C++ implementation presented in the next section.}.

Two readily available grids are provided. The first one is \textsf{oneRoadGrid}. It represents a $111\times 12 m^2$ one direction road that goes from left to right as can be seen in Fig.~\ref{fig:roadAni}. The cell length\footnote{again, this has nothing to do with the discretization parameter $cellLength$.} of this grid is one (meaning that the smallest static obstacle is a $1 m^2$ square). The second grid (\textsf{oneRoadGrid2}) has the same dimensions as the first one but the length of one cell in the grid is $2 m$ (so the \textsf{obstacles} matrix is half the size of that of the first one and the smallest static obstacle possible is a $4 m^2$ square). It was used to ensure that the search is independent of the number of cells in the grid, which was successfully confirmed. Note that the direction of this road is always to the right ($\phi = 0$ all over the road). This makes it simpler to do the planning and use the successor functions discussed earlier. To allow this package to deal with curved roads, however, a way must be found to encode the information about the forward direction of the road in any specific location in the grid (which is not done in the current version of the package). 

The package includes two different static obstacles checking function. The function \textsf{isFree} returns true simply when the center position of the agent (which is modeled as a circle) is on a cell that does not contain a static obstacle in the \textsf{obstacles} matrix. This makes the calculation very fast. However, it doesn't guarantee that the complete agent circle doesn't touch any static obstacle. \textsf{isFreeR} guarantees this by checking the positions of the four corners (two horizontal and two vertical) of the agent circle as well as the center position. However, it takes a considerably longer time.

The successor function \textsf{succRightCSV} is a direct implementation of the DynSucc function presented in the previous section. In this function, $\phi$ is assumed to be always 0 and hence it is only suitable for roads where the forward direction is rightward. The heuristic function \textsf{hEuclideanMaxV} is also a direct implementation of the heuristic shown in the previous section. Both the successor and the heuristic functions take as inputs a state as well as an instance of the \textsf{searchProblem} structure. At the same time, handles of the functions are kept in \textsf{searchProblem}, hence the double arrows in the block diagram.

Moving obstacles are modeled as circles with different radii and fixed velocities and are saved in a matrix named \textsf{agents} inside the $searchProblem$ structure, where each row of the matrix represents the information of one obstacle. An obstacle can be added using the function \textsf{addAgent} which takes the initial position, radius, speed, direction (angle). Positions, speeds and other parameters of the moving obstacles can be modified during the fixed\hyp time simulation by directly accessing the matrix or by using functions like \textsf{updateAgentsBounce} and \textsf{updateAgentsRepeat}. Both of them change the current position of the agent by extrapolating their initial positions to the current time in the simulation. The first function ``bounces back'' any agent touching the borders of the grid, by changing its direction of motion to the reflection of the angle this agent touched the border with. The second one, brings the agent, when it goes out of the grid, from the opposite side of the grid with the same direction of motion. The package also allows adding moving obstacles with zero speed. In this case, they are effectively static obstacles.

The search functions in this package are implemented in a completely different way than it was done in the first package. This change is primarily because of the fact that, unlike the 8-connected grid\hyp based planning done in the first package, the maximum number of nodes to be considered in the search here is far greater than the number of cells in the grid. So, $g$-values cannot be saved in matrices of the same size as the \textsf{obstacles} matrix. For that, a matrix named $Open$ is used to keep all the search nodes and the associated information, including keys, $g$-values, back pointers... etc. Each row in this matrix represents one node and hence, the number of nodes that can be considered is not limited in any way. Two search functions were implemented in the package. \textsf{ARAStar}, which is a direct implementation of the algorithm shown in the previous section, and \textsf{AStar}, which is the same but without the anytime feature (it performs the search for only one value of the inflation factor $\epsilon$ and does not change $\epsilon$ during the search). A path is exctracted in a very similar way to what was done before by the function \textsf{tracePath}. The results of the search are saved in the familiar \textsf{searchInfo} structure together with some other useful information like the execution time of the search.

The executable test script \textsf{oneRoadTests} is where everything is wrapped together. The script initializes a \textsf{searchProblem} and calls a search function to find a path. \textsf{oneRoadTests} runs a fixed\hyp step simulation. The time step can be determined by the user. At each step, the locations of the moving obstacles are updated. If any change happened (detected by the function \textsf{isChanged}) such that the trajectories of the obstacles are changed, the search function is called to start searching from scratch again. Graphing is done via a function named \textsf{plotD} in a similar way to what is done in the previous package. However, the current graphing function is optimized to make animation very fast by changing the locations of the drawn objects at each time step rather than deleting and redrawing them again. The animation loop is run inside the test scrip itself and no separate function is provided for that.

\section{Results and Analysis}
Various tests were done to explore the behavior, advantages and limitations of the proposed scheme. A C++ implementation of the modified A* algorithm shown in section~\ref{newPosition} was written and tested over a real road data from DLR. In addition to that, MATLAB simulations, using the package presented in the previous section, were also done to validate the ideas shown in section ~\ref{newSpeed}.

\subsection{C++ Implementation and Tests}\label{cpp}

A C++ Implementation of all the ideas presented in section~\ref{newPosition} was done. The code was written to be used with environments described by a usual format of road data that can be obtained from navigation systems. In this format, the road is described by position points along its center line. From these points, the curvature of the road can be known. For each point along the center line, there is a value for $\phi$ (the road forward direction) and a parameter named $s$ indicating the center line distance from the start point of the road. Combined with the width of the road, this representation has all the needed information to plan a path. The forward direction of the road at any point can be easily obtained by finding the closest point on the center line and getting its associated $\phi$.

\begin{figure}[htb]
\centering
\includegraphics[width=\textwidth]{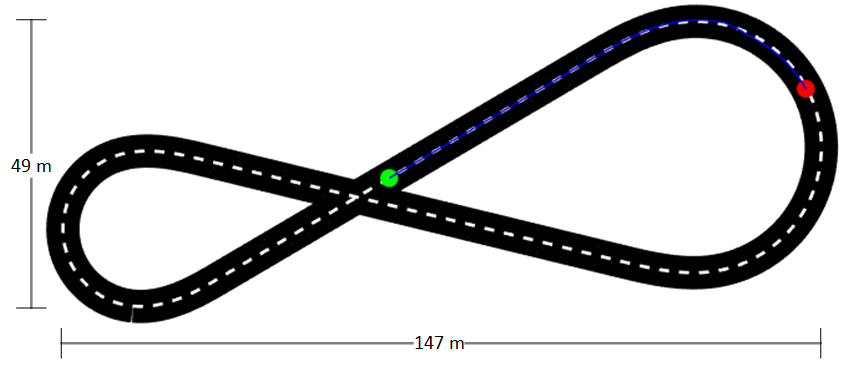}
\caption{The test road used in C++ tests and the resultant plan (blue) from start (green) to goal (red) with $cellLength = 0.5$ and $moveLength = 1$}
\label{fig:road}
\end{figure}

Data from a real test road in DLR was used to test the implementation. The road can be seen in Fig.~\ref{fig:road}. Planning was done from several start points to ensure that the search method is working properly. The implemented method is able to plan from any start location in the road to any goal location. The plan, if found, guarantees that the complete agent circle will completely stay within the road. The current implementation does not consider any static obstacle other than the road borders. However, once a suitable representation of static obstacles within the road is decided, the code can be easily extended to avoid them with minimal additional computational cost.

Table\ref{cppTests}, shows execution times as well as other important information about the search, including number of nodes inserted to $Open$, number of expanded nodes (nodes inserted to $Closed$), number of nodes in the found path and length of the found path (in meters). These metrics are shown for a specific search problem, where the agent starts from the green circle shown in Fig.~\ref{fig:road} and plans for 100 ahead to the red circle. The execution times are calculated by averaging the times of one thousand executions to get more reliable results. Note that these times where measured by running the code in a desktop operating system. In a practical situation where the code is run in an embedded processor with a real time operating system (RTOS), the execution time would be much shorter. The cost function used is the Euclidean distance added to the absolute value of the deviation angle ($\theta$) from the road direction ($\phi$), which causes the planner to favor paths that closely follow the center line of the road (this can be seen in the resultant plan shown in Fig~\ref{fig:road}). The used heuristic is the center line distance between a location and the goal, which is simply calculated as the difference between two values of the $s$ parameter that comes with the road data.

\begin{table}[htb]
\centering
\caption{Planning data for different values of $cellLength$ and $moveLength$.}\label{cppTests}
\resizebox{\textwidth}{!}{\begin{tabular}{cccccccc}
\hline
$moveLength$ & time (ms) & \# $Open$ & \# Expanded & \# Path & Path length & Path cost\\
\hline
\multicolumn{7}{c}{$cellLength$ = 0.5}\\
\hline
0.4 & -     & 1 & 1 & - & -\\
0.5 & 1.392 & 729 & 694 & 202 & 100.5 & 108.005\\
0.6 & 1.615 & 682 & 543 & 168 & 100.2 & 101.564\\
0.7 & 2.858 & 875 & 856 & 144 & 100.1 & 101.671\\
0.8 & 2.645 & 819 & 797 & 127 & 100.8 & 101.847\\
0.9 & 2.184 & 731 & 688 & 113 & 100.8 & 101.673\\
1.0 & 1.820 & 693 & 577 & 101 & 100.0 & 101.047\\
\hline
\multicolumn{7}{c}{$cellLength$ = $0.5\cdot moveLength$}\\
\hline
0.5 & 11.842 & 2346 & 1735 & 202 & 100.5 & 101.373\\
1.0 & 1.752  & 693  & 577  & 101 & 100.0 & 101.047\\
2.0 & 0.266  & 174  & 149  & 51  & 100.0 & 100.873\\
3.0 & 0.170  & 108  & 108  & 35  & 102.0 & 103.745\\
\hline
\end{tabular}}
\end{table}

The upper part of the table explores the effect of having different ratios between $cellLength$ and $moveLength$. It can be seen that when $moveLength$ is shorter than $cellLength$, no path can be found at all. Only one node is expanded and when the successor function is called, it returns the successors, but all of them are not accepted in the $Open$ queue as they lie on the same cell as the parent. When $moveLength$ is the same as $cellLength$ or just longer than it (second and third rows of the table) a path could be found, but it can be seen that the number of nodes inside $Open$ is lower than the case when $moveLength$ is higher (at 0.7 and 0.8) this is because many of the nodes are rejected from $Open$ for the same reason mentioned above. In some cases, this might prevent the planner from finding a path, but here it did not, most likely because of the position of the start node. However it can be seen that the cost of the path when $moveLength=cellLength$ is much higher than the length of the path. This indicates that only successors with large deviation angles were accepted in $Open$ and others were rejected. To prevent nodes from being initially rejected from $Open$ because they lie on the same cell as their predecessors, $moveLength$ must be greater than $\sqrt{2}\cdot cellLength$ as explained in section~\ref{newPosition}. 

At $moveLength$ = 0.9 and 1.0, the number of nodes inserted to $Open$ as well as the number of expanded nodes decrease again. This is because one step  along the planned path with a larger $moveLength$ gets the agent closer to the goal than one step with a smaller value. It can be noticed that, unexpectedly, when the resolution of the planned path is decreased to $1 m$, the found path has a smaller length than when the resolution is higher (at 0.8 - 0.9). This is because of two facts: 1) the length of the path is always a multiple of $moveLength$.\footnote{path length = (number of nodes in the path - 1)$\cdot moveLength$} 2) the cost is composed of the distance and the deviation angle. So with $moveLength$ = (0.8 - 0.9), the smallest number of steps needed to reach the goal at minimum cost made the path longer. It can be seen, however, that for higher resolutions in (0.6 - 0.7) the goal cell can be reached with shorter distances because higher resolution allows more curvature in the planned path and offers more flexibility as the length of the path is a multiple of a smaller numbers. It can be concluded that choosing $moveLength$ to be double the length of the cell offers a good trade\hyp off between allowing as much successors as possible to be included in the search per one node and at the same time keeping the search more efficient.

In the lower part of the table, $cellLength$ is fixed to be half the length of one movement. The effect of changing $moveLength$ is then seen on the efficiency and quality of the search. It can be noticed that as the value of $moveLength$ decreases, the execution time increases exponentially, so is the number of included\footnote{those inserted in $Open$.} and expanded nodes. This suggests that $moveLength$ should be chosen as long as possible with respect to the desired resolution of the road. It can be seen that when $moveLength$ is doubled from 0.5 to 1, the execution time decreases by a factor of 10. At the same time, the length of the resultant path decreases by $0.5 m$, which means that computational time is saved at almost no cost over the quality. The cost and path length when $moveLength = 0.5$ is higher than when $moveLength = 1$ because with smaller values of $moveLength$, the planner is able to produce paths that follow the center line closely, which requires more deviations and more distance. As $moveLength$ is further increased, the execution time still drops significantly but the length of the found path is larger and the cost is higher. This is primarily due to the fact that, with larger values of $moveLength$, the planner is less able to find paths with precise curves such that the center line is followed and small deviations have greater effects on the trajectory of the agent. This shows that the suggested scheme offers a way to increase the ``maneuverability'' at the cost of the computational complexity.

\subsection{MATLAB Simulation Results}

\begin{figure}[htb]
\centering
\includegraphics[width=\textwidth]{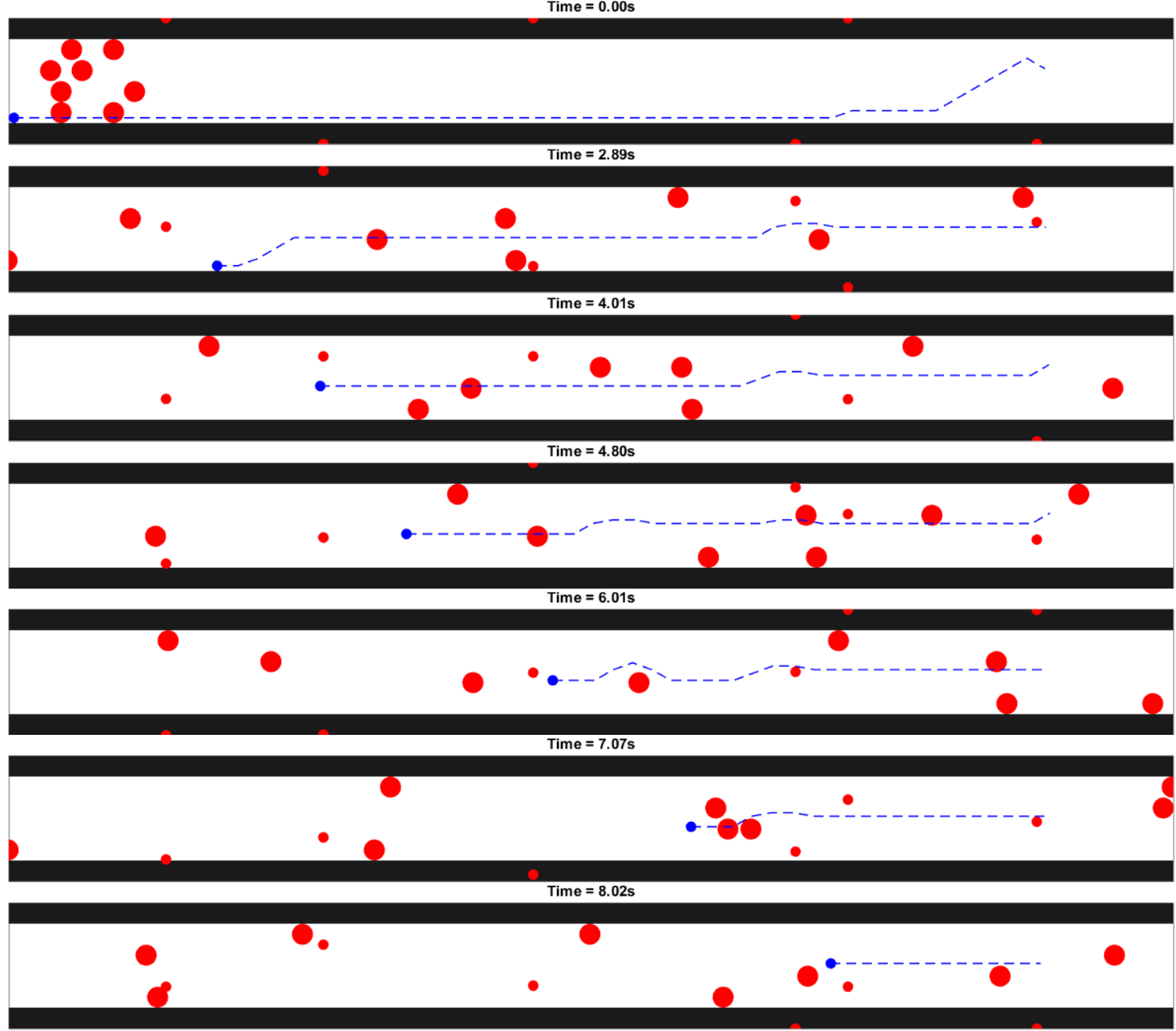}
\caption{Instances from a fixed time\hyp step animation of an agent planning using Inflated A* and optimizes time. $\epsilon = 1.1$, $moveLength = 2m$, $cellLength = 1m$, $speedRange = 0.5m/s$ and initial speed $= 5 m/s$.}
\label{fig:roadAni}
\end{figure}

The ideas presented in section~\ref{newSpeed} were verified through MATLAB simulations using the package described in section~\ref{MATLAB2}. Two search algorithms were implemented and tested. One of them is the same as the modified ARA* algorithm presented in section~\ref{newSpeed}, and the other is the same method but without the anytime feature as mentioned before. At first, the behaviors of the algorithms were tested using fixed time\hyp step animations, which can be created in the package for any \textsf{searchProblem} provided by the user. Figure~\ref{fig:roadAni} shows instances of one of these animations at different points in time. At these shown instances, the agent changes its previously planned path as some changes are observed on the trajectories of the moving obstacles. The agent starts with an initial speed of $5m/s$, and moves to a goal near the end of the road. Note that the 2D figure does not show any information about the timing of the path. Thus, although the shown path may go through some obstacles, the car does not collide with any of them because as it moves the obstacles also move.

\begin{figure}[htb]
\centering
\begin{subfigure}{\textwidth}
  \centering
  \includegraphics[width=\textwidth]{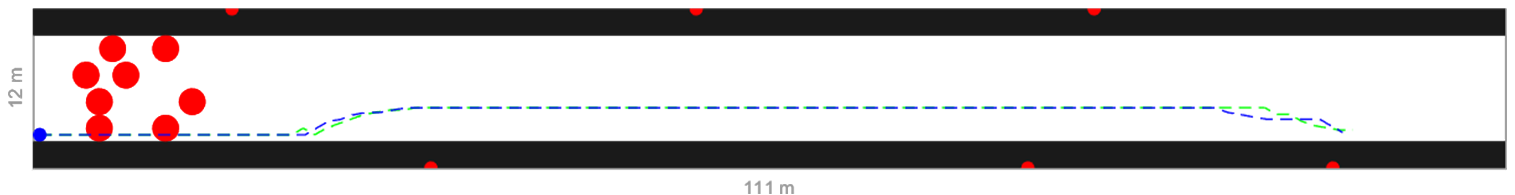}
  \caption{path}
\end{subfigure}
\par\medskip
\begin{subfigure}{\textwidth}
  \centering
  \includegraphics[width=.985\textwidth]{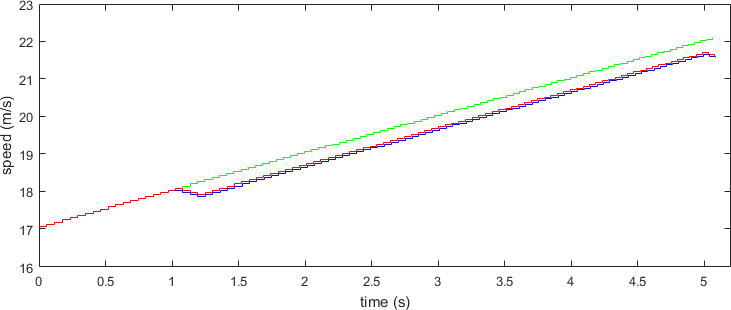}
  \caption{speed profile}
\end{subfigure}
\caption{Results of inflated A* search while optimizing time (green), distance (blue) and both time and distance (red). $\epsilon = 1.1$, $moveLength = 1m$, $cellLength = 0.5m$, $speedRange = 0.01m/s$ and initial speed $= 17m/s$.}
\label{fig:costdt}
\end{figure}

Tests were done also to see the effect of changing the weighting factors of the cost and the heuristic. Figure~\ref{fig:costdt} shows the paths and speed profiles that result from running an inflated A* search with $\epsilon = 1.1$, $moveLength = 1m$, $cellLength = 0.5m$, $speedRange = 0.01m/s$ and initial speed $= 17m/s$. Note that an inflated search is used in the tests presented in this section because running a completely optimal A* search takes a very long time. In the upper part of Fig.~\ref{fig:costdt}, the paths when the search optimizes time ($w_t = 1, w_c = 0$), distance ($w_t = 0, w_c = 1$) and both of them ($w_t = 0.5, w_c = 0.5$) are shown. The speed profiles of these search are shown in the lower part of the figure. It can be seen that search results in the same path and very similar speed profiles when $w_t=w_c = 0.5$.

Table~\ref{costsTests} shows more detailed data from the same search. It can be notices that when the search is optimized for time, the found path takes the shortest time, but the length of the path is slightly longer. This is because, for the agent to continue accelerating, it needs to take longer paths to avoid moving obstacles and arrive faster. It can be noticed also that for this search, optimizing for time made the search faster, expanding less nodes than other weighting factors settings. When the two weighting factors are equal, the planner tries to find paths that have as small distances as possible keeping the needed time as short as possible. It can be seen that this made the search the most computationally expensive among the shown settings for the weighting factors causing expanding the greatest number of nodes. This is primarily due to the used non\hyp informing heuristic as explained before in section~\ref{newSpeed}. The found path is then shorter but the time it takes is slightly larger than when the planner optimizes time only. In the case that distance is what is optimized for, the found path is the shortest possible (equal to the previous case), but the planner gives no attention to the time the path needs. So, the time of the found path is the largest among other settings. The computational cost of this search is just below that of the previous search.

\begin{table}[htb]
\centering
\caption{Inflated A* search data for various cost weights. $\epsilon = 1.1$, $moveLength = 1m$, $cellLength = 0.5m$ and initial speed $= 17 m/s$.}
\label{costsTests}
\resizebox{\textwidth}{!}{\begin{tabular}{ccccccc}
\hline
$w_t$ & $w_c$ & time (ms) & \# $Open$ & \# Expanded & Path length & Path time (ms)\\
\hline
1.0 & 0.0 & 199.8 & 895  & 117 & 100 & 5107.4\\
0.5 & 0.5 & 247.7 & 1113 & 156 & 99  & 5122.9\\
0.0 & 1.0 & 241.6 & 1104 & 150 & 99  & 5133.5\\
\hline
\end{tabular}}
\end{table}

\begin{figure}[htb]
\centering
\includegraphics[width=\textwidth]{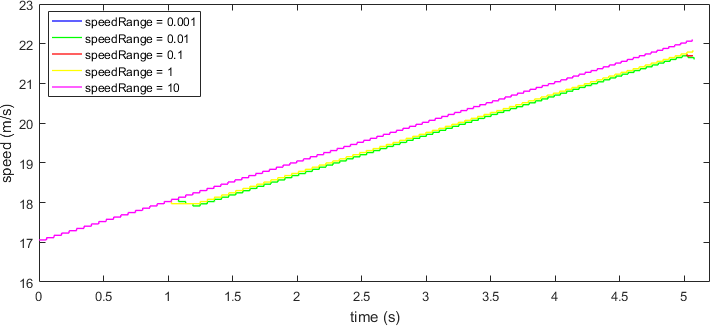}
\caption{Inflated A* speed profiles for various values of $speedRange$. $w_t = w_c = 0.5$, $\epsilon = 1.1$, $moveLength = 1m$, $cellLength = 0.5m$ and initial speed $= 17 m/s$.}
\label{fig:speedR}
\end{figure}

In Table~\ref{speedRTests}, the effect of changing the speed discretization parameter $speedRange$ is examined. The search problem here is the same as the one in Table~\ref{costsTests} with $w_t=w_c=0.5$. It can be seen that as $speedRange$ decreases, the execution time and also the nodes in $Open$ and in $Closed$ decrease. However, this decrease is more like a logarithmic rather than a linear one. Resultant paths have the same distance and time when $speedRange=0.001-0.01$. The time is slightly less when $speedRange = 0.1 - 1$. This is because this search is inflated, so both values for time satisfy the 1.1 optimization bound. As $speedRange$ increases to 10, the path time decreases, but the distance increases. This is because the number of speeds per one cell was very limited such that only accelerating speeds were allowed as they were expanded first (they are favored by the heuristic) blocking any other speed. This can be seen clearly in Fig.~\ref{fig:speedR}, where the speed profiles of these searches are shown. The profile when $speedRange = 10$ is a linearly accelerating one. The profiles when $speedRange=0.001-0.01$ are equivalent to each other, and those when $speedRange=0.1-1$ are very similar. It can be concluded form this test while decreasing $speedRange$ can save computational time and still produce good results, overdoing this can result in losing the optimality of the search. This, however, can be a great start for a new anytime planner that starts with a large value of $speedRange$ and, if time allows, perform successive search cycles gradually decreasing $speedRange$.

\begin{table}[htb]
\centering
\caption{Inflated A* search data for various values of $speedRange$. $w_t = w_c = 0.5$, $\epsilon = 1.1$, $moveLength = 1m$, $cellLength = 0.5m$ and initial speed $= 17 m/s$.}
\label{speedRTests}
\resizebox{\textwidth}{!}{\begin{tabular}{cccccc}
\hline
$speedRange$ & time (ms) & \# $Open$ & \# Expanded & Path length & Path time (ms)\\
\hline
0.001 & 256.3 & 1122 & 163 & 99  & 5122.9\\
0.010 & 247.7 & 1113 & 156 & 99  & 5122.9\\
0.100 & 205.9 & 728  & 128 & 99  & 5113.2\\
1.000 & 179.3 & 397  & 114 & 99  & 5112.7\\
10.00 & 172.1 & 303  & 112 & 100 & 5107.4\\
\hline
\end{tabular}}
\end{table}

The performance of the anytime planner, shown in section~\ref{newSpeed}, on the same search problem used in the previous two tests can be seen in table~\ref{newara}. The planner starts at $\epsilon=\epsilon_0 = 2$ until reaching $\epsilon=1.1$ with a step equal to $\epsilon_{step} = 0.1$. It can be clearly seen that almost all of the search effort is done in the first search cycle. As $\epsilon$ decreases, the planner finds that the planned path already meets the new optimality bounds, and hence no improvement is needed. This continues until moving from $\epsilon=1.3$ to $\epsilon=1.1$ where very small additional number of states are considered. Comparing this to the number of expanded nodes and the exeution time when using an inflated A* search for the same problem (Table~\ref{costsTests}, second row), this test shows that, in many cases, using a normal inflated A* search can be more efficient than using an anytime planner. This same finding was pointed out to in~\cite{hgu}.

\begin{table}[htb]
\centering
\caption{ARA* search data as $\epsilon$ is decreasing. $w_t = w_c = 0.5$, $\epsilon_0 = 2$, $\epsilon_{step} = 0.1$, $moveLength = 1m$, $cellLength = 0.5m$, $speedRange = 0.01$ and initial speed $= 17 m/s$.}
\label{newara}
\begin{tabular}{ccccc}
\hline
$\epsilon$ & time (ms) & \# $Open$ & \# Expanded & \# $Incons$\\
\hline
2.0 & 300.7  & 1055 & 121 & 0  \\
1.9 & 4.2    & 0    & 0 & 0  \\
1.8 & 4.2    & 0    & 0 & 0\\
1.7 & 4.2    & 0    & 0 & 0\\
1.6 & 4.2    & 0    & 0 & 0\\
1.5 & 4.2    & 0    & 0 & 0\\
1.4 & 4.2    & 0    & 0 & 0\\
1.3 & 4.2    & 0    & 0 & 0\\
1.2 & 6.4    & 4    & 2 & 0\\
1.1 & 32.1   & 29   & 17 & 1\\
\hline
\end{tabular}
\end{table}

\chapter*{Conclusion}
\addcontentsline{toc}{chapter}{Conclusion}
In this report, the problem of planning in dynamic environments, using incremental heuristic search and forcing kinematic and dynamic constraints, was tackled. The report started with a clarification of the importance of the path planning component in a highly automated driving system. This was followed by a brief survey of the main methods in the field. It was concluded that incremental heuristic search techniques are the most suitable for the application dealt with here.

In chapter~\ref{chapterStatic}, the different state\hyp of\hyp the\hyp art incremental heuristic\hyp based path planning algorithms were reviewed and discussed in terms of their applications, strengths and weaknesses in the context of static partially\hyp known environment. All of these methods can be considered as improvements over the original Dijkstra's algorithms that finds the shortest paths from a start node to all nodes in a directed graph. This is achieved by inserting nodes in a queue as they are encountered and \emph{expanding} them based on their closeness to the start. 

Broadly speaking, all the methods discussed in chapter~\ref{chapterStatic} try to achieve better results by two main things: use a better strategy for selecting the nodes to be inserted in the queue, and\slash or prioritize them in a way that makes the search more efficient. A*, for example, focuses the search towards the goal by using a heuristic (a lower bound estimate on the cost from a node to the goal) to order the nodes in the queue based on the anticipated total cost from the start to the goal. D* Lite improves that by adding a replanning feature. This allows using previous search efforts to fix the path when a change happens in the graph by focusing the search on the nodes that were affected by the change and are important to fix the path. ARA* has an anytime planning capabilities, allowing it to produce quick suboptimal plans and improve them while resources are available. It limits the nodes to be inserted to the queue such that the search is efficient while at the same time guaranteeing a bound on the optimality of the result. Finally, AD* combines the ideas of the last two methods to have a planner that can plan and replan in an anytime fashion. 

A complete simulation grid-based environment was designed to test and see the differences between the discussed algorithms. The algorithms were also implemented as well as optimized version of them where applicable. This was followed by various tests both in terms of behavior and performance on three designed grids of different sizes. The tests gave valuable insights about the operation of the different algorithms. Special attention was given to the Anytime Dynamic A* (AD*) where the effect of altering the values of its parameters on the performance and behavior was explored. 

Chapter~\ref{chapterDynamic} focused on planning in dynamic environments. It was seen how backward search, with methods like D* lite or AD*, in the configuration\hyp time space makes it very difficult and computationally expensive to produces paths consistent with the kinematic and dynamic constraints of a car\hyp like agent. This is mainly due to three reasons: 1) using backward search requires inserting all possible goal states to the search at the beginning. 2) Applying dynamic constraints in the configuration\hyp time space requires either very dense roadmaps or planning in a 4D space. 3) building the roadmap and performing the preprocessing required to get the heuristic and cost functions is often very time consuming and not suitable in reactive environments. It was also concluded, based on the tests done in chapter~\ref{chapterStatic}, that D* and AD* are not very efficient for the relatively small environments the designed path planning component is supposed to work in.

A new efficient scheme was then suggested based on forward incremental heuristic\hyp based search. The proposed scheme plans in the configuration\hyp speed space, builds the roadmap as needed while searching, and checks for collisions with moving obstacles using a new method named ``minimum collision time''. It produces paths that are consistent with the kinematic and dynamic constraints of a car\hyp like agent and safe from colliding with static and dynamic obstacles in the environments. The provided scheme has also an anytime feature allowing it to quickly produce a suboptimal path and continue improving the produced path as time allows. In addition to that, it allows searching for paths that optimize time, distance and\slash or other factors. 

The proposed scheme was tested using a designed MATLAB simulation package and an artificial road containing many moving obstacles, as well as a C++ implementation used with real DLR road data. The various tests show the efficiency and capabilities of the technique. It was found that using normal A* might be, in many cases, more efficient than using an anytime planner.

The performed tests show that the proposed method has a great potential. The next step in this research should be continuing to explore the new scheme with extensive tests as well as trying to provide theoretical mathematical proofs for the statements in sections~\ref{newPosition} and \ref{newSpeed} regarding position and speed discretization. More efforts should be put also into finding a better informing heuristic, especially with respect to time.

\chapter*{Acknowledgments}
\addcontentsline{toc}{chapter}{Acknowledgments}
I would like to express my gratitude to my supervisor in DLR, Dr.~Reza Dariani for giving me this opportunity to work on such an interesting topic. Thanks for your time and efforts in helping me.

\end{document}